\documentclass[10pt,journal,compsoc]{IEEEtran}
\ifCLASSOPTIONcompsoc
  \usepackage[nocompress]{cite}
\else
  \usepackage{cite}
\fi

\usepackage[T1]{fontenc}

\usepackage{xcolor,soul,framed} 

\colorlet{shadecolor}{yellow}
\usepackage[pdftex]{graphicx}
\graphicspath{{../pdf/}{../jpeg/}}
\DeclareGraphicsExtensions{.pdf,.jpeg,.png}

\usepackage[cmex10]{amsmath}
\usepackage{array}
\usepackage{mdwmath}
\usepackage{url}
\usepackage{chngcntr}
\usepackage{subfigure}
\usepackage{multirow}
\usepackage{cite}
\usepackage{amsmath}
\usepackage{amssymb}
\usepackage{booktabs}
\usepackage{setspace}
\usepackage{enumerate}
\makeatletter  
\newif\if@restonecol  
\makeatother

\usepackage[linesnumbered,ruled, vlined]{algorithm2e}
\usepackage{algpseudocode}  
\begin{document}
%
\title{Performance-aware Approximation of Global Channel Pruning for Multitask CNNs}
%
%
%

\author{Hancheng~Ye,
        Bo~Zhang,
        Tao~Chen,~\IEEEmembership{Senior Member,~IEEE},
        Jiayuan~Fan,
        and Bin~Wang,~\IEEEmembership{Senior Member,~IEEE}
\IEEEcompsocitemizethanks{\IEEEcompsocthanksitem H. Ye, T. Chen, and B. Wang are with School of Information Science and Technology, Fudan University, Shanghai 200433, China. E-mail: \{yehc20, eetchen, wangbin\}@fudan.edu.cn.\IEEEcompsocthanksitem B. Zhang is with Shanghai AI Laboratory, Shanghai 200232, China. E-mail: zhangbo@pjlab.org.cn.\IEEEcompsocthanksitem
J. Fan is with Academy for Engineering and Technology, Fudan University, Shanghai 200433, China. E-mail: jyfan@fudan.edu.cn. \IEEEcompsocthanksitem This work is supported by National Natural Science Foundation of China (No. 62071127, U1909207, and 62101137), Zhejiang Lab Project (No. 2021KH0AB05). (Corresponding author: T. Chen)}}%

%
%

\markboth{IEEE TRANSACTIONS ON Pattern Analysis and Machine Intelligence}%
{Ye \MakeLowercase{\textit{et al.}}: Performance-aware Approximation of Global Channel Pruning for Multitask CNNs}
%



\IEEEtitleabstractindextext{
\begin{abstract}

Global channel pruning (GCP) aims to remove a subset of channels (filters) across different layers from a deep model without hurting the performance. Previous works focus on either single task model pruning or simply adapting it to multitask scenario, and still face the following problems when handling multitask pruning: 1) Due to the task mismatch, a well-pruned backbone for classification task focuses on preserving filters that can extract category-sensitive information, causing filters that may be useful for other tasks to be pruned during the backbone pruning stage; 2) For multitask predictions, different filters within or between layers are more closely related and interacted than that for single task prediction, making multitask pruning more difficult. 
Therefore, aiming at multitask model compression, we propose a Performance-Aware Global Channel Pruning (PAGCP) framework. We first theoretically present the objective for achieving superior GCP, by considering the joint saliency of filters from intra- and inter-layers. Then a sequentially greedy pruning strategy is proposed to optimize the objective, where a performance-aware oracle criterion is developed to evaluate sensitivity of filters to each task and preserve the globally most task-related filters. Experiments on several multitask datasets show that the proposed PAGCP can reduce the FLOPs and parameters by over 60\% with minor performance drop, and achieves 1.2x$\sim$3.3x acceleration on both cloud and mobile platforms. Our code is available at http://www.github.com/HankYe/PAGCP.git.

\end{abstract}

}
\maketitle
\IEEEdisplaynontitleabstractindextext

%
\IEEEpeerreviewmaketitle

\ifCLASSOPTIONcompsoc
\IEEEraisesectionheading{\section{Introduction}\label{sec:introduction}}
\else
\section{Introduction}
\label{sec:introduction}
\fi
\IEEEPARstart{T}{o} date, Convolutional Neural Network (CNN)-based models have developed rapidly in large-scale multitask prediction, yet most of them still suffer from huge computational cost and storage overhead, preventing their deployment in real-time applications, such as the autopilot system and edge computing devices. Model compression as an effective technique to relieve this problem, has advanced a lot and can be roughly divided into several types including knowledge distillation \cite{romero2014fitnets, polino2018model, chung2020feature}, network quantization \cite{wu2016quantized, gu2019projection}, and model pruning \cite{han2015deep, 8416559, yu2018nisp, frankle2018lottery, liu2020autocompress}. Knowledge distillation attempts to transfer the learned knowledge of a large model into a small one by ensuring the prediction consistency between two models, and network quantization aims to reduce the model size by projecting the network weights from float-point type to low-bit type. On the other hand, model pruning as one prevalent compression technique has been proposed, which can be divided into two categories: 1) weight pruning~\cite{han2015deep, frankle2018lottery} and 2) channel pruning~\cite{8416559, yu2018nisp, liu2020autocompress}. The former maintains the model accuracy by generating non-structured sparse connections. But such a non-structured connection inevitably produces filters with irregular sizes, further resulting in the poor generalization ability across different hardware devices with a specialized design for efficient inference.

In contrast, channel pruning can reduce the inference time by introducing the structured sparsity to the whole filter or channel. For example, by minimizing the reconstruction error of the important feature maps, a channel pruning technique is developed to correctly select the representative channels \cite{8237417}. Besides, ThiNet \cite{8416559} determines the current layer's filter saliency based on the output of the next layer, thus developing a greedy algorithm to select a subset of filters to compress the model. More recently, a data-independent pruning method \cite{coreset} is designed, which constructs a filter coreset layer-by-layer according to whether the layer can be approximated by such a filter coreset. Joo \textit{et al. }\cite{Joo_Yi_Baek_Kim_2021} propose a linearly replaceable filter pruning method, which suggests that a filter can be pruned if it can be linearly approximated by other filters.

\begin{figure*}
    \centering
    \scalebox{0.6}{\includegraphics{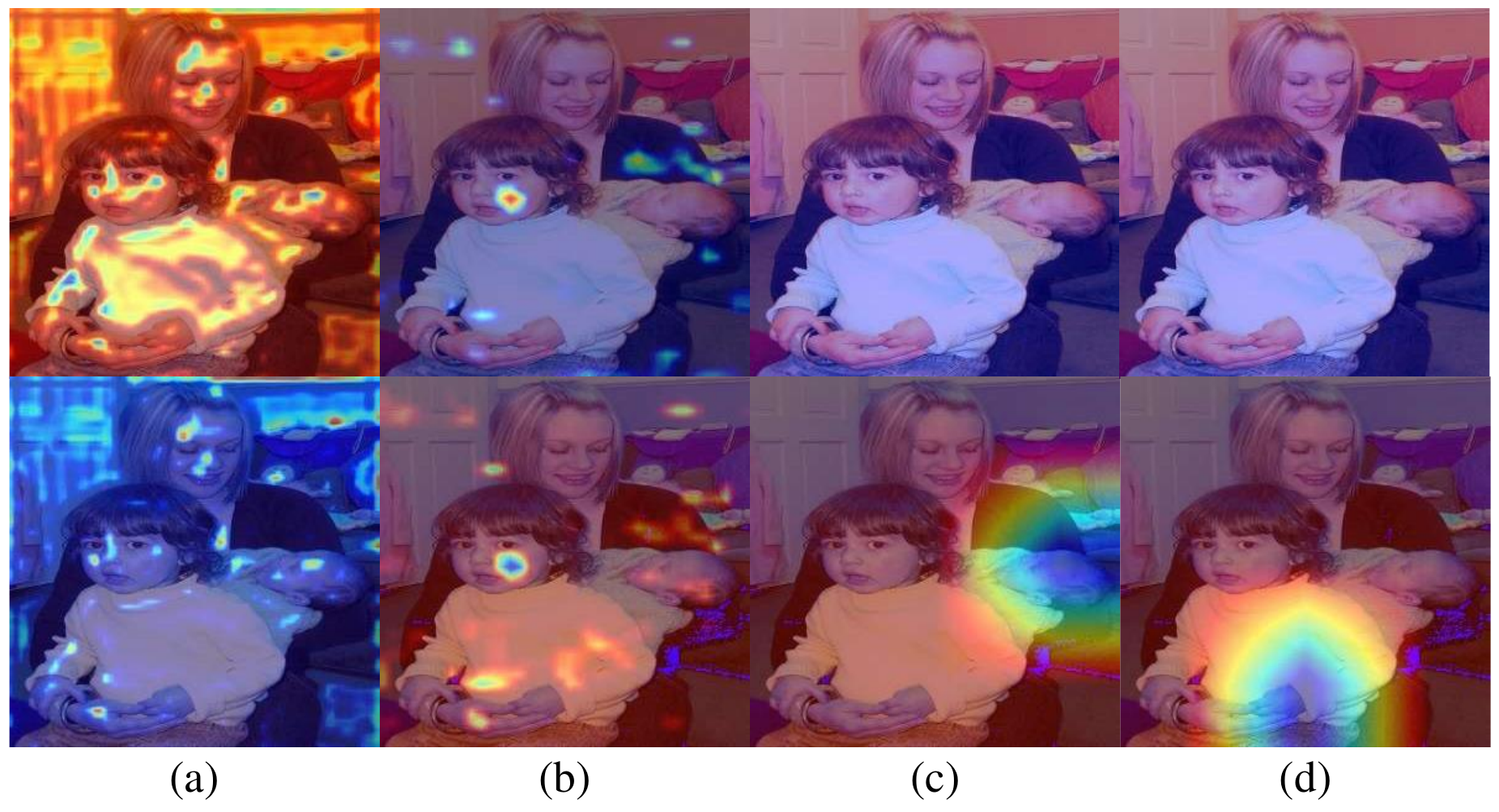}}
    \caption{Visualization of the co-influence among different filters during the pruning process. We take a single-stage detector VGG16-SSD \cite{liu2016ssd} as an example, and use the Grad-CAM technology \cite{gradcam} to generate the attention map of each filter in SSD. The upper and lower rows respectively illustrate the attention maps of filters in layer (a) conv3\_2, (b) conv4\_1, (c) extras.3, and (d) extras.5 before and after several pruning steps. Column (a) shows the response decrease after pruning other filters, while the other three columns show the response increase after pruning other filters.}
    \label{fig1}
\end{figure*}

Pruning multitask networks is more challenging than pruning single-task ones, \textit{e.g.}, an object classification network, since filters in the multitask networks often need to learn representations that can simultaneously serve multiple tasks. As a result, the filters across different layers present stronger relevance and interactions, and the pruning of one filter will also impact another filter's performance for different tasks. This is neglected by previous single task and multitask based compression research. A typical example is that the sensitivity ranking of multiple tasks for the same filter (meaning the performance drop of each task caused by dropping the filter) may change, if other filters are pruned. Another example is the change of response areas after pruning other filters, which is illustrated in Fig. \ref{fig1}. It shows that the high response area in the attention map of a filter may be greatly impacted by the pruning of other filters. Overall, the dynamically changing filter importance (meaning the saliency of a filter) in the multitask pruning process, may cause the difficulty in finding an optimal balance between the multitask performance drop and compression ratio.

Even though some works start to design pruning techniques for multitask applications, they are mainly based on classification-based pruning strategies, \textit{i.e.}, fine-tuning the pre-pruned classification backbone with multiple task-specific branches on multitask benchmarks \cite{liu2018rethinking}, or adapting classification-based pruning strategies \cite{Singh_2019, XIE2020400} under multitask settings. However, these approaches face two limitations as follows.

First, the features from the pre-pruned classification-based backbone can not well serve for multitask use, which can be considered as a filter mismatch problem between the single-task image classification and different multitask models. For example, many shallow layer channels with rich spatial details may be pre-pruned in a classification backbone, as these channels contain limited high-level semantic information required for classification. But such channel pruning will negatively impact a detection model's localization performance as spatial details are important for object positioning, which finally restrains the total detection performance when fine-tuning a classification-based backbone on a detection task.

Secondly, the joint impact of pruning filters from different layers on the compression performance is neglected in previous works. Such neglect may not hurt the compression performance of classification models, since much redundant information such as background features and low-level features, that is of little help to the classification task exists and can be discarded. In other words, massive correlated filters from different layers in classification models are redundant and have great potentials to be removed. However, for multitask models, as mentioned above, the complex interactions cause the pruning of one filter to highly impact the performance of another one. Thereby, the neglect of joint impact may result in a worse evaluation of filters' pruning potential in multitask models when adapting various classification-based pruning strategies under multitask settings.

\begin{figure*}
    \centering
    \scalebox{0.5}{\includegraphics{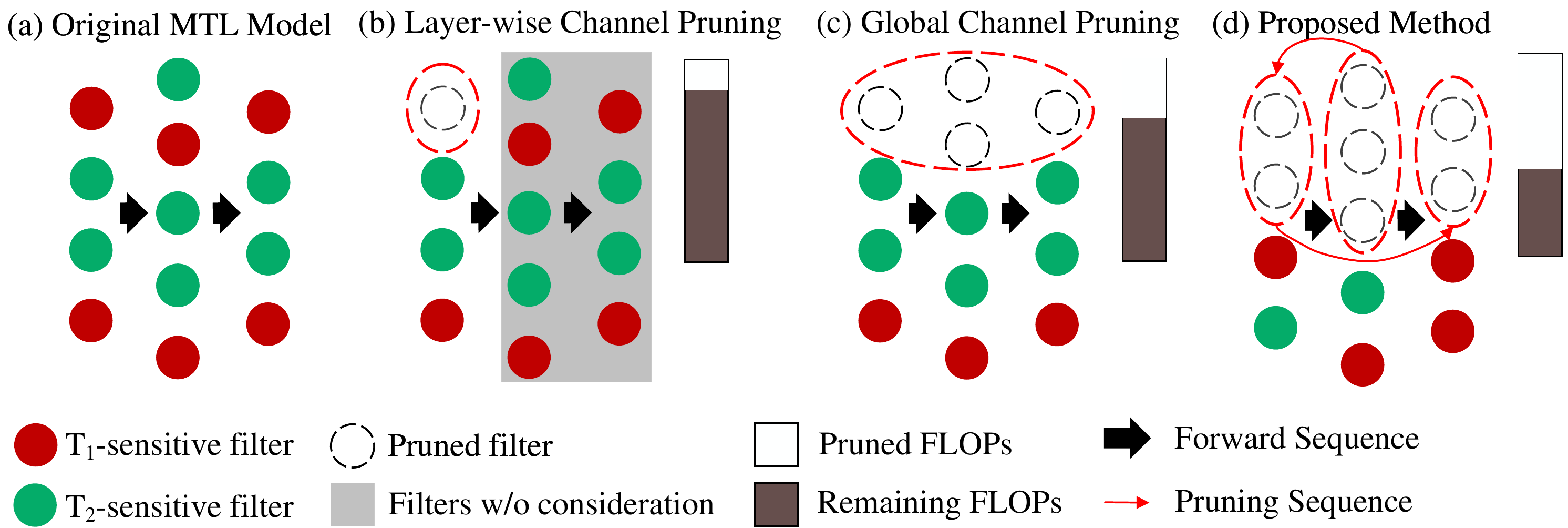}}
    \caption{\label{fig2.1} The distinction between different pruning strategies in a single pruning iteration. The red dashed circle denotes the set of filters to be pruned at each iteration. (a) A general multitask model, in which each filter has different sensitivity to different tasks. The red circle refers to the filter that is most sensitive to Task 1, and the green circle refers to the filter that is most sensitive to Task 2. (b) Layer-wise channel pruning, which evaluates and prunes filters in a single layer and fine-tunes the model at each iteration. The filters in the gray area are not considered in the current pruning iteration. (c) Previous global channel pruning, which evaluates and prunes filters across all layers based on static filters' relations at each pruning iteration (represented in the unchanged colors of filters). (d) Our PAGCP method, which evaluates and prunes the filters in a single layer dynamically based on the previously pruned structure, and globally considers both intra- and inter-layer filter relations for the compression performance of multitask models (represented in the color change of filters). The layer pruning sequence is determined by each layer's contribution to the total FLOPs reduction.} 
    
\end{figure*}

To this end, we focus on multitask model compression, and propose a Performance-Aware Global Channel Pruning (PAGCP) framework. First, we present the formularized objective of Global Channel Pruning (GCP) under the multitask setting, where the \textbf{joint filter saliency} is fully considered in the GCP process. \textbf{Joint filter saliency refers to the co-importance of multiple filters from intra- and inter-layers evaluated by a designed performance-related metric}. According to the theoretical derivation, a sequentially greedy pruning strategy and a performance-aware oracle criterion are developed to solve the objective problem of GCP. The developed sequential pruning strategy dynamically computes the joint filter saliency based on the previously pruned structure, and selects the filters and layers to be pruned in a greedy fashion. Moreover, to address the filter mismatch issue, the performance-aware oracle criterion evaluates the sensitivity of filters to each task, and adaptively preserves the most task-related filters in each layer. In this way, the model can maintain superior multitask performances by controlling the local performance drop of the most sensitive task in each layer.

Compared with the previous methods that adapt the filter saliency criterion for classification models to the multitask ones, our approach dynamically computes the saliency of each filter in a greedy fashion to approximate an optimal GCP, and automatically controls the performance drop of pruning by the performance-aware oracle criterion, which offers the following advantages. First, our pruning is performed without penalty, yet with a large reduction in both parameter numbers and FLOPs. Second, compared with previous works, the performance drop achieved by our method is more controllable. Traditional determination of the pruning ratio relies on either the independent sensitivity analysis in layer-wise pruning strategies or a global small threshold in global pruning strategies, yielding imperceptible performance drop after pruning, as shown in Fig. \ref{fig2.1}.

To demonstrate that the proposed approach can be generalized to different multitask settings, we apply the proposed PAGCP on detection models that contain classification and localization tasks, \textit{e.g.}, SSD \cite{liu2016ssd}, Faster R-CNN \cite{Ren2015Faster}, CenterNet \cite{zhou2019objects} and YOLOv5m \cite{glenn2021yolov5}, and other conventional multitask models that include multiple dense prediction tasks, \textit{e.g.}, MTI-Net \cite{vandenhende2020mti}. We conduct experiments on several object detection and Multi-Task Learning (MTL) datasets including PASCAL VOC \cite{everingham2015pascal}, COCO \cite{lin2014microsoft}, NYUD\_v2 \cite{silberman2012indoor} and PASCAL Context \cite{context}. The empirical results and insightful analyses show that the proposed PAGCP achieves superior performance over the state-of-the-art methods on these models under the same compression constraint. And we also specialize the PAGCP to the single-task setting, \textit{i.e.}, image classification models like ResNet \cite{resnet}, further validating its superiority for compressing various CNN models. Finally, the latency of the pruned model on both cloud and mobile platforms is reported to show the acceleration effect of our proposed approach.

The main contributions of our work can be summarized into the following three points:
\begin{itemize}
    \item[1)] We formulate the global channel pruning for multitask model as a joint filter saliency optimization problem, which considers the co-influence of filters from both intra- and inter-layers on the compression performance of multitask models, and theoretically prove that such joint saliency of filters is upper-bounded by the sum of independent saliency and conditional saliency. To our knowledge, this is the first work to study the joint filter saliency in multitask model channel pruning.
    \item[2)] According to the theoretical analysis, we develop a novel pruning paradigm containing a sequentially greedy channel pruning algorithm and a performance-aware oracle criterion, to approximately solve the derived objective problem of GCP. The developed pruning strategy dynamically computes the filter saliency in a greedy fashion based on the pruned structure at the previous step, and control each layer's pruning ratio by the constraint of the performance-aware oracle criterion.
    \item[3)] We conduct extensive experiments under different multitask settings, including the typical object detection and the conventional multitask learning, on different benchmarks. Results show that our method outperforms existing pruning-based multitask models and other lightweight models consistently for all tasks. Meanwhile, our pruned model achieves faster inference speed on both cloud and mobile platforms, approaching the latency of other lightweight models.
\end{itemize}

The reminder of this paper is organized as follows. Section \ref{related_works} briefly introduces the related works on multitask models, channel pruning and channel pruning specially for multitask models. In Section \ref{method}, we first theoretically model the global channel pruning process, and give the formulated objective of GCP. Then to solve the objective problem, a sequentially greedy channel pruning strategy and a performance-aware oracle criterion are developed and presented in detail. Extensive experimental results of our method are presented in Section \ref{experiment}, followed by the discussion in Section \ref{discussion} and conclusion in Section \ref{conclusion}.

\section{Related Works}
\label{related_works}
Our work is closely related to multitask models and channel pruning. We review the related works on the two fields and compare our method with the existing methods in the following part.
\subsection{Multitask Models}
Multitask models here refer to CNN-based models with multiple task-specific heads and optimization objectives, such as typical object detectors \cite{he2015Spatial,Ren2015Faster, liu2016ssd,zhang2018single,redmon2016you} which include classification and regression heads. Besides, conventional multitask models \cite{misra2016cross, xu2018pad, gao2019nddr, liu2019end, vandenhende2020mti} which include multiple dense prediction tasks like semantic segmentation, depth estimation, surface normal estimation, etc., are also studied for long and become the mainstream for MTL research. In the following, we will review the object detection models and the conventional multitask models that are the two main pruning focuses in this work.

\subsubsection{Object Detectors}
CNN-based object detection can be roughly categorized into two classes: two-stage detectors \cite{he2015Spatial,Ren2015Faster} and single-stage detectors  \cite{liu2016ssd,zhang2018single,redmon2016you}. Typically, two-stage detectors divide the detection process into the initial proposal generation and the subsequent proposal refinement stage, where the Region-of-Interest (RoI) pooling plays a crucial role by mapping proposals of different sizes to fixed-size features for subsequent proposal refinement. In contrast, single-stage object detectors as a dense prediction task detect multiple objects without relying on the RoI pooling. Two-stage detectors often have more tasks than single-stage ones, which need to generate the coarse location and objectness scores of proposal objects in the first stage, followed by the object classification and localization in the second stage.

These object detection models are hard to be deployed on resource constrained devices, which is mainly due to their huge computational cost and storage overhead. Thus, it is necessary to compress detectors and accelerate the object detection speed for resource-constrained mobile applications.

\subsubsection{Conventional Multitask Models}
Conventional multitask models \cite{misra2016cross, xu2018pad, gao2019nddr, liu2019end, vandenhende2020mti} perform multiple dense prediction tasks such as semantic segmentation, depth estimation, surface normal estimation, simultaneously. These MTL models require more interactions between multiple tasks in different stages to learn more complementary features from different tasks. CNN-based multitask models can be divided into encoder-focused and decoder-focused models. Commonly, both types adopt the one-encoder-multiple-decoder structure. The former conducts the interaction in the encoding stage by sharing parameters \cite{misra2016cross, gao2019nddr}, while the latter does this in the decoding stage \cite{xu2018pad, liu2019end, vandenhende2020mti}, which can perceive more discriminative information from ground-truth labels.

Although multitask models save weight parameters through sharing the encoder, they still suffer from the heavy computation and memory cost in the interaction stage, where cross-task features are propagated by many attention operators \cite{liu2019end, vandenhende2020mti}. Thus, they still need compression for resource-constrained mobile applications. In addition, as filters in MTL models present more interactive correlations to serve multiple tasks together, the pruning of MTL models is thus more difficult than that of single task models.

\subsection{Channel Pruning}
Existing channel pruning approaches can be roughly categorized into two classes: saliency-based and sparsity-based. Saliency-based approaches hypothesize that the weight values indicate their importance to the final prediction. A heuristic idea in this way is that those weights with smaller magnitudes make little contribution to the output, and thus are less informative to the model. $\ell_1$ norm on either filters \cite{KDSG17} or activation maps \cite{hu2016network} is consequently computed as the filter saliency for model compression. Liu \textit{et al.} \cite{liu2017learning} hold that the saliency of filters could be represented by the scaling factor of the Batch Normalization layer. Yet, some works study the filter saliency from the oracle view based on each filter's direct contribution to the final loss. Molchanov \textit{et al.} \cite{8953464} approximate the importance of filters via ranking the first-order Taylor coefficients. Considering that the global loss computation is time-consuming, a few works focus on using the local reconstruction loss as the saliency criterion. In \cite{8237417, 8416559}, the filters with little impact on the output feature maps are considered as less important. Apart from these, there are other criteria to rank the filter importance \cite{8953212, 9156677}.

Sparsity-based approaches aim at exploring the pruning ratio of each layer in a global or local manner. In the global manner, the sizes of all layers in the model are preset with the same compression ratio before pruning. Howard \textit{et al.} \cite{howard2017mobilenets} apply the same pruning ratio to all layers, while Tan \textit{et al.} \cite{pmlr-v97-tan19a} further scale the depth and resolution together with the channels. Such pruning schemes are coarse for the ignorance of individual sensitivity in each layer, which is decisive to the pruning performance. In contrast, the local manner personalizes the pruning ratio of each layer with two general norms. One norm is sensitivity analysis based on saliency criteria \cite{KDSG17,8237417,liu2018rethinking}. The other norm is the global automatic search for the pruning ratio of each layer, such as \cite{liu2017learning, liu2019metapruning, AMP, ye2022efficient, ye2022beta, li2022pruning, guo2021towards}. Although automatic searching can avoid lots of hand-crafted choices, it also involves more optimization operations, resulting in time-consuming channel pruning.

Thus far, both paradigms have made much progress but still suffer from two dilemmas: the neglect of the joint impact on the compression performance by simultaneously pruning multiple filters, and the imperceptibility of the performance drop during the pruning process. The joint impact of multiple filters remains under-explored in the current works, which should be emphasized in the compression of multitask models due to the complex interaction among different filters. The performance drop as another key issue, is directly related to the prunability \cite{resrep}, \textit{a.k.a.} pruning ratio of a model. Most current works empirically set a small pruning threshold for each layer, which does not fully consider different layers' pruning potentials and different filters' co-influence for the multiple tasks, and thus are insufficient in controlling the performance drop at an optimal level.

\begin{figure*}[t]

\setlength{\abovecaptionskip}{0.1cm}
\setlength{\belowcaptionskip}{-0.5cm}

  \centering
  \scalebox{0.27}{
  \includegraphics{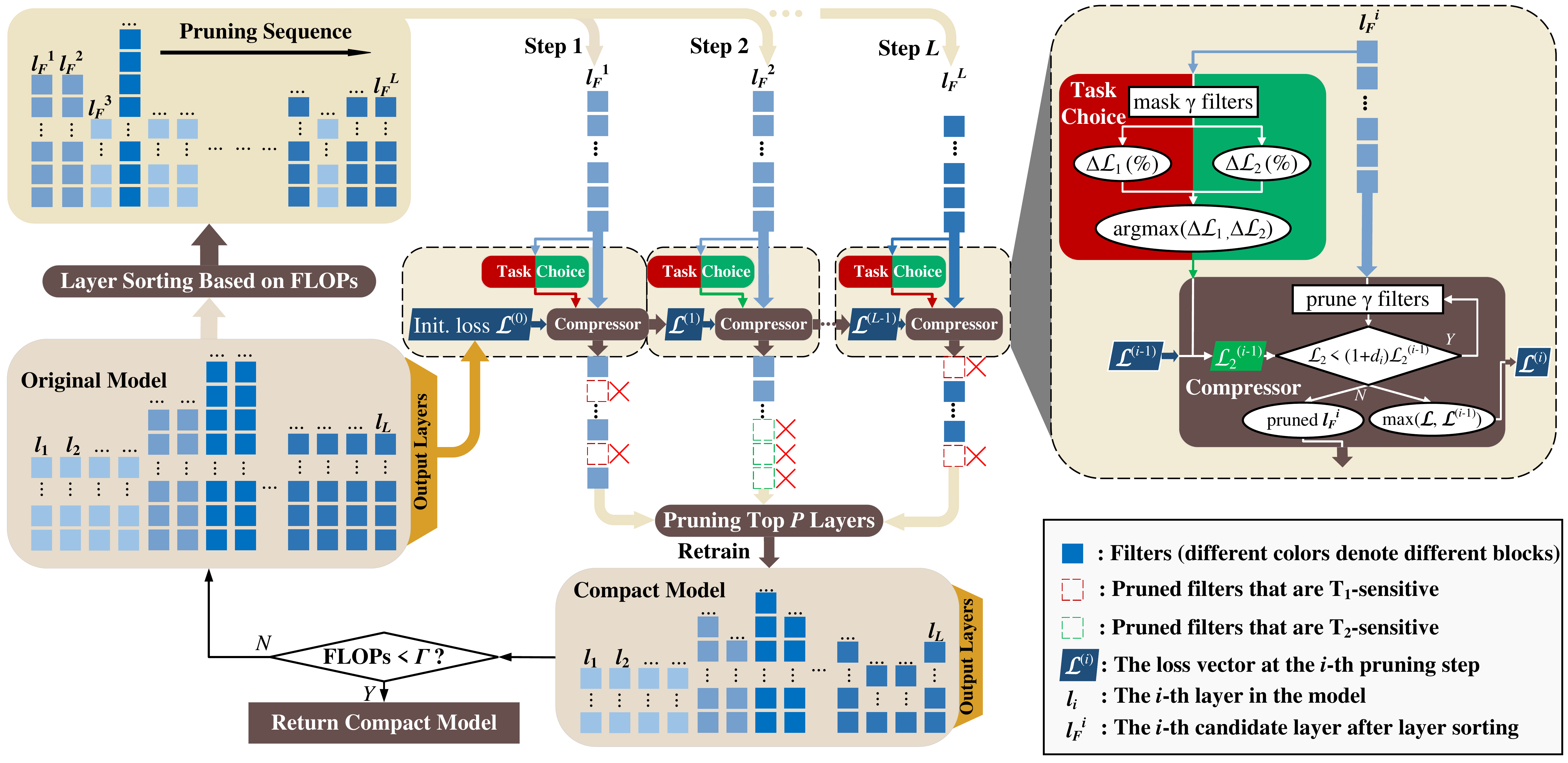}}
  \caption{The overview of the proposed Performance-Aware Global Channel Pruning (PAGCP) framework. Given an original well-trained multitask model, we sort all target layers in a new sequence based on each layer's contribution to total FLOPs reduction (computed by subtracting the FLOPs of the pruned model from the FLOPs of the original model), and compress each layer in such sequence. Specially, the original model produces an initial loss $\boldsymbol{\mathcal{L}}^{(0)}$ for the estimation of pruning ratio in $l_F^1$. Then at each step $i$, we select the task with largest performance drop when masking $\gamma$ filters in $l_F^i$ as the most sensitive task for the target filters to be pruned. In the compressor, based on the selected task and the saliency criterion, we maximize the compression ratio of $l_F^i$ under local constraints with $\boldsymbol{\mathcal{L}}^{(i-1)}$ generated from step $i-1$ as a reference. The compressor outputs a list of the pruned filters and updates $\boldsymbol{\mathcal{L}}^{(i)}$ for step $i+1$. After all layers are compressed, we reorder the pruned layers by evaluating their compression contributions again, and compress the top $P$ layers with the highest pruning ratios. Finally, we retrain the pruned model and repeat the above procedure until the reduction requirement of FLOPs or parameters is satisfied.}
  \label{fig2} 
\end{figure*}

\subsection{Channel Pruning for Multitask Models}


Typical channel pruning methods for multitask models mainly follow two practices: fine-tuning the pre-pruned classification backbone on multitask benchmarks (fine-tuning based pruning), and adapting the classification-based pruning strategy to multitask model pruning (adapting based pruning). Liu \textit{et al. } \cite{liu2018rethinking} validate the filter mismatch problem in the fine-tuning based pruning methods, and argue that the pruned model of fine-tuning based methods can perform well by training from scratch. ThiNet \cite{8416559} transfers the greedy channel pruning with the local reconstruction regularization to the detector compression. MLP \cite{Singh_2019} adapts the group pruning methods in \cite{KDSG17} to compress the SSD models by $\ell_1$ norm. A localization-aware auxiliary network is designed in \cite{XIE2020400} to find out important channels of detectors, which is adapted from the discrimination-aware channel pruning for classification \cite{zhuang2018discrimination}. PAM \cite{he2021pruning} merges multiple networks into a multitask network to eliminate redundancy across tasks before network pruning, and the pruning strategy is inspired by the existing classification-based pruning method. These methods face one common drawback that the co-importance of filters from different layers or groups is neglected, which should get more attention in the multitask setting, since filters are coupled and interacted more tightly to represent the multitask features than those in the single task models.

\section{Problem and Solution} 

\label{method}
The purpose of this work is to study the superior global channel pruning by considering the joint saliency of filters across different layers. Similar to the joint filter saliency, here the saliency denotes the filter importance evaluated by a certain metric, determining which filter to prune. Usually, a higher saliency score indicates more chances to keep this filter. Based on this definition, we first derive the formulated objective of performance-aware global channel pruning, and then propose our pruning strategy to solve the objective problem. The overall framework of the strategy is shown in Fig. \ref{fig2}.

\subsection{Preliminaries}\label{preliminary}

Given a well-trained model with $L$ target layers (which denotes the layers to be pruned\cite{resrep}), traditional saliency-based methods aim at minimizing the penalty-attached objective, in which the penalty denotes the designed saliency criterion and can be formulated as follows:

\begin{equation}\label{traditional objective}
\begin{aligned}
\min_{\boldsymbol{\theta }\in \varOmega, \varTheta} \,\,& \mathbb{E}_{x\sim D_x}\left[\mathcal{L}\left(x; \boldsymbol{\theta},\varTheta, y\right) + \beta\mathcal{S}\left(x;\boldsymbol{\theta},\varTheta, y \right) \right], \\
&\text{s.t. }\boldsymbol{g}\left( \boldsymbol{\theta } \right) \leqslant \boldsymbol{\alpha}, 
\end{aligned}
\end{equation}
where $\mathcal{L}(\cdot)$ denotes the task-related objective function (\textit{e.g., }cross-entropy for the classification task, or the weighted sum of multiple objectives for multitask learning) and $\mathcal{S}(\cdot)$ denotes the saliency criterion such as $\ell_1$ norm of filters, which is used to evaluate the importance of filters. The $x$ and $y$ represent the input image and its ground-truth label(s), respectively, where $x$ obeys the distribution of $D_x$. For convenience, we omit $y$ in the following part. Besides, $\boldsymbol{g}(\cdot)$ denotes the constraint vector determined by several factors including the reduction rates of FLOPs and parameters, while $\boldsymbol{\alpha}$ is a threshold vector with each dimension corresponding to each above factor. The $\varOmega$ is the set of all pruning choices (solutions) in the original model, and $\boldsymbol{\theta}$ represents one choice in $\varOmega$, in which the element $\theta_{lk}$ is a binary variable denoting the pruning state of the $k$-th filter in the $l$-th layer, \textit{i.e.}, $\theta_{lk} \in \left\{ 0, 1 \right\}, l=1, 2, \dots, L, k=1, 2, 3, \dots, K_{l}$, where $K_{l}$ is the filter number of the $l$-th layer. $\varTheta$ represents the total set of parameters. Usually, the constraint $\boldsymbol{g}(\cdot)$ is parameterized by $\boldsymbol{\theta}$, but its expression formula varies in different typologies of models. An explicit constraint on $\boldsymbol{\theta}$ is imposed as: $\lVert \boldsymbol{\theta } \rVert _0\leqslant \eta$, where $\lVert \cdot \rVert _0$ refers to the zero-norm of $\boldsymbol{\theta}$ and $\eta$ is the threshold of the remaining filter number.

As analyzed in ResRep \cite{resrep}, once the second item (penalty) is attached to the objective, the gradient of the first term (performance loss) will compete against the second. The gradient of the penalty makes the parameters of every filter deviate from the optima of the original objective function, changing the original importance of every channel. Different from ResRep \cite{resrep} which designs a mask and a compactor to decouple the optimization process of the two terms, our method proposes to fix the parameters of the well-trained model and evaluate the importance of every channel based on original filter parameters. In this way, the first term in Eq. (\ref{traditional objective}) can be dropped without impacting filters' contributions to the final performance, thus eliminating the competition between two items. Note that we add the performance loss into the constraint list to compensate for the neglect of performance loss after discarding the first item. The motivation behind this adjustment is that we can compress a well-trained model without penalizing weights by exploring the potential maximum number of redundant filters in a model, which will not hurt the important filters' impact on the final performance. Thus, our goal is to adaptively determine the compression ratio, grounded by the current network structure and filter weights. 

Therefore, we transform Eq. (\ref{traditional objective}) into Eq. (\ref{pruning model}), which aims at removing those filters producing the minimum saliency scores subject to several compression constraints.
\begin{equation}\label{pruning model}
\min_{\boldsymbol{\theta }\in \varOmega} \,\,\mathbb{E}_{x\sim D_x}\left[ \mathcal{S}\left( x; \boldsymbol{\theta}\right) \right] \,\,\text{s.t. }\boldsymbol{g}\left( \boldsymbol{\theta } \right) \leqslant \boldsymbol{\alpha}.
\end{equation}

To get the optimal filter subset $\boldsymbol{\theta}$ to be pruned, unlike previous works that calculate the filter saliency independently without considering the impact on the saliency from pruning other filters, we propose a \textbf{P}erformance-\textbf{A}ware \textbf{G}lobal \textbf{C}hannel \textbf{P}runing (PAGCP) framework that considers the joint filter saliency from intra- and inter-layers, and controls the pruning ratio of each target layer by the performance-aware criterion. In the following, we first derive the optimization objective of PAGCP, then introduce an approximated solution: sequentially greedy channel pruning based on a performance-aware oracle criterion.

\subsection {Theoretical Derivation of the PAGCP Objective}
Generally, the saliency of a candidate filter \textit{w.r.t} the sample $x$ in different pruning choices (\textit{i.e.}, $\theta \in \left\{0,1\right\}$, 0 for pruning and 1 for not) can be formulated as:
\begin{equation}\label{sensitivity}
\begin{aligned}
     \mathcal{S}\left(x;\theta_{lk}\right) &= \lVert \boldsymbol{f}\left( x; \theta _{lk}\right) -\boldsymbol{f}\left( x;\theta _{lk}=0\right) \rVert _r\\
     &\triangleq\mathcal{S}_{l}^{k}\left(x \right), 
\end{aligned}
\end{equation}
where $\boldsymbol{f}(\cdot): \mathbb{R}^m\longrightarrow\mathbb{R}^n$ is a pre-defined state function of a model, for instance, the model prediction loss \cite{8953464}, or the output of the activation map in the next layer \cite{8237417}. $m$, $n$ and $r$ denote the dimension of input $x$, state $\boldsymbol{f}$ and the norm, respectively. Such a norm-based saliency fomulation attempts to measure the distance between two high-dimension states. (See more rationality analysis in Appendix \ref{norm}) $\mathcal{S}_{l}^{k}\left(x \right): \mathbb{R}^n\longrightarrow\mathbb{R}$ refers to the saliency of the $k$-th filter in the $l$-th layer taking the sample $x$ as input.

\subsubsection{Joint Saliency of Two Arbitrary Filters}
To perform the global channel pruning, we first consider a simple case: the joint saliency of two arbitrary filters, $\theta_{pu}$ and $\theta_{qv}$, which can be flexibly extended to larger number of filters. According to Eq. (\ref{sensitivity}), the joint filter saliency can be defined as:
\begin{equation}\label{joint_sensitivity}
\begin{aligned}
    \mathcal{S}\left(x;\theta_{pu},\theta_{qv} \right)&=\lVert \boldsymbol{f}(x; \theta _{pu}, \theta_{qv}) -\boldsymbol{f}( x;\theta _{pu}, \theta_{qv}=0) \rVert _r\\
     &\triangleq \mathcal{S}_{pq}^{uv}\left(x \right),
\end{aligned}
\end{equation}
where $\theta_{pu}, \theta_{qv}=0$ means both filters are pruned. For simplicity, we neglect $r$ in the following derivation. However, the conclusion applies for all values of $r\in \mathcal{N}^\star$. Now we have the following theorem about the joint filter saliency.

\newtheorem{thm}{\bf Theorem}
\begin{thm}\label{thm1}
Suppose $\mathcal{S}_{p}^{u}$ and $\mathcal{S}_{pq}^{uv}\left(x \right)$ satisfy Eq.  (\ref{sensitivity}) and Eq. (\ref{joint_sensitivity}), respectively. Similarly, the conditional saliency $\mathcal{S}_{q}^{v}|_{p}^{u}\left( x \right)$ is defined as the derived saliency of filter $\theta_{qv}$ after pruning filter $\theta_{pu}$. Then the following in-equation holds:
\begin{equation}\label{conditional_sensitivity}
    \mathcal{S}_{pq}^{uv}\left( x \right) \leqslant \mathcal{S}_{p}^{u}\left( x \right) + \mathcal{S}_{q}^{v}|_{p}^{u}\left( x \right). 
\end{equation}
The equality sign of Eq. (\ref{conditional_sensitivity}) holds if and only if the designed state $\boldsymbol{f}$ satisfies:
\begin{equation}\label{equality_sign}
\scalebox{0.86}{
$\begin{aligned}
\boldsymbol{f}\in\varPhi _f=\{ \boldsymbol{f}&|\boldsymbol{f}\left( x;\theta _{pu}=0 \right)\\
=&\mathrm{mid}\{ \boldsymbol{f}\left( x;\theta _{pu}=1 \right) , \boldsymbol{f}\left( x;\theta _{pu}=0 \right) ,\\
&\boldsymbol{f}\left( x;\theta _{pu},\theta _{qv}=0 \right)\},\forall x\in D_x\},
\end{aligned}$}
\end{equation}
where mid(·) means the element-wise median map.
\end{thm}

The proof is detailed in Appendix \ref{proof}. According to Theorem \ref{thm1}, the upper bound of the joint saliency is determined by two parts: the independent saliency of $\theta_{pu}$ and the conditional saliency of $\theta_{qv}$. Inspired by the typical classification task where the \textit{softmax} operation is employed to map the output logits into a probabilistic space, we further explore mapping the original saliency value to a new probability space where the probability denotes the filter pruning chance. Let $\mathcal{P}_{pq}^{uv}\left( x \right)$ denote a negative exponential map of $\mathcal{S}_{pq}^{uv}\left( x \right)$, \textit{i.e.}, $\mathcal{P}_{pq}^{uv}\left( x \right)\triangleq\text{exp}\left[-\mathcal{S}_{pq}^{uv}\left( x \right)\right]$. Similarly, $\mathcal{P}_{p}^{u}\left( x \right)\triangleq\text{exp}\left[-\mathcal{S}_{p}^{u}\left( x \right)\right]$ as well as $\mathcal{P}_{q}^{v}|_{p}^{u}\left( x \right)\triangleq\text{exp}\left[-\mathcal{S}_{q}^{v}|_{p}^{u}\left( x \right)\right]$. Then Eq. (\ref{conditional_sensitivity}) is transformed into the following formulation:
\begin{equation}\label{joint_probability}
    \mathcal{P}_{pq}^{uv}\left( x \right) \geqslant \mathcal{P}_{p}^{u}\left( x \right) \mathcal{P}_{q}^{v}|_{p}^{u}\left( x \right), 
\end{equation}
where for symmetry, $\mathcal{P}_{pq}^{uv}\left( x \right) \geqslant \mathcal{P}_{q}^{v}\left( x \right) \mathcal{P}_{p}^{u}|_{q}^{v}\left( x \right)$.

Two insights can be observed from Eq. (\ref{joint_probability}). First, since $\mathcal{P}\in(0, 1]$ due to the non-negative property of $\mathcal{S}$ (explained more in Appendix \ref{exponential}), channel pruning can be understood from a probabilistic view by designing a proper state function $\boldsymbol{f}$ to satisfy Eq. (\ref{equality_sign}), that is, a pruning probability (chance that a filter is pruned) is introduced for each filter. As a result, the joint pruning of two random filters is also probabilistic.

\label{approximation}
Another insight is that despite its in-equivalence to the probability when Eq. (\ref{equality_sign}) fails, Eq. (\ref{joint_probability}) in fact indicates a lower bound of $\mathcal{P}_{pq}^{uv}$. Therefore, as proved by the Expectation Maximization (EM) algorithm \cite{dempster1977maximum} which iteratively optimizes the lower bound to approach the local maximum (See Appendix \ref{EM}), we can push up the lower bound derived from the multiplication of marginal and conditional `probability', which is easier to obtain than $\mathcal{P}_{pq}^{uv}$, to globally approximate the maximum value of $\mathcal{P}_{pq}^{uv}$.

\subsubsection{General Joint Saliency of Arbitrary Filters}
Next, we further generalize Eq. (\ref{conditional_sensitivity}) and Eq. (\ref{joint_probability}) to the case with an arbitrary number of filters, for example, $N$ filters. The formulas are listed as follows:
\begin{equation}\label{multi_sensitivity}
\begin{array}{lll}
& \mathcal{S}_{l_1l_2...l_N}^{k_1k_2...k_N}\left( x \right) \leqslant \mathcal{S}_{l_1}^{k_1} \left( x \right) +\sum_{i=2}^N{\mathcal{S}_{l_i}^{k_i}|_{l_1...l_{i-1}}^{k_1...k_{i-1}}\left( x \right)};\\
& \vspace{1ex} \\
& \mathcal{P}_{l_1l_2...l_N}^{k_1k_2...k_N}\left( x \right) \geqslant \mathcal{P}_{l_1}^{k_1} \left( x \right) \prod_{i=2}^N{\mathcal{P}_{l_i}^{k_i}|_{l_1...l_{i-1}}^{k_1...k_{i-1}}\left( x \right)},
\end{array}
\end{equation}
where $l_i \in \left\{1,2,..., L\right\}, k_i \in $\{$1,2,..., K_{l_i}\}, K_{l_i}$ is the filter number of the $l_i$-th layer. This can be proved by employing the in-equation property of absolute value.

\subsubsection{Step-wise Optimization Objective}
Substituting the upper bound of Eq. (\ref{multi_sensitivity}) into Eq. (\ref{pruning model}), we obtain the following objective:
\begin{equation}\label{objective}
\begin{aligned}
     \underset{\boldsymbol{\theta}}{\min}\,\, & \mathbb{E}_{x\sim D_x}\left[ \mathcal{S}_{l_1}^{k_1} \left( x \right) +\sum_{i=2}^N{\mathcal{S}_{l_i}^{k_i}|_{l_1...l_{i-1}}^{k_1...k_{i-1}}\left( x \right)} \right],\\
     & \text{s.t.}\,\,\,\, \boldsymbol{g}\left( \boldsymbol{\theta } \right) \leqslant \boldsymbol{\alpha }.
\end{aligned}
\end{equation}

That is, the objective of global channel pruning that minimizes the joint saliency, is approximated by a combinatorial objective including multiple conditional saliency items subject to the same compression requirement. Since Eq. (\ref{objective}) comprises sums of conditional saliency items, its optimization can be naturally split into a multiple-step procedure associated with local constraints in each step.

\subsection{Solution: Sequential Channel Pruning Strategy}\label{solution}
Specifically, we take filters in the same layer as a candidate pruning set in one optimization step, fix the filters in other layers and decompose the global constraints of $\boldsymbol{g}\left(\boldsymbol {\theta}\right)$ into the local ones, which thus decomposes the objective Eq. (\ref{objective}) into a concatenation of layer-wise channel pruning problems.

To solve the step-wise optimization problem, we propose to greedily find the candidate pruning layer in each step and the candidate pruning filters in each pruning layer, which is named as a sequentially greedy pruning strategy. In this strategy, the relations between filters for the compression performance can be divided into two types: inter- and intra-layer filter relations, which are both explicitly denoted by the saliency criterion and the compression constraints. Note that such relations change with the pruning state of every filter, thus being dynamic during the pruning process, which is different from previous methods treating the filters' relations statically and screening out all candidate filters simultaneously. The computation of filter saliency is detailed in Section \ref{sga}.

Moreover, since filters of each layer are supposed to have different sensitivities to different tasks, the decomposition of global constraints fully considers each layer's contribution to each task's performance, and the compression potential in each pruning layer. Generally, the layers with more contribution to the FLOPs reduction will be assigned with a larger tolerance in the relative performance drop constraint. The constraint of the performance drop depends on which task the layer is most sensitive to, and is denoted as performance-aware oracle criterion. The detailed setting of the criterion is shown in Section \ref{task-aware criterion}.

\subsubsection{State Function and Saliency Criterion}
\label{sga}
As mentioned above, we decompose the objective into the concatenation of layer-wise pruning steps to approximate the optimal GCP. Let $\boldsymbol{\theta}_l$ denote the candidate pruning set in the $l$-th layer. For the pruning step in the $l$-th layer, the local objective can be formulated as the following expression:
\begin{equation}
\begin{aligned}
     \underset{\boldsymbol{\theta}_l}{\min}\,\, & \mathbb{E}_{x\sim D_x}\left[ \mathcal{S}_{l}^{k_1}\left( x;\boldsymbol{\theta} \right) + \sum_{i=2}^{K_l}{\mathcal{S}_{l}^{k_i}|_{l}^{k_1...k_{i-1}}\left( x;\boldsymbol{\theta} \right)} \right],\\
     &\text{s.t.}\,\,\,\, \boldsymbol{g}\left( \boldsymbol{\theta}_l \right) \leqslant \boldsymbol{\alpha}_l,
\end{aligned}
\end{equation}
where $\mathcal{S}_l^{k_i}$ denotes the saliency of the $k_i$-th filter in the $l$-th layer. $K_l$ is the filter number of the candidate set in the $l$-th layer. The $\boldsymbol{\alpha}_l$ is the local constraint derived from the global constraint, which is detailed in Section \ref{task-aware criterion}.

\label{independence_analysis}
Therefore, from the above derivation, the objective in Eq. (\ref{conditional_sensitivity}) can be optimized by combining both the intra-layer and inter-layer pruning in a greedy mode. Generally, the greedy pruning inside a single layer or between layers should be a serial screening process; however, due to the large screening space in most layers, the computational complexity in each layer becomes the main bottleneck during pruning, which reaches $\mathcal{O}({K_l}^2)$. To tackle this, we can design a saliency criterion satisfying that the saliency of filters in the same layer is independent of each other (meaning the saliency score of a filter is irrelative to other filters in the same layer. See Appendix \ref{independence} for more independence analysis) and the saliency of those in different layers keeps dependent on each other, so that the computational complexity in each layer is reduced to $\mathcal{O}(K_l)$. 

Specially, the state function and the saliency criterion can be reasonably designed as the form of $\ell_1$ norm as follows:
\begin{equation}\label{state_function}
\begin{aligned}
     f(x;\theta_{l_{1}k_{1}},..., \theta_{l_{N}k_{N}})
     &=\sum_{i=1}^{N}{\theta_{l_{i}k_{i}}}\lVert \boldsymbol{\omega}_{l_{i}k_{i}}(\theta_{l_{1}k_{1}}, ..., \theta_{l_{N}k_{N}}) \rVert_1; \\
     \mathcal{S}_{l_1,..., l_N}^{k_1,..., k_N}(x) &= \sum_{i=1}^{N}\lVert \boldsymbol{\omega}_{l_{i}k_{i}}(\theta_{l_{1}k_{1}}, ..., \theta_{l_{N}k_{N}}) \rVert_1,
\end{aligned}
\end{equation}
where $\boldsymbol{\omega}_{l_{i}k_{i}}(\theta_{l_{1}k_{1}}, ..., \theta_{l_{N}k_{N}})$ denotes the weights of the $k_i$-th filter in the $l_i$-th layer at the state of $(\theta_{l_{1}k_{1}}, ..., \theta_{l_{N}k_{N}})$. The state function defined in Eq. (\ref{state_function}) satisfies the equality constraint in Eq. (\ref{equality_sign}), such that minimizing the joint saliency of filters is equivalent to maximizing their joint pruning probability. When the candidate filters are in the same layer, the conditional saliency of each filter is equal to its marginal saliency due to the independence of filter weights. Consequently, the local objective of pruning each layer is transformed into the following objective ($x$ is omitted): 
\begin{equation}\label{final_objective}
     \underset{\boldsymbol{\theta}_l}{\min}\,\,\sum_{i=1}^{K_l}{\mathcal{S}_{l}^{k_i}\left( \boldsymbol{\theta} \right)}\,\,\,
     \text{s.t.}\,\boldsymbol{g}\left( \boldsymbol{\theta}_l  \right) \leqslant \boldsymbol{\alpha}_l.
\end{equation}
\subsubsection{Performance-aware Oracle Criterion}\label{task-aware criterion}
Since the objective function optimization is decomposed into the layer-wise pruning steps, we should also decompose the constraint into layer-wise steps to align with the objective function decomposition. As mentioned in Section \ref{preliminary}, the performance drop is added into the constraint to compensate for the discarding of performance loss item in the objective. To decompose the constraint of total performance drop into layer-wise pruning steps, we consider both the performance loss decomposition and the threshold decomposition.

\begin{algorithm}[t]
\caption{The step-wise optimization of the whole pruning process}
\label{algorithm_2}
\LinesNumbered 
\KwIn{Target layer number, $L$; The initial pruning state $\boldsymbol{\theta}$; The list of local performance drop thresholds $\{$$d_i$\}; The maximum number of reserved filters $\eta$.}
\KwOut{The optimized pruning state $\boldsymbol{\theta}$.}
\While {$\lVert \boldsymbol{\theta } \rVert _0 > \eta$}
{
    {  
        \For{$l$ = 1:$L$}
        {
            $\boldsymbol{\theta}_l^{\star} = \underset{\boldsymbol{\theta}_l}{\text{argmin}}\sum_{k=1}^{K_l}{\mathcal{S}_{l}^{k}\left(\boldsymbol{\theta} \right)}, \text{s.t.\,\,} g(\boldsymbol{\theta})=d_l$\;
            update $\boldsymbol{\theta}$ with $\boldsymbol{\theta}_l^{\star}$\;
        }
    }
}
\textbf{return} $\boldsymbol{\theta }$\;
\end{algorithm}  

For the decomposition of performance loss, we argue that features extracted by different filters present various importance to each task across all layers. Taking object detection as an example, most filters in shallow layers focus on extracting more detailed and spatial information that benefits the localization task. Thereby, shallow layers are more robust to the localization task and more sensitive to the classification task. On the contrary, most filters in deep layers are robust to the classification task and sensitive to the localization task due to rich semantic information extracted by filters. Therefore, the local performance drop constraint of each layer is supposed to pay attention to the performance drop of the most sensitive task. In other words, when the performance drop of the most sensitive task reaches the local constraint boundary, the number of prunable filters in a layer will reach a maximum, which is namely the performance-aware oracle criterion. Specially, the formula of the criterion $g(\cdot)$ is the $\ell_{\infty}$ norm of the multitask loss changes:
\begin{equation}\label{task-aware constrain}
\begin{aligned}
    g(\boldsymbol{\theta}) &=\lVert \varDelta\mathcal{L} _1( \boldsymbol{\theta}) , \varDelta\mathcal{L} _2( \boldsymbol{\theta}) , ..., \varDelta\mathcal{L} _T( \boldsymbol{\theta}) \rVert _{\infty}\\
    &=\underset{1\leqslant t\leqslant T}{\max}| \varDelta\mathcal{L} _t(\boldsymbol{\theta})|,
\end{aligned}
\end{equation}
where $\varDelta\mathcal{L}_t(\boldsymbol{\theta})$ denotes the relative loss change of the $t$-th task at the pruning state $\boldsymbol{\theta}$. In this way, the criterion is insensitive to the scaling factor and the absolute loss changes. $T$ is the task number of the model. Note that the task number here refers to the number of loss components in the learning objective. Therefore, some models designed for two tasks, \textit{e.g.}, semantic segmentation and depth estimation, may contain more than two loss components, \textit{e.g.}, losses for multi-scale predictions for each task. \footnote{For example, in MTI-Net, there are two prediction stages. In the initial (first) stage, MTI-Net outputs multi-scale predictions (4 scales for HRNet-w18) of each task as initial predictions. In the final (second) stage, MTI-Net predicts the final output of each task. Taking the training setting in NYUD-v2 as an example, where the semantic segmentation map and depth estimation map are to be predicted, the actual task number in our criterion is 2 (seg. and dep.)×5 (4 initial stage scales prediction and 1 final stage prediction) = 10.}

It can be observed that the global constraint $g(\cdot)$ is in the same form as Eq. (\ref{sensitivity}), where the vector of $\mathcal{L}_t$ corresponds to $\boldsymbol{f}$; therefore, it can also be approximated by a combinatorial constraint, including the independent constraints for the first pruning layer and multiple conditional constraints (meaning the performance drop constraint on condition of the previous pruning state).

For the decomposition of the local threshold, we consider that the redundancy of each target layer is different and thus propose to decompose the global threshold of the performance drop into the progressive drop constraints as follows:
\begin{equation}
\label{scaling_factor}
    \prod_{i=1}^{L}\left(1+d_1\lambda^{i-1}\right) = \alpha,
\end{equation}
where $d_1$ denotes the initial drop threshold for the first pruning layer candidate (meaning the maximum acceptable relative performance drop of the first pruning layer is $d_1$), and $\lambda$ denotes the constant scaling factor of each step's threshold based on the previous drop threshold. $\alpha$ is the threshold of the global performance drop constraint. As a result, the drop threshold of the constraint at the $i$-th pruning step, $d_i$ can be computed by scaling $d_{i-1}$ with $\lambda$.

\begin{algorithm}[t]
\caption{The proposed PAGCP framework}
\label{algorithm_1}
\LinesNumbered 
\KwIn{ 
  The pre-trained model, $M_0$; The dataset, $D$; The reserved ratio of FLOPs or parameters, $\Gamma$; The performance drop threshold, $\alpha$; The initial drop threshold $d_1$; The masking ratio of filters, $\gamma$; The filtering ratio of pruning layers, $P$.}
\KwOut{
  The compressed model, $M_{p}$ satisfying the compression requirements $\Gamma$.}
initialize $p\longleftarrow 1$, $M_p\longleftarrow M_0$\;
\While {$p > \Gamma$}
{
    reorder the pruning layer sequence according to the FLOPs reduction at a certain pruning ratio of each single layer$\longrightarrow l_F$\;
    compute $\lambda$ by Eq. (\ref{scaling_factor})\;
    \For{$i, l$ in enumerate($l_F$)}
    {  
        \For{$j$ = 1:$K_i$}
        {
            compute $\mathcal{S}_l^{j}$ by Eq. (\ref{state_function})\;
        }
        mask $\gamma$ filters with lowest value in $\mathcal{S}_l$ and find the most sensitive task $t_l$ by Eq. (\ref{task-aware constrain})\;
        prune filters with lowest value in $\mathcal{S}_l$ until the relative performance drop of Task $t_l$ oversteps $d_i$\; 
        $R_l\longleftarrow$ ratio of pruned filters in layer $l$\;
        $M_p\longleftarrow$ pruned model\;
        $d_{i+1}\longleftarrow \lambda d_i$\;
    }
    sort the compression contribution $\boldsymbol{C}$ of each layer at the ratio of $\boldsymbol{R}=\{R_1, ..., R_{L}\}$\, and 
    select top $P$ layers as the pruning layer set$\longrightarrow L_p$\;
    reset model $M_p\longleftarrow M_0$\;
    \For {l in $l_F$}
    {
        \If {l in $L_p$}
        {
            prune filters of $M_p$ in $\mathcal{S}_l$ with the ratio of $R_l\longrightarrow M_p$ \;
        }
    }
    $M_0 \longleftarrow \text{fine-tuned }M_p$\;
    update $p$\;
}
\textbf{return} $M_p$\;
\end{algorithm}  

Therefore, the final step-wise objective optimization of pruning process is illustrated in Alg. (\ref{algorithm_2}). Note that in order to achieve high pruning efficiency per iteration, we impose looser local constraints for those layers with more contributions to the FLOPs reduction. Specifically, we first evaluate the degree of FLOPs reduction of all candidate layers individually with the same pruning ratio, and sort these layers in descending order of the FLOPs reduction if $\lambda < 1$ else in ascending order (See the sequences difference with different pruning ratios in Appendix \ref{pruning_sequence}). Then for each step, we pop out one layer to prune from the reordered queue. Considering the trade-off between model lightweight and prediction performance, we further evaluate the contribution of each filter to be pruned to the total FLOPs reduction, and remove those with little compression contribution but certain performance drop. In experiments, we denote the variable $P$ to adjust the filtering ratio of pruning layers. The framework of the pruning strategy is shown in Alg. (\ref{algorithm_1}).

\section{Experiments and Results}\label{experiment}
The proposed method is empirically evaluated on two types of multitask learning: object detection, which is an important vision task, and the conventional multitask learning. For object detection, we conduct the compression on three benchmark models (SSD\cite{liu2016ssd}, Faster R-CNN\cite{Ren2015Faster} and CenterNet\cite{zhou2019objects}) and a more recent lightweight detector, namely YOLOv5m\footnote{YOLOv5m code: https://github.com/ultralytics/yolov5/tree/v5.0} \cite{glenn2021yolov5}. The benchmark datasets of object detection are PASCAL VOC \cite{everingham2015pascal} and COCO \cite{lin2014microsoft}. For the conventional multitask learning, we select to compress the state-of-the-art model MTI-Net \cite{vandenhende2020mti} on two benchmarks: NYUD-v2 \cite{silberman2012indoor} and PASCAL Context \cite{context}. For sufficient comparisons between our method and other channel pruning methods for multitask model compression, besides finetuning-based methods, we apply several state-of-the-art pruning methods on the multitask model pruning that are experimented on classification models in official papers but relatively easy to transfer to multitask model pruning, namely the adapted-based methods. Further, considering that the study of channel pruning methods for multitask model compression is still under-explored, we also compare the compressed models with lightweight models obtained by other compression techniques like knowledge distillation, Neural Architecture Search (NAS) and model scaling, to validate the superiority of our method.

We first introduce the benchmark datasets in our work. Then, we give the detailed experimental setups, including the pruning scheme of skip connections, the training setup and the compression setup. Finally, we show the compression results of \textbf{five typical models}.

\subsection{Dataset}
\textbf{PASCAL VOC} \cite{everingham2015pascal} contains 20 categories of real-world objects, including 5,011 \textit{trainval} images for PASCAL VOC 2007, 11,540 \textit{trainval} images for PASCAL VOC 2012, and 4,952 \textit{test} images, respectively. We follow the official split setting to train models on both \textit{trainval} lists, and evaluate the proposed method on PASCAL VOC 2007 \textit{test} set. 

\noindent\textbf{COCO2017} \cite{lin2014microsoft} is the latest version of COCO dataset, containing 118K training images, 5,000 validation images and 41K test images. We train models on training images and evaluate them on the validation set.

\noindent\textbf{NYUD-v2}\cite{silberman2012indoor} is the indoor scene dataset, which has multiple task labels, including semantic segmentation and depth estimation. The dataset contains 1,449 images, with 795 \textit{training} images and 654 \textit{test} images.

\noindent\textbf{PASCAL Context}\cite{context} is a larger benchmark dataset for multitask learning, which contains 10,103 images, with 4,998 for training and 5,105 for test. Multiple tasks are available in this dataset, including semantic segmentation, depth estimation, saliency estimation, and surface normal estimation, etc.

The details of task choices for NYUD-v2 and PASCAL Context are shown in Table \ref{task_1}.

\begin{table}[htbp]
\centering
\caption{\label{task_1} Tasks choices for NUYD-v2 and PASCAL Context, where Seg., Dep., Sal., and Norm., refer to semantic segmentation, depth estimation, saliency estimation, and surface normal estimation, respectively.}
\resizebox{\columnwidth}{!}{
\begin{tabular}{ccccc}
\toprule
\multirow{2}{*}{} & Seg. & Dep. & Sal. & Norm. \\\hline
NYUD-v2 & \checkmark & \checkmark & &   \\
PASCAL Context & \checkmark &  &\checkmark&\checkmark  \\\hline
Metrics & mIoU (\%) & rmse & mIoU (\%) & mErr \\\bottomrule
\end{tabular}}
\end{table}

\subsection{Experimental Setup}
\subsubsection{Pruning of Skip Connections}
\begin{figure}[t]
    \centering
    \scalebox{0.5}{
    \includegraphics{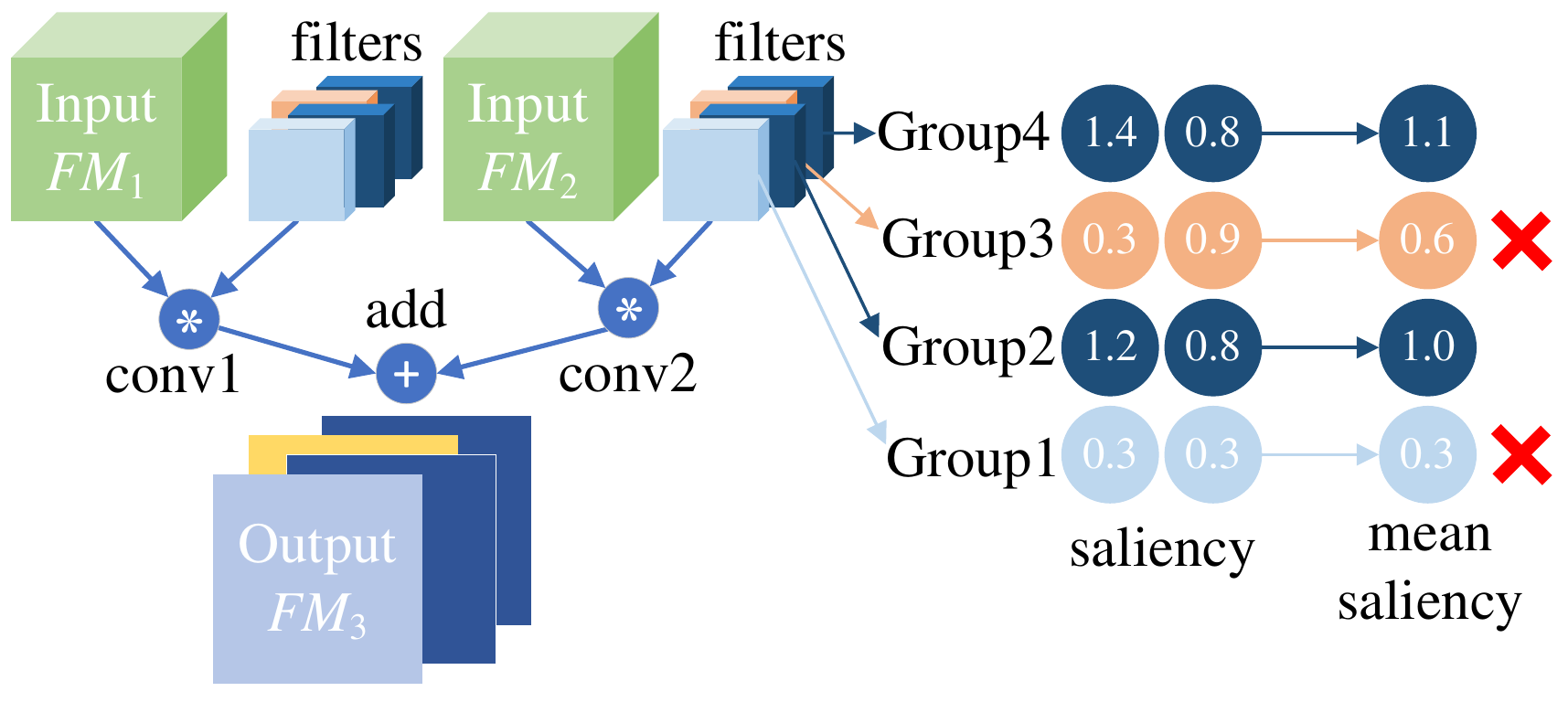}}
    \caption{The pruning scheme of skip connections in our method. Filters with the same index across different layers share the same group. Input $FM_1$ and $FM_2$ generated by different layers denote the general input feature maps of the skip connection layer. Filters of Group 1 and Group 3 are finally removed in this figure, as well as the corresponding channels of output feature maps $FM_3$.
    }
    \label{skipconnection}
\end{figure}

Traditional channel pruning for CNNs usually considers the pruning of only convolutional layers without pruning the skip connection layers. In our method, to maximize the pruning ratio of a model, we take \textbf{the skip connection module} into consideration. Similarly, we evaluate the importance of skip connection based on Eq. (\ref{state_function}). To consider the joint impact of multiple filters on one channel in the skip connection module, we compute the importance of a channel in the skip connection module by the group mean of all corresponding filter saliency scores. Fig. \ref{skipconnection} illustrates the pruning scheme of skip connections.

\begin{table*}[t]
\centering
\caption{Compression results on PASCAL VOC. \#Params denotes the parameter numbers. ``$\clubsuit$'' denotes the model is trained on VOC07 \textit{trainval} dataset. ↓ denotes the absolute decrease of the metric, and ↓(\%) denotes the relative decrease (\%) of the metric.}
\label{pascal}
\resizebox{\textwidth}{!}{
\begin{tabular}{@{}lllcccccc@{}}
\toprule
& \multicolumn{1}{l}{Method} & Model & mAP (\%) & FLOPs (B) & \#Params (M) & mAP↓ & FLOPs↓(\%) & \#Params↓(\%) \\ \midrule
\multirow{12}{*}{SSD300} & \multicolumn{1}{l}{Baseline} & VGG16-SSD \cite{liu2016ssd} & 77.5 & 31.5 & 26.3 & - & - & - \\ \cmidrule(l){2-9}
& \multirow{3}{*}{Lightweight} & Mobile-SSD \cite{howard2017mobilenets} & 70.0 & 2.6 & 8.8 & - & - & - \\
&  & DSOD300-small \cite{shen2017dsod} & 73.6 & 5.3 & 5.9 & - & - & - \\
& & SSDM\_7.5 \cite{8078500} & 73.1 & 7.5 & 10.1 & - & - & - \\ \cmidrule(l){2-9}
& \multirow{2}{*}{Finetuning} & Res10-SSD \cite{8078500} & 64.8 & 2.3 & 6.7 & - & 92.7 & 74.5 \\
& & VGG11\_0.5-SSD \cite{liu2018rethinking} & 68.1 & 15.4 & 20.4 & 9.4 & 51.1 & 22.4 \\ \cmidrule(l){2-9}
& \multirow{3}{*}{Adapting} & ThiNet \cite{8416559} & 72.7 & 7.9 & 6.6 & 4.8 & 74.9 & 74.9 \\
& & MLP \cite{Singh_2019} & 75.0 & 13.5 & 3.9 & 2.5 & 57.1 & 85.2 \\
& & LCP \cite{XIE2020400} & 75.2 & 7.9 & 6.6 & 2.3 & 74.9 & 74.9 \\ \cmidrule(l){2-9}
& \multicolumn{1}{l}{} & Ours & \textbf{75.6} & \textbf{8.1} & \textbf{3.8} & \textbf{1.9} & \textbf{74.3} & \textbf{85.6} \\ \midrule
\multirow{7}{*}{Faster R-CNN} & Baseline & VGG16-Faster \cite{Ren2015Faster} & 75.7 & 400.2 & 137.1 & - & - & - \\ \cmidrule(l){2-9} 
 & Finetuning & VGG11\_0.5-Faster \cite{liu2018rethinking} & 57.6 & 292.4 & 129.0 & 18.1 & 26.9 & 5.9 \\ \cmidrule(l){2-9} 
 & \multirow{2}{*}{Distillation} & VGG11-I Faster \cite{Wang_2019}$^\clubsuit$ & 67.6 & 156.4 & 133.9 & - & 60.9 & 2.3 \\
 &  & Res50-I Faster \cite{Wang_2019}$^\clubsuit$ & 72.0 & 50.62 & 48.3 & - & 87.4 & 64.8 \\ \cmidrule(l){2-9} 
 &  & Ours & \textbf{73.8} & \textbf{100.9} & \textbf{37.9} & \textbf{1.9} & \textbf{74.8} & \textbf{72.4} \\ \midrule
\multirow{12}{*}{CenterNet} & Baseline & Res50-CenterNet & 76.9 & 26.9 & 30.7 & - & - & - \\ \cmidrule(l){2-9}
& \multirow{2}{*}{Finetuning} & Res50\_0.75-CenterNet \cite{li2020eagleeye} & 76.3 & 21.6 & 24.8 & 0.6 & 19.7 & 19.2 \\
& & Res50\_0.5-CenterNet \cite{li2020eagleeye} & 75.3 & 16.2 & 19.4 & 1.6 & 39.8 & 36.8 \\ \cmidrule(l){2-9}
 & \multirow{2}{*}{} & \multirow{2}{*}{Ours} & \textbf{76.9} & \textbf{20.3} & \textbf{23.3} & \textbf{0.0} & \textbf{24.5} & \textbf{24.1} \\ 
 & &  & \textbf{76.4} & \textbf{17.9} & \textbf{19.9} & \textbf{0.5} & \textbf{33.5} & \textbf{35.2} \\ \cmidrule(l){2-9}
 & Baseline & Res101-CenterNet & 76.6 & 46.3 & 49.7 & - & - & - \\ \cmidrule(l){2-9}
 & Finetuning & Res101\_0.7-CenterNet \cite{sfp} & 76.6 & 30.5 & 33.4 & 0.0 & 34.1 & 32.8 \\ \cmidrule(l){2-9}
 & \multirow{2}{*}{} & \multirow{2}{*}{Ours} & \textbf{77.2} & \textbf{30.6} & \textbf{31.0} & \textbf{-0.6} & \textbf{33.9} & \textbf{37.6} \\
 & &  & \textbf{76.1} & \textbf{23.0} & \textbf{22.9} & \textbf{0.5} & \textbf{50.3} & \textbf{55.7}\\ \midrule
\multirow{6}{*}{YOLOv5} & Baseline & YOLOv5m \cite{glenn2021yolov5} & 83.2 & 32.4 & 21.4 & - & - & - \\ \cmidrule(l){2-9}  
 & Scaling & YOLOv5s \cite{glenn2021yolov5} & 79.5 & 10.2 & 7.3 & 3.7 & 68.5 & 65.9 \\ \cmidrule(l){2-9}
 & \multirow{2}{*}{Adapting} & NS-YOLOv5m \cite{liu2017learning} & 81.9 & 19.5 & 14.2 & 1.3 & 39.8 & 33.6\\
 & & Eagleye-YOLOv5m \cite{li2020eagleeye} & 76.4 & 11.4 & 8.9 & 6.8 & 64.8 & 58.4\\\cmidrule(l){2-9}
 &  & Ours & \textbf{81.7} & \textbf{11.4} & \textbf{4.7} & \textbf{1.5} & \textbf{64.8} & \textbf{78.0} \\ \bottomrule
\end{tabular}}
\end{table*}

\begin{table*}[t]
\small
\centering
\caption{Compression results on COCO. ``$\dagger$'' denotes the model input size is 300$\times$300 pixels.}
\label{coco}
\resizebox{\textwidth}{!}{
\begin{tabular}{@{}lllcccccc@{}}
\toprule
& \multirow{2}{*}{Method} & \multirow{2}{*}{Model} & AP (\%) & mAP (\%) & FLOPs  & \#Params & FLOPs & \#Params \\
& & & @50 & @(0.5:0.95) & (B) & (M) & ↓(\%) & ↓(\%) \\\midrule 
\multirow{5}{*}{SSD512} & \multicolumn{1}{l}{Baseline} & VGG16-SSD \cite{liu2016ssd} & 48.3 & 29.4 & 98.7 & 36.0 & - & - \\ \cmidrule(l){2-9}
& Finetuning & VGG11\_0.5-SSD \cite{liu2018rethinking} & 31.5 & 18.0 & 53.3 & 30.1 & 46.0 & 16.4 \\ \cmidrule(l){2-9}
& Adapting & MLP \cite{Singh_2019}$^\dagger$ & - & 21.7 & - & 12.7 & - & 64.7 \\\cmidrule(l){2-9}
& & Ours & \textbf{47.6} & \textbf{28.6} & \textbf{55.9} & \textbf{21.1} & \textbf{43.4} & \textbf{41.4} \\ \midrule
\multirow{9}{*}{CenterNet} & Baseline & Res50-CenterNet \cite{zhou2019objects} & 54.1 & 34.8 & 26.9 & 30.7 & -  & - \\ \cmidrule(l){2-9}
& \multirow{2}{*}{Finetuning} & Res50\_0.75-CenterNet \cite{li2020eagleeye} & 54.2 & 35.0 & 21.6 & 24.8 & 19.7 & 19.2  \\
& & Res50\_0.5-CenterNet \cite{li2020eagleeye} & 52.7 & 34.0 & 16.3 & 19.4 & 39.4 & 36.8 \\ \cmidrule(l){2-9}
 &  & Ours & \textbf{54.3} & \textbf{35.3} & \textbf{21.0} & \textbf{17.8} & \textbf{21.9} & \textbf{42.0} \\ \cmidrule(l){2-9}
 & Baseline & Res101-CenterNet\cite{zhou2019objects} & 52.6 & 34.1 & 46.4 & 49.7 & - & - \\ \cmidrule(l){2-9}
 & Finetuning & Res101\_0.7-CenterNet \cite{sfp} & 51.9 & 33.4 & 30.5 & 33.4 & 34.3 & 32.8 \\ \cmidrule(l){2-9}
 & & Ours & \textbf{52.3} & \textbf{34.0} & \textbf{33.6} & \textbf{28.6} & \textbf{27.6} & \textbf{42.5} \\ \midrule
 \multirow{12}{*}{YOLOv5} & Baseline & YOLOv5m \cite{glenn2021yolov5} & 62.7 & 43.6 & 51.3 & 21.4 & - & - \\ \cmidrule(l){2-9}
  & \multirow{5}{*}{Lightweight} & FairDARTS-C \cite{chu2020fair} & 51.9 & 31.9 & - & 5.3 & - & - \\
  & & DARTS-A \cite{chu2020darts} & 52.8 & 32.5 & - & 5.5 & - & - \\
  & & YOLOv4-Tiny \cite{wang2021scaled} & 40.2 & 21.7 & 7.0 & 6.1 & - & - \\
  & & EfficientDet-D1 \cite{tan2020efficientdet} & 58.8 & 40.2 & 6.1 & 6.6 & - & - \\
  & & YOLOX-S \cite{yolox2021} & 58.7 & 39.6 & 26.8 & 9.0 & - & - \\ \cmidrule(l){2-9} 
 & Scaling & YOLOv5s \cite{glenn2021yolov5} & 56.5 & 36.7 & 17.0 & 7.3 & 66.9 & 65.9 \\ \cmidrule(l){2-9}
 & \multirow{2}{*}{Adapting} & NS-YOLOv5m \cite{liu2017learning} & 58.5 & 40.2 & 27.9 & 15.0 & 45.6 & 29.9\\
 & & Eagleye-YOLOv5m \cite{li2020eagleeye} & 59.2 & 40.6 & 25.7 & 9.7 & 49.9 & 54.7\\\cmidrule(l){2-9}
 & \multirow{2}{*}{} & \multirow{2}{*}{Ours} & \textbf{60.7} & \textbf{41.5} & \textbf{23.5} & \textbf{7.7} & \textbf{54.2} & \textbf{64.0} \\
 & &  & \textbf{60.3} & \textbf{41.0} & \textbf{20.0} & \textbf{6.2} & \textbf{61.0} & \textbf{71.0}\\ \bottomrule
\end{tabular}}
\end{table*}

\subsubsection{Training Setup}
\textbf{Object Detection.} We evaluate the proposed PAGCP on three benchmark object detectors: VGG16-based SSD \cite{liu2016ssd}, VGG16-based Faster R-CNN \cite{Ren2015Faster} and ResNet-based CenterNet \cite{zhou2019objects}. We also apply the method on a recent lightweight detector, namely YOLOv5m\cite{glenn2021yolov5}.

In the experiments of compressing SSD, the input images are resized to 300$\times$300 pixels and 512$\times$512 pixels on PASCAL VOC and COCO2017 datasets, respectively. On PASCAL VOC, we train the base model of SSD300 with batch size of 32. The common learning rate is initialized as 0.001, multiplied by 0.1 at iteration 80,000 and 100,000, and terminated after 120,000 iterations. On COCO, we train the base model of SSD512 with the batch size of 12. The initial learning rate is set as 0.001, and decays by 0.1 at the iteration of 280,000, 360,000. The overall training process reaches 400,000 iterations. For the fine-tuning of the pruned model, we set the number of training epochs proportional to the compression rate of FLOPs, which means that we will fine-tune the model with smaller FLOPs for more epochs. 

For the compression of Faster R-CNN, we conduct experiments on PASCAL VOC. The shorter side of each image is set to 600 pixels. We train the Faster R-CNN with the batch size of 1, and the learning rate is initialized as 0.001, multiplied by 0.1 at epoch 10 and terminated at epoch 14. The fine-tuning setting of the pruned model keeps the same as the base training setting, since the original training epoch number is small. 

For the compression of CenterNet, we conduct experiments on PASCAL VOC and COCO. The image size is set to 512 $\times$ 512 pixels on both datasets. On PASCAL VOC, we train the base model with batch size of 32 and the initial learning rate is set as 1.25e-4, and decays by 0.1 at epoch 45 and 60. The whole training process reaches 70 epochs. On COCO, we train the base model with batch size of 96 and the learning rate is initialized at 3.75e-4, multiplied by 0.1 at epoch 90 and 120, terminated after 140 epochs. The fine-tuning schedule of the pruned model follows the same strategy as the compression of SSD.

For the compression of YOLOv5m, we conduct experiments on PASCAL VOC and COCO. The image size is set to 512$\times$512 pixels on PASCAL VOC and 640$\times$640 pixels on COCO, and we train the base model from scratch with batch size of 64 on both benchmark datasets. The training setting follows the official strategy. The fine-tuning setting follows the same strategy in the compression of SSD.

\noindent\textbf{Conventional Multitask Learning.} We evaluate the PAGCP on the state-of-the-art multitask learning model, MTI-Net\cite{vandenhende2020mti}, which takes HRNet \cite{hrnet} as backbone. The image size is set to 480 $\times$ 640 on NYUD-v2 and 512 $\times$ 512 on PASCAL Context. We train the base model with batch size of 32 and the learning rate is initialized as 1.5e-4, updated in the polynomial form, and terminated after 200 epochs on both benchmark datasets. The fine-tuning setting of the pruned model follows the same strategy as the compression of SSD.

\subsubsection{Compression Setup}
For all models in our experiments, we select all layers except the output heads as target layers. Specially, the dynamic convolution (DCN-v2) layers are reserved in ResNet-based CenterNet since the pruning of offset channels will affect the successive convolution weights. The filtering ratio $P$ is set to $80\%$ of target layers to be pruned. The performance drop is evaluated by the loss function in the base training.

For the compression of SSD, we constrain the global performance drop $\alpha$ as 6 for PASCAL VOC and 5 for COCO in each pruning iteration, while the initial performance drop threshold $d_1$ is set as 6\% on both PASCAL VOC and COCO. For the compression of other models, we set two hyper-parameters as follows: $\alpha=6, d_1=6\%$ on PASCAL VOC for Faster R-CNN, $\alpha=6, d_1=6\%$ on PASCAL VOC and $\alpha=5, d_1=6\%$ on COCO for CenterNet, $\alpha=10, d_1=6\%$ on PASCAL VOC and $\alpha=8, d_1=4\%$ on COCO for YOLOv5m, and $\alpha=10, d_1=6\%$ on both NYUD-v2 and PASCAL Context for MTI-Net.

\subsection{Results of Object Detection}
\subsubsection{Results on PASCAL VOC}
Table \ref{pascal} compares the pruned detectors using PAGCP with other methods under the same test setting. Notably, except for the lightweight models using other compression methods instead of channel pruning and the adapting-based pruned SSD models, whose results are referred from their original papers, we \textbf{reproduce} all other methods under the same finetuning setup.
In general, the compressed models using the PAGCP method achieve higher accuracy and larger reduction in FLOPs and parameters. Specially, on low-accuracy regime, our method achieves the highest compression ratio for SSD300 and the best accuracy among all SSD variants, obtaining at most 2.9\% accuracy increase than adapting-based channel pruning methods \cite{8416559, Singh_2019, XIE2020400} with over 85\% parameter reduction. On high-accuracy regime, the pruned YOLOv5m model based on PAGCP also outperforms both adapting-based pruned models and the recent efficient detector YOLOv5s \cite{glenn2021yolov5}, which is based on scaling down the width and depth of YOLOv5m. In particular, our method achieves 2.2\% higher accuracy than YOLOv5s, and controls the accuracy decrease at 1.5\% with 78\% reduction in parameters and 64.8\% reduction in FLOPs. The compression results of the other two detectors, Faster R-CNN and CenterNet, also validate the consistent effectiveness and versatility of our method in different types of detectors.

\begin{table}[]
\small
\centering
\caption{Compression results of SSD300 on PASCAL VOC at each iteration.}
\label{ssd300_voc_periteration}
\scalebox{1.0}{
\begin{tabular}{@{}lccccc@{}}
\toprule
\multirow{2}{*}{Iter} & mAP & FLOPs & \#Params & FLOPs & \#Params\\ 
& (\%) &  (B) & (M) & ↓(\%) & ↓(\%) \\\midrule
0   & 77.5           & 31.5          & 26.3        & - & - \\
1   & 76.2           & 12.8          & 8.6        & 59.4 & 67.3  \\
2   & 75.9           & 10.6          & 5.7        & 66.3 & 78.3 \\
3   & 75.7           & 9.1           & 4.7        & 71.1 & 82.1  \\
4   & 75.6           & 8.1           & 3.8        & 74.3 & 85.6 \\ \bottomrule[0.8pt]
\end{tabular}
}
\end{table}

In addition, in Table \ref{ssd300_voc_periteration}, we enumerate the pruning results in each iteration of compressing SSD to the comparable size with other methods. In the initial pruning iteration, it can be observed that the reduction of both FLOPs and parameters is significant and reaches over 50\%, while after several iterations, the available compression space gradually decreases. This observation conforms to the prior knowledge that the small, well-trained models may have less redundancy than large, well-trained models.

\begin{table*}[t]
\centering
\caption{Compression results on conventional multitask learning benchmarks. F. and P. denotes FLOPs and parameters, respectively. S2M\_seg, S2M\_dep, S2M\_sal and S2M\_norm denote the compressed MTI-Net models pruned by PAGCP on the semantic segmentation, depth estimation, saliency estimation and surface normal estimation, respectively, and then finetuned on the multitask setting.}
\label{nyud}
\resizebox{\textwidth}{!}{
\begin{tabular}{@{}lllccccccc@{}}
\toprule
\multirow{2}{*}{Dataset} & \multirow{2}{*}{Method} & \multirow{2}{*}{Model} & Seg. & Dep. & FLOPs & \#Params  & F.↓ & P.↓\\ 
& & & (mIoU) & (rmse) & (B) & (M) & (\%) & (\%) \\\midrule 
\multirow{10}{*}{NYUD-v2} & Baseline & HRNet18-MTI\cite{vandenhende2020mti} & 38.5 & 0.592 & 16 & 9 & - & -\\ \cmidrule(l){2-10} 
& \multirow{2}{*}{Lightweight} & MTL\cite{vandenhende2020mti} & 33.9 & 0.636 & 6 & 4 & - & -\\
& & PAD-Net\cite{xu2018pad} & 36.0 & 0.630 & 82 & 12 & - & -\\ \cmidrule(l){2-10}
& \multirow{2}{*}{Finetuning} & S2M\_seg & 37.8 & 0.596 & 10.7 & 7.4 & 33.1 & 17.8 \\
& & S2M\_dep & 34.6 & 0.632 & 10.4 & 6.8 & 35.0 & 24.4\\ \cmidrule(l){2-10}
& \multirow{3}{*}{Adapting} & $\ell_1$ norm \cite{KDSG17} & 36.6 & 0.603 & 8.9 & 6.3 & 44.4 & 30.0 \\
& & NS-MTI \cite{liu2017learning} & 36.8 & 0.605 & 10.7 & 5.9 & 33.1 & 34.4\\
& & Eagleeye-MTI \cite{li2020eagleeye} & 36.0 & 0.612 & 8.6 & 5.1 & 46.3 & 43.3 \\ \cmidrule(l){2-10} 
& \multirow{2}{*}{} & \multirow{2}{*}{Ours} & \textbf{37.9} & \textbf{0.593} & \textbf{9.9} & \textbf{6.7} & \textbf{38.1} & \textbf{25.6}\\
& & & \textbf{37.5} & \textbf{0.598} & \textbf{7.8} & \textbf{5.9} &\textbf{51.3} & \textbf{34.4} \\ \midrule \midrule
 & \multirow{2}{*}{Method} & \multirow{2}{*}{Model} & Seg. & Sal. & Norm. & FLOPs & \#Params  & F.↓ & P.↓\\ 
& & & (mIoU) & (mIoU) & (mErr) & (B) & (M) & (\%) & (\%) \\\midrule
\multirow{10}{*}{PASCAL Context} & Baseline & HRNet18-MTI\cite{vandenhende2020mti} & 61.0 & 66.7 & 14.7 & 22 & 13 &- &-\\ \cmidrule(l){2-10}
& \multirow{2}{*}{Lightweight} & MTL\cite{vandenhende2020mti} & 54.1 & 65.3 & 14.8 & 7 & 4 & - &-\\
& & PAD-Net\cite{xu2018pad} & 49.1 & 66.3 & 14.8 & 168 & 12 & - & -\\ \cmidrule(l){2-10}
& \multirow{3}{*}{Finetuning} & S2M\_seg & 57.5 & 65.9 & 14.7 & 15.4 & 11.0 & 30.0 & 15.4 \\
& & S2M\_sal & 57.1 & 66.9 & 14.6 & 15.3 & 11.1 & 30.5 & 14.6 \\ 
& & S2M\_norm & 54.5 & 65.8 & 14.4 & 14.1 & 11.1 & 35.9 & 14.6 \\ \cmidrule(l){2-10}
& \multirow{3}{*}{Adapting} & $\ell_1$ norm \cite{KDSG17} & 59.1 & 66.2 & 14.4 & 13.1 & 9.8 & 40.5 & 24.6 \\
& & NS-MTI \cite{liu2017learning} & 56.2 & 65.9 & 14.6 & 12.9 & 7.2 & 41.4 & 44.6 \\
& & Eagleeye-MTI \cite{li2020eagleeye} & 57.5 & 65.6 & 14.7 & 11.2 & 8.2 & 49.1 & 36.9 \\ \cmidrule(l){2-10} 
& & Ours & \textbf{59.7} & \textbf{66.3} & \textbf{14.3} & \textbf{11.5} &\textbf{9.7} & \textbf{47.7} & \textbf{25.4}  \\ \bottomrule
\end{tabular}}
\end{table*}

\subsubsection{Results on COCO}
We further conduct the compression experiments of multiple detectors on COCO, whose results are reported in Table \ref{coco}. The experimental scheme and selected comparison methods keep the same as Table \ref{pascal}. Differently, we mainly focus on the high-accuracy regime in the compression on COCO, where we compare the pruned YOLOv5m with the most recent lightweight detectors based on Neural Architecture Search (NAS) \cite{chu2020fair, chu2020darts, tan2020efficientdet}and other state-of-the-art YOLO variants \cite{wang2021scaled, yolox2021}. We report the AP@50 and mAP@(0.5:0.95) in the experiments as the metrics for comparison. From the perspective of lightweight performance, the pruned YOLOv5m using the proposed PAGCP surpasses recent detectors by a large margin (+2\%) in both metrics under the similar parameter magnitude. And compared with YOLOX-S \cite{yolox2021}, which is the most recent efficient YOLO-variant model, the pruned YOLOv5m has fewer FLOPs and parameters, but achieves 1.9\% better mAP(0.5:0.95) than YOLOX-S. From the perspective of compression quality, the pruned YOLOv5m using PAGCAP also outperforms both adapting- and scaling-based models, from which the pruned YOLOv5m exceeds YOLOv5s by more than 4\% in both AP@50 and mAP@(0.5:0.95) under the similar compression ratio. And the compression results of SSD512 and CenterNet further validate the capability of maintaining high performances on large-scale datasets while pruning 40$\sim$70\% FLOPs or parameters.

\subsection{Results of Conventional Multitask Learning}

In general, our pruning method can be adapted to any multitask model. Therefore, we are also interested in the compression of models for conventional multitask learning. For simplicity, we perform the proposed PAGCP on the compression of the state-of-the-art model, MTI-Net \cite{vandenhende2020mti}. To compare with other pruning methods, we adopt several finetuning- and adapting-based methods to compress MTI-Net. For finetuning-based methods, since the backbone of MTI-Net, HRNet, has never been compressed in previous pruning methods, we apply our method on pruning the backbone of the single-task MTI-net, \textit{i.e.}, segmentation-only MTI-Net or depth-only MTI-Net, and then finetune the pruned backbone under the multitask setting. Specifically, we select the final prediction of each task as the single task to prune MTI-Net, and apply our method on the single-task pruning. The compression setup of single-task pruning keeps the same as our method, \textit{i.e.}, $d_1$=0.06, $\alpha$=10, $P$=0.8. As for the adapting-based methods, we pick the same pruning methods as detectors’ ones, \textit{i.e.}, Network Slimming \cite{liu2017learning} and Eagleeye \cite{li2020eagleeye}, and adapt them to prune MTI-Net. The compression setups of adapting-based methods are aligned with the official papers and the training setups keep the same as ours. Additionally, we adopt the performance-aware oracle criterion to the layer-wise pruning method \cite{KDSG17} as one of adapting-based methods for comparisons. Table \ref{nyud} presents the compression results of MTI-Net on NYUD-v2 and PASCAL Context, respectively. 

\begin{table}[t]
\centering

\caption{Compression results of SSD300 based on four algorithms. ``$\dagger$'' denotes the results from several iterations of pruning. ``$\clubsuit$'' denotes the result from one pruning iteration.}
\label{pruning_algorithm_compare}
\resizebox{0.9\columnwidth}{!}{
\begin{tabular}{@{}lccc@{}}
\toprule
Strategy & mAP (\%) & FLOPs (B) & \#Params (M) \\ \midrule
Layer-wise$^\dagger$    & 73.2   & 8.1 & 4.2  \\
Global \cite{KDSG17}$^\clubsuit$        & 74.3   & 19.2 & 8.3  \\
Group \cite{Singh_2019}$^\dagger$         & 75.2   & 13.5 & 3.9  \\ \midrule
Ours$^\dagger$          & \textbf{75.6}   & \textbf{8.1} & \textbf{3.8} \\ \bottomrule
\end{tabular}}
\end{table}

\begin{table}[]
\centering

\caption{Compression results of SSD300 based on four performance drop constraints with one pruning iteration.}
\label{task-aware_compare}
\resizebox{0.9\columnwidth}{!}{
\begin{tabular}{@{}lccc@{}}
\toprule
Constraint & mAP (\%) & FLOPs(B) & \#Params (M) \\ \midrule
$\ell_1$ (loss sum) & 75.4 & 10.7 & 9.4 \\
$\ell_2$ & 75.6 & 10.9 & 6.3 \\
min & 49.0 & 0.8 & 2.6 \\ \midrule
Ours (max) & \textbf{75.9} & \textbf{10.6} & \textbf{5.7} \\ \bottomrule
\end{tabular}}
\end{table}

\subsubsection{Results on NYUD-v2}\label{Results on NYUD-v2}
The base model is \textbf{reproduced} in our trainer and the results of two lightweight models \cite{vandenhende2020mti, xu2018pad} are referred from \cite{vandenhende2020mti}. The results on NYUD-v2 show that PAGCP can maintain higher performances for all tasks at about 50\% compression ratio of FLOPs, outperforming both finetuning- and adapting-based pruning methods. An observation from Table \ref{nyud} is that the reduction rate of parameters is relatively smaller than that of FLOPs. A possible reason is that MTI-Net contains lots of attention blocks in the feature propagation stage, which occupy most FLOPs but with few parameters for the computation of self-correlation matrices. Therefore, the pruning of attention blocks may bring a smaller gain of parameter reduction than FLOPs reduction. Still, the pruned MTI-Net achieves at most +3.6\% mIoU for semantic segmentation and -0.28 rmse for depth estimation than existing multitask models.

\subsubsection{Results on PASCAL Context}
We also re-implement the base model training on PASCAL Context, and the results of two lightweight models \cite{vandenhende2020mti, xu2018pad} are referred from \cite{vandenhende2020mti}. We can conclude from Table \ref{nyud} that the compressed model using PAGCP can significantly reduce the FLOPs number by 47.7\%, yet achieve comparable performance for each task with the original model. Further, the parameter reduction by the compressed model is 25.4\%, which is not as significant as FLOPs reduction due to the same reason as analyzed in Section \ref{Results on NYUD-v2}.

\section{Insights and Discussion}
\label{discussion}
In this section, we discuss the impact of each component on the pruning performance in our method, including the sequentially greedy pruning fashion, the performance-aware oracle criterion and several hyper-parameters. Furthermore, to test the compatibility of our method with the single task setting, we apply the PAGCP to the classification models. Then, we are interested in the acceleration performance of the pruned models and test the latency of various models on different platforms. Finally, we visualize the compressed model structure, the sensitivity distribution of each layer along the pruning process and several prediction results of pruned models for a clear analysis. 

\begin{table}[t]
\centering

\caption{The study of the performance-aware criterion on NYUD-v2. The ``init. pred.'' represents the output set of the corresponding task at the initial stage. The ``final pred.'' refers to the final output at the final stage.}
\label{task_number}
\resizebox{\columnwidth}{!}{
\begin{tabular}{@{}cccccccc@{}}
\toprule
\multicolumn{2}{c}{Seg.} & \multicolumn{2}{c}{Dep.} & \multirow{2}{*}{mIoU} & \multirow{2}{*}{rmse} & \multirow{2}{*}{FLOPs} & \multirow{2}{*}{\#Params} \\ \cmidrule(r){1-4}
Init. & Final & Init. & Final & (\%) &  & (B) & (M) \\ 
pred. & pred. & pred. & pred. & & & & \\\midrule
$\checkmark$ &  &  &  & 35.9 & 0.622 & 5.8 & 5.0 \\
 & $\checkmark$ &  &  & 34.8 & 0.636 & 5.3 & 3.6 \\
 &  & $\checkmark$ &  & 35.1 & 0.618 & 6.1 & 4.7 \\
 &  &  & $\checkmark$ & 34.5 & 0.613 & 7.0 & 4.7 \\
$\checkmark$ & $\checkmark$ &  &  & 36.9 & 0.616 & 6.6 & 4.9 \\
 &  & $\checkmark$ & $\checkmark$ & 34.9 & 0.614 & 7.2 & 4.6 \\
$\checkmark$ & $\checkmark$ & $\checkmark$ & $\checkmark$ & 37.4 & 0.595 & 9.3 & 6.5 \\ \bottomrule
\end{tabular}
}
\end{table}

\subsection{Ablation Study}
\label{ablation}
\noindent\textbf{Sequentially Greedy Pruning.}
We compare the proposed method with three other pruning strategies, including layer-wise, global, and group pruning. For layer-wise pruning, we prune filters with the same ratio in each layer as our method does. For global pruning, we use mean $\ell_1$ norm \cite{KDSG17} to get the pruning set and prune them in one iteration. It can be observed from Table \ref{pruning_algorithm_compare} that the sequentially greedy algorithm performs much better in evaluating the joint saliency of filters than regular pruning strategies, thus achieving the highest accuracy with the largest compression ratio.

\begin{figure*}[t]
\centering
\begin{minipage}{.245\textwidth}
  \centering
  {\includegraphics[width=0.95\linewidth]{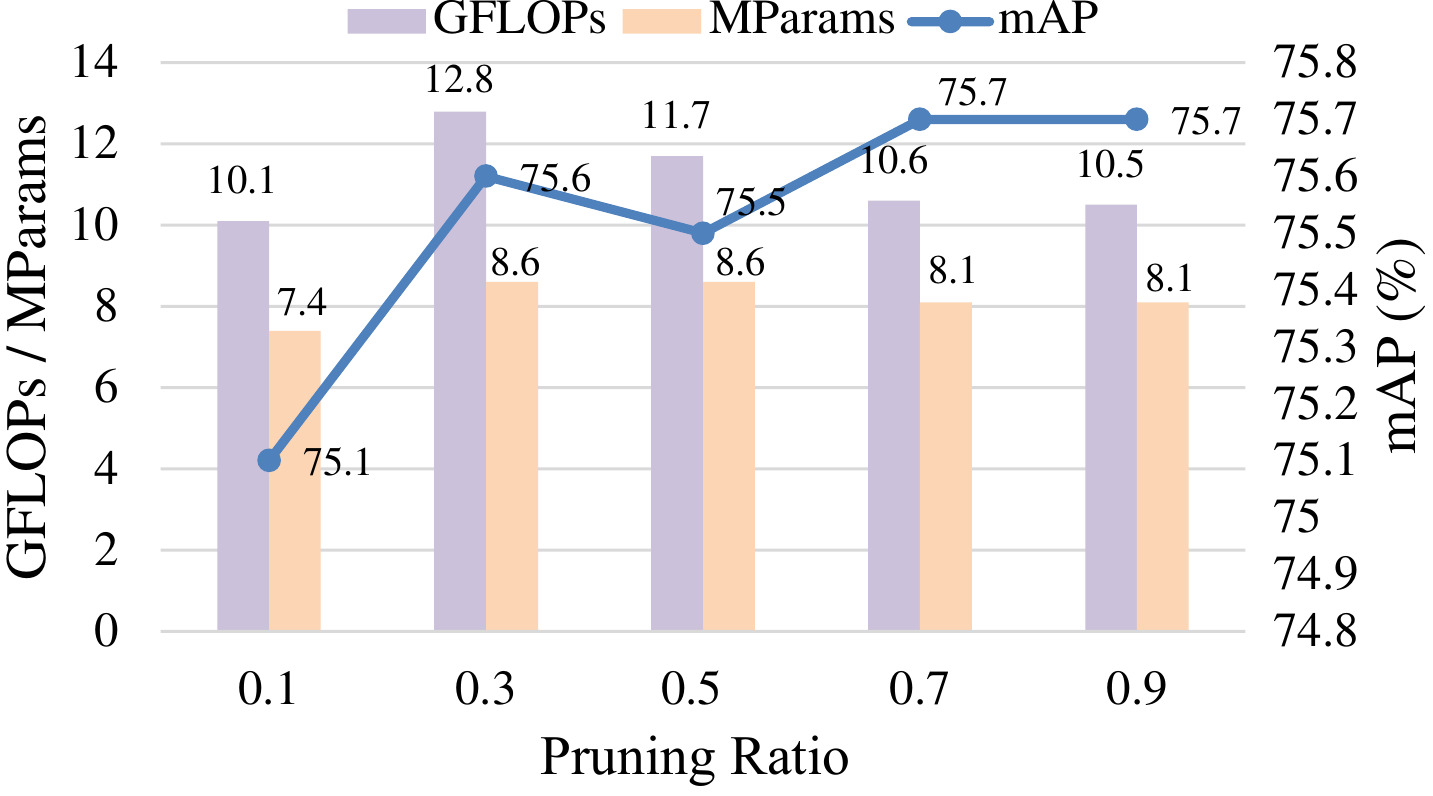}}
  \label{pruning_ratio}
\end{minipage}
\begin{minipage}{.245\textwidth}
  \centering
  {\includegraphics[width=0.95\linewidth]{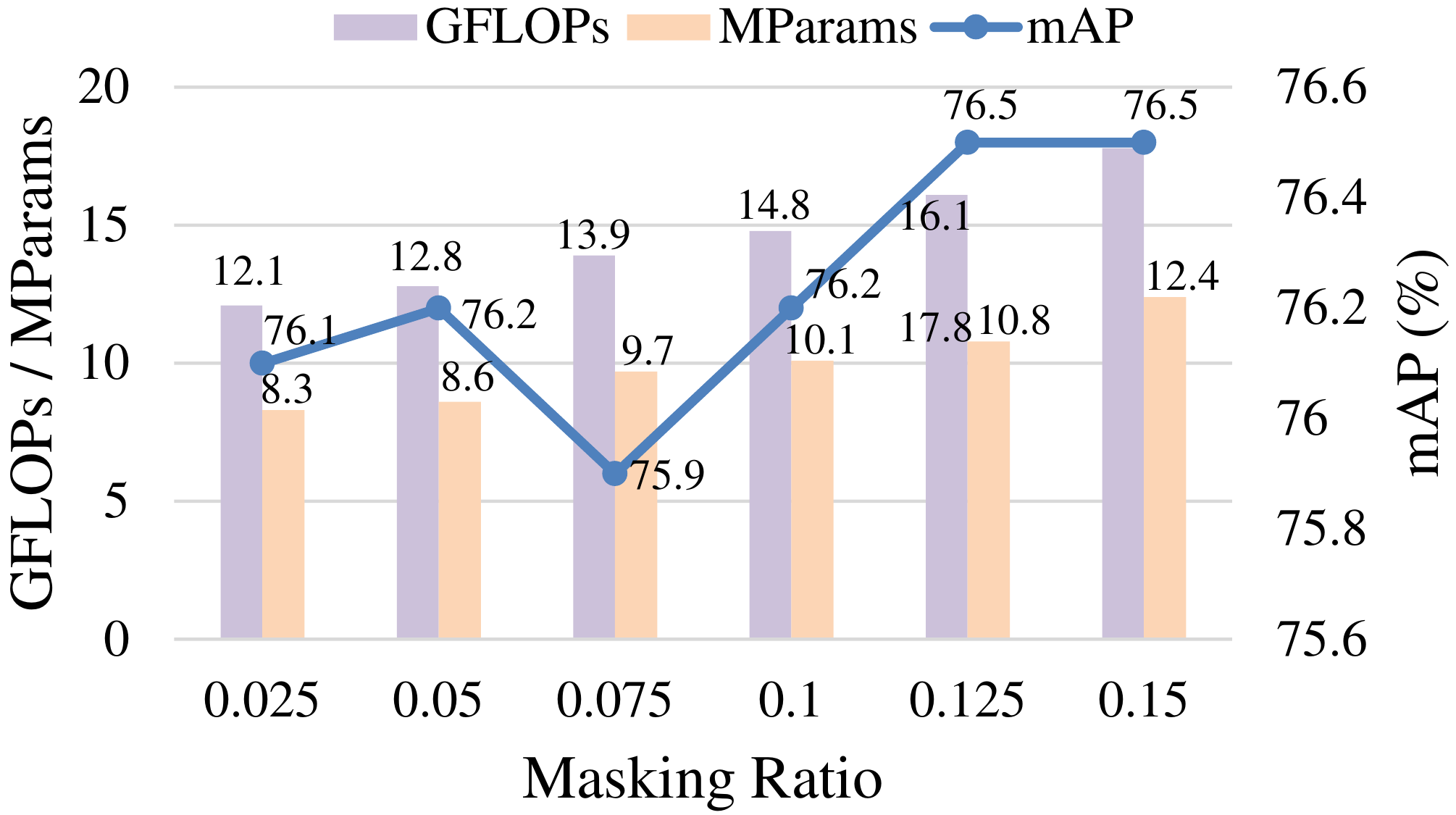}}
  \label{masking_ratio}
\end{minipage}
\begin{minipage}{.245\textwidth}
  \centering
  {\includegraphics[width=0.95\linewidth]{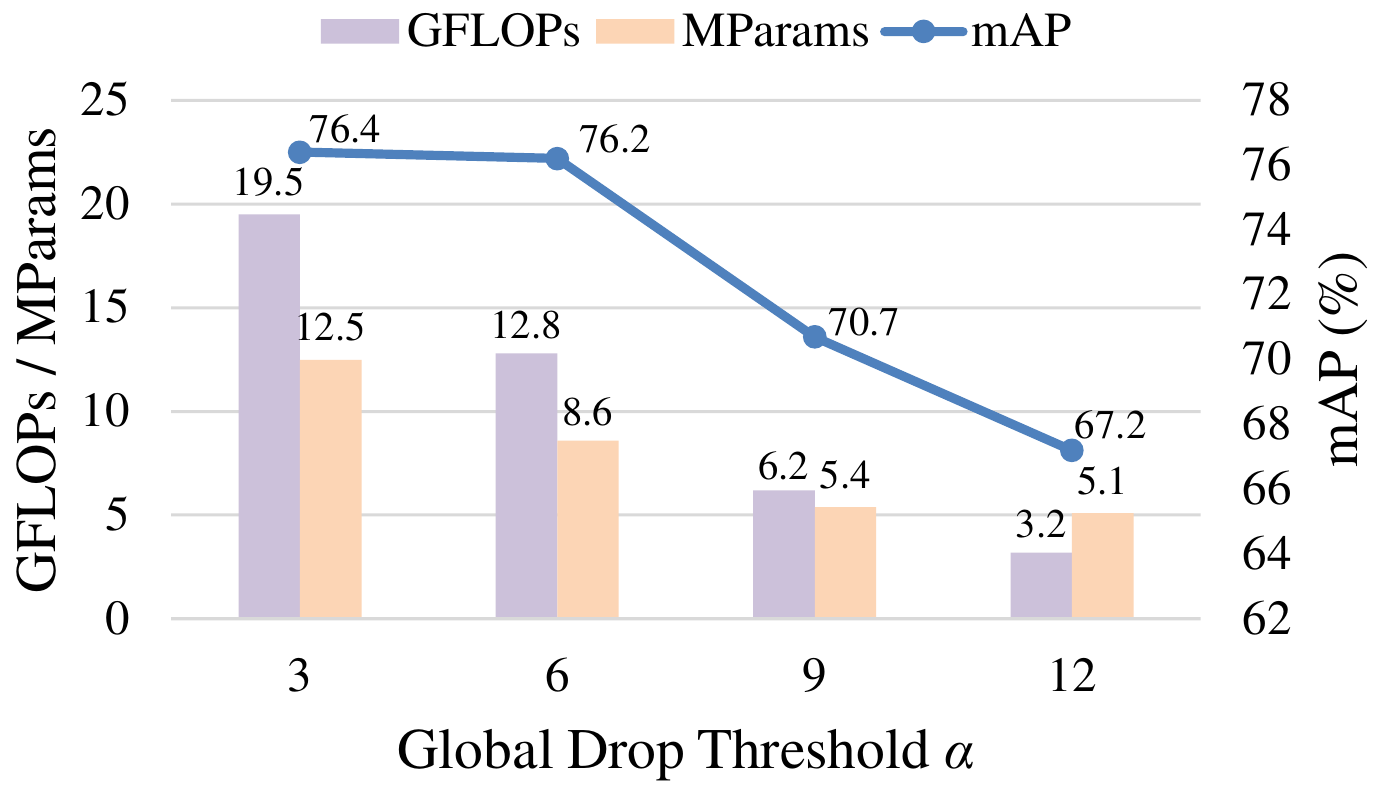}}
  \label{alpha}
\end{minipage}
\begin{minipage}{.245\textwidth}
  
 \centering {\includegraphics[width=0.85\linewidth]{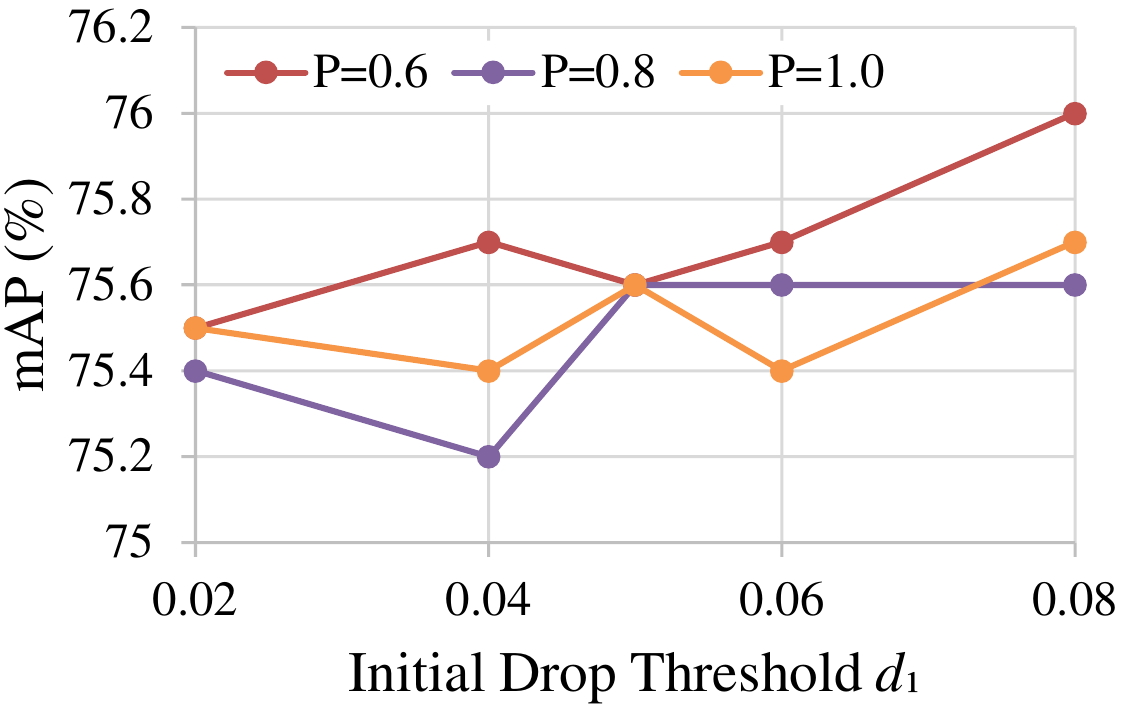}}
\end{minipage}
\caption{Hyper-parameter analysis. Column 1 figure denotes the study on the pruning ratio during the initialization of the pruning sequence. Column 2 figure shows the study on the masking ratio during the pruning in each layer. Column 3 figure presents the study on the global performance drop $\alpha$. Column 4 shows the study on the initial drop threshold $d_1$ and the filtering ratio of pruning layers $P$. Except for Column 4 figure, where all schemes iteratively prune SSD300 with 70\% FLOPs reduction, the rest experiments in the other three charts are conducted by pruning SSD300 only once.}
\label{sensitivity_hyp}
\end{figure*}

\noindent\textbf{Performance-aware Oracle Criterion.}
\label{tapm}

1) \textbf{Effect of $\ell_{\infty}$ Metric}: We test three more metrics, \textit{i.e.}, $\ell_1$ norm, $\ell_2$ norm and the minimum mapping. Specifically, we conduct the compression experiments for SSD300 on PASCAL VOC with the same compression setup for different metrics-based oracle criteria ($\alpha$=6, $d_1$=0.06, $P$=0.8). The results are shown in Table \ref{task-aware_compare}, from which we can observe that the minimum mapping achieves the highest compression ratio yet hurts the compression performance severely, and the maximum-based criterion achieves the best compression performance with acceptable FLOPs and parameters reduction. The reason for the severe performance degradation by the minimum norm is that, the criterion aims at finding the minimal pruning ratio to ensure the performance of all tasks drops below the local threshold without differentiating these tasks’ different sensitivity for various filters, which inevitably prunes many important filters.

2) \textbf{Effect of the Task Number}: We discuss the compression performance under different task numbers to be considered in the performance-aware oracle criterion. Taking MTI-Net on NYUD-v2 as an example, to simplify the exploration of the performance-aware criterion, we categorize the tasks into four sets of subtasks on NYUD-v2 according to the located stage of each task. The four multi-scale outputs of one task are wrapped as one task, with the final prediction bound as another task. In this way, the tasks on NYUD-v2 are divided into four subtasks, annotated as initial and final subtasks under two tasks respectively in Table \ref{task_number}. The experiments in Table \ref{task_number} show that despite the slightly larger compression ratio (\textit{e.g.}, 3.6 MParams \textit{vs.} 6.5 MParams), reducing the task number for compression seriously impairs the performance. Conversely, the model can recover more when more tasks are included in the compression criterion.

\noindent\textbf{Hyper-parameter Analysis.}
We further study the impact of changing hyper-parameters in our PAGCP on the compression performance. In particular, there are five hyper-parameters that may affect the compression performance, including the pruning ratio during the initialization of the pruning sequence, the masking ratio $\gamma$ during the pruning of each layer, the initial drop threshold $d_1$, the filtering ratio of pruning layers $P$, and the global performance drop threshold $\alpha$. Intuitively, $d_1$ and $\alpha$ determines the main reduction ratio of FLOPs, while the filtering ratio variable $P$ is devoted to balance the pruning ratio and the pruning quality. For the study on the pruning ratio and the masking ratio, we sample them from the sets $\{$0.1, 0.3, 0.5, 0.7, 0.9$\}$ and $\{$0.025, 0.05, 0.075, 0.1, 0.125, 0.15\}, respectively. Other hyper-parameters are set as: $d_1$=0.6, $\alpha$=6, $P$=0.8. For the study on $d_1$ and $P$, we sample $d_1, P$ from the sets $\{$0.02, 0.04, 0.05, 0.06, 0.08$\}$ and $\{$0.6, 0.8, 1.0\}, respectively. Other hyper-parameters are set as: $\alpha$=6, pruning ratio=0.3, masking ratio=0.05. For the study on $\alpha$, we sample $\alpha$ from the set $\{$3, 6, 9$\}$ and set other hyper-parameters as: pruning ratio=0.3, masking ratio=0.05, $d_1$=0.6, $P$=0.8. The compression results by tuning these parameters are shown in Fig. \ref{sensitivity_hyp}. We can observe that (1) the compression performance keeps relatively stable when the pruning ratio gets larger than 0.3, (2) a small masking ratio can get a large compression ratio due to the more fine-grained selection for the candidate filters, (3) a large performance drop threshold for the first pruning layer along with a mild filtering ratio of each layer and a mild global performance drop threshold can maintain a good performance under the same pruning ratio, which suggests the necessity of exploring the pruning space of each layer. 

\begin{table}[t]
\centering
\caption{Compression results of ResNet50 on ImageNet.}
\label{imagenet}
\resizebox{\columnwidth}{!}{%
\begin{tabular}{@{}llccccc@{}}
\toprule
 &  & \multicolumn{2}{c}{Top-1(\%)} & \multicolumn{2}{c}{Top-5 (\%)} & FLOPs↓ \\ \cmidrule(lr){3-6}
 \multirow{-2}{*}{Model} & \multirow{-2}{*}{Method} & Baseline & Pruned & Baseline & Pruned & (\%) \\ \midrule
 & FPGM\cite{He_2019_CVPR} & 76.15 & 75.59 & 92.87 & 92.63 & 42.2 \\
 & Taylor \cite{8953464} & 76.18 & 74.50 & - & - & 44.9 \\
 & GAL\cite{Lin_2019_CVPR} & 76.15 & 71.95 & 92.87 & 90.94 & 43.0 \\
 & RRBP \cite{zhou2019accelerate} & 76.10 & 73.00 & 92.90 & 91.00 & 54.5 \\
 & TAS\cite{dong2019tas} & 77.46 & 76.20 & 93.55 & 93.07 & 43.5 \\
 & Autopruner \cite{luo2020autopruner} & 76.15 & 74.76 & 92.87 & 92.15 & 48.7 \\
 & PFP \cite{liebenwein2019provable} & 76.13 & 75.21 & 92.87 & 92.81 & 10.8 \\
 & HRank\cite{9156677} & 76.15 & 74.98 & 92.87 & 92.33 & 43.7 \\
 & SCOP\cite{tang2020scop} & 76.15 & 75.95 & 92.87 & 92.79 & 45.3 \\
 & NPPM\cite{NPPM} & 76.15 & 75.96 & 92.87 & 92.75 & 56.0 \\
 & DCP\cite{DCP} & 76.01 & 75.15 & 92.93 & 92.20 & 52.4 \\
 & ResRep \cite{resrep} & 76.15 & 75.30 & 92.87 & 92.47 & 62.1 \\
 & CHIP \cite{sui2021chip} & 76.15 & 75.26 & 92.87 & 92.53 & 62.8 \\ \cmidrule(l){2-7} 
\multirow{-14}{*}{ResNet50} & \textbf{Ours} & 76.15 & \textbf{76.01} & 92.87 & \textbf{92.81} & \textbf{61.0} \\ \bottomrule
\end{tabular}%
}
\end{table}

\subsection{Specialization of PAGCP on Classification Task}
Since the classification task is a special case of multitask learning, we also test the compatibility of our PAGCP with the single task setting. In this case, the performance-aware oracle criterion degenerates into the vanilla oracle form, \textit{i.e.}, the loss of classification predictions. For fair comparisons, we follow previous pruning works and compress ResNet50 \cite{resnet} on ImageNet \cite{deng2009imagenet}. The compression setup is experimentally set as: $d_1$=0.06, $\alpha$=5, $P$=0.8. We finetune the pruned model with the standard training setup, \textit{i.e.}, batchsize=256, initial learning rate=0.01 and cosine annealing for 180 epochs. The compression result is presented in Table \ref{imagenet}. Compared with state-of-the-art pruning methods, our method maintains the smallest accuracy drop in both Top-1 (-0.14\%) and Top-5 (-0.06\%) while achieving a large compression ratio (61.0\% drop in FLOPs), which further demonstrates the essence of joint saliency and the effectiveness of PAGCP (See Appendix \ref{cifar_benchmark} for more results).

\begin{table}[]
\small
\centering
\caption{Acceleration of the inference latency (ms/image) for various pruned models on both cloud and mobile platforms. models suffixed with `\_p' are pruned by \textbf{PAGCP}.}
\label{latency_compare}
\resizebox{\columnwidth}{!}{
\begin{tabular}{@{}llcccccc@{}}
\toprule
\multirow{2}{*}{Dataset} & \multirow{2}{*}{Model} & \multicolumn{2}{c}{RTX 3090} & \multicolumn{2}{c}{ARM} & \multicolumn{2}{c}{OpenCL} \\ \cmidrule(l){3-8} 
 &  & 320 & 640 & 320 & 640 & 320 & 640 \\ \midrule
\multirow{3}{*}{PASCAL VOC} & SSD & 23.9 & 39.2 & 316.5 & 1196 & 164.6 & 667.4 \\
 & SSD\_p & 8.3 & 14.7 & 99.35 & 366.9 & 61.23 & 224.5 \\ \cmidrule(l){2-8} 
 & Accelerate & \textbf{2.9x} & \textbf{2.7x} & \textbf{3.2x} & \textbf{3.3x} & \textbf{2.7x} & \textbf{3.0x} \\ \midrule
\multirow{7}{*}{COCO} & YOLOv5m & 1.6 & 4.1 & 130.4 & 515.2 & 52.9 & 168.4 \\
 & YOLOv5s & 0.9 & 1.7 & 55.5 & 225.2 & 29.3 & 76.2 \\
 & YOLOX-S & 0.7 & 1.9 & 68.4 & 273.8 & 28.53 & 99.7 \\
 & NS-YOLOv5 & 0.9 & 2.6 & 85.5 & 337.8 & 39.2 & 114.7 \\
 & Eagleeye-YOLOv5 & 0.9 & 2.6 & 76.3 & 305.5 & 33.4 & 106.3 \\
 & YOLOv5m\_p & 0.8 & 2.3 & 64.8 & 264.7 & 30.0 & 89.9 \\ \cmidrule(l){2-8} 
 & Accelerate & \textbf{2.0x} & \textbf{1.8x} & \textbf{2.0x} & \textbf{1.9x} & \textbf{1.8x} & \textbf{1.9x} \\ \midrule
\multirow{5}{*}{NYUD-v2} & MTI-Net & 10.8 & 11.8 & 166.8 & 496.3 & - & - \\
 & NS-MTI & 12.0 & 13.0 & 134.5 & 401.7 & - & - \\
 & Eagleeye-MTI & 10.3 & 11.8 & 112.3 & 330.1 & - & - \\
 & MTI-Net\_p & 9.4 & 11.2 & 117.7 & 350.0 & - & - \\ \cmidrule(l){2-8} 
 & Accelerate & \textbf{1.2x} & \textbf{1.1x} & \textbf{1.4x} & \textbf{1.4x} & - & - \\ \bottomrule
\end{tabular}}
\end{table}

\subsection{Acceleration Study}
We further explore the acceleration results after pruning. We conduct the experiments on three benchmark datasets, with sufficient validations on various models of different types of resolution and platforms. Specifically, we measure the latency on two platforms, including the cloud platform (a single NVIDIA GeForce RTX 3090), and a mobile platform based on a single Snapdragon 855 processor. The latency calculation adopts images with two types of resolutions (320×320 and 640×640) as input on both platforms. On the cloud platform, we report the latency based on batch size of 32. And on the mobile platform, we report the latency based on batch size of 1 with both ARM (CPU) and OpenCL (GPU) frameworks. For fair comparisons, we test several lightweight models based on previous compression methods on the same platforms,\textit{ e.g.}, YOLOX-S, the slimming-based pruned models and the Eagleeye-based pruned models. The results are shown in Table \ref{latency_compare}. Generally, we can obtain 1.2x$\sim$3.3x acceleration for the pruned models on all platforms with both types of resolution and different benchmarks. Our method achieves the most latency acceleration for SSD models and the similar latency to YOLOv5s and YOLOX-S for YOLOv5m. For MTI-Net, we report the latency time on ARM and RTX3090 since some operations are not supported for OpenCL, which also explains why the acceleration of MTI-Net is not as significant as other two models. Still, compared with previous pruning methods, PAGCP can achieve the most latency acceleration on different models and datasets (See Appendix \ref{pruning_complexity} for more analysis).

\begin{figure}[t]
    \centering
    \resizebox{\columnwidth}{!}{
    \includegraphics{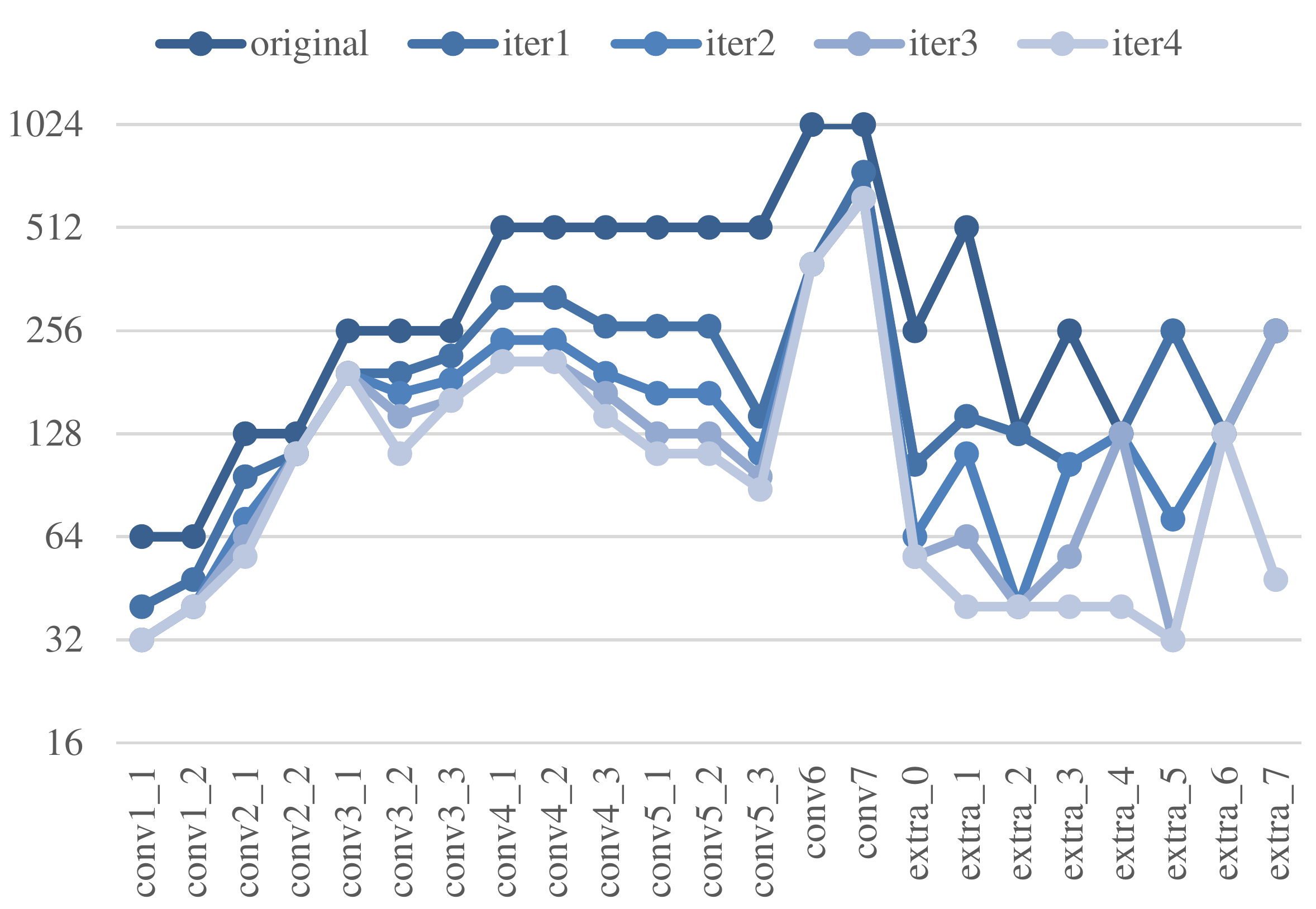}}
    \caption{Widths of layers in the pruned SSD300 at each pruning iteration.}
    \label{pruning_structure}
\end{figure}

\begin{figure}[t]
    \centering
    \resizebox{\columnwidth}{!}{
    \includegraphics{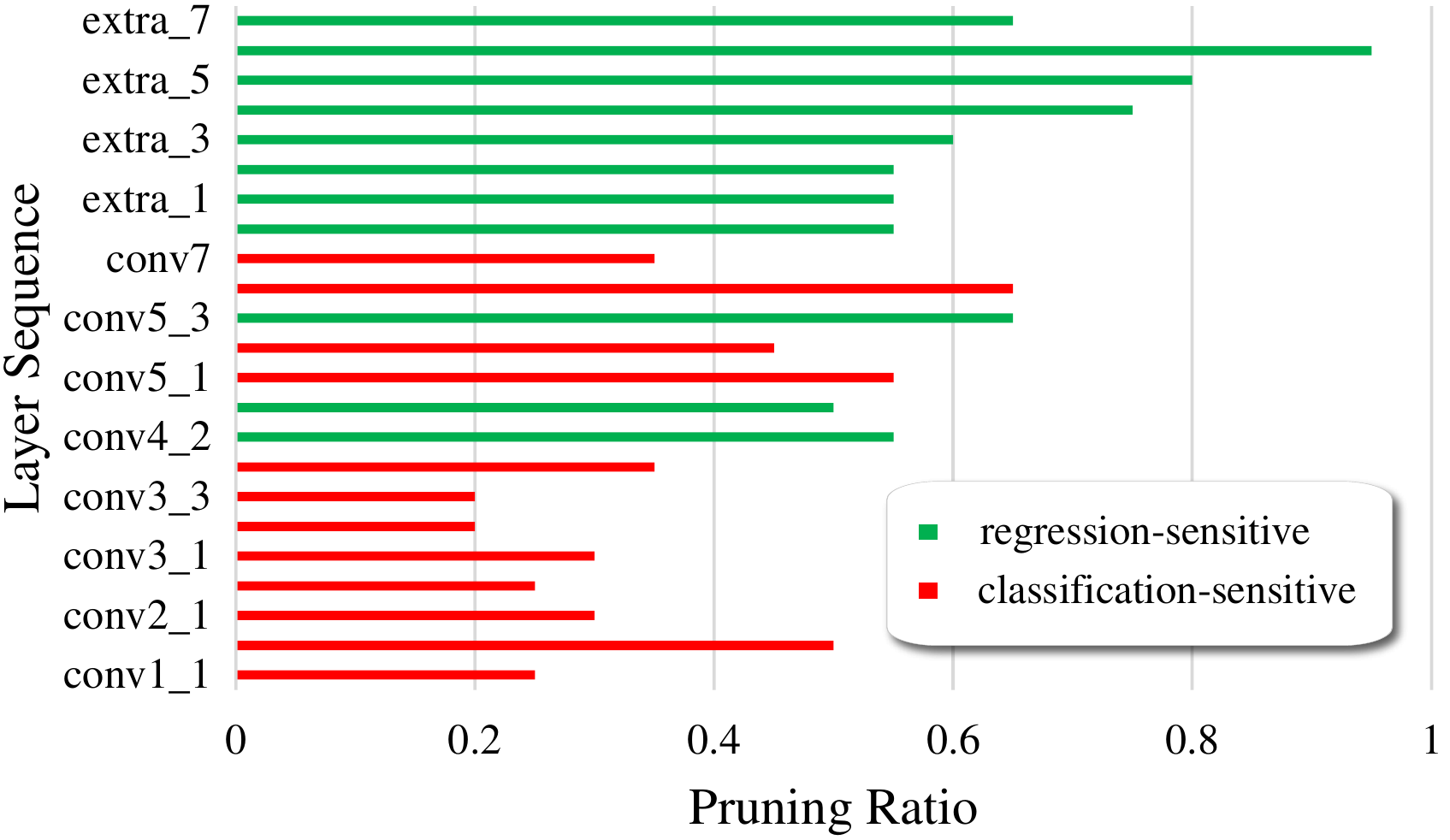}}
    \caption{The sensitivity in each layer of SSD300 with two tasks for the model. The vertical and horizontal axis denote the layer sequence and the pruning ratio, respectively. Bars with different colors represent different task-sensitive groups.}
    \label{sensitivity_distribution}
\end{figure}
\subsection{Visualizations}
\noindent\textbf{Visualization of the Pruned Structure.} For a clear visualization of the structure change, we choose VGG16-SSD, which has a simple and clear structure, as the target model to visualize the pruning procedure. Results are shown in Fig. \ref{pruning_structure}. We list the width, \textit{i.e.}, channel number, of each target layer at each iteration of pruning SSD300 on PASCAL VOC dataset. From Fig. \ref{pruning_structure}, it can be observed that 1) deep layers are pruned more than shallow layers, which is consistent with the prior work \cite{resrep} that SSD300 maybe over-fit on PASCAL VOC dataset, and 2) the width ratios between neighboring layers are preserved in most layers, which reflects that PAGCP relies more on mining the neighboring inter-layer filter correlations for multitask model pruning.

\begin{figure}[ht]
    \centering
    \resizebox{\columnwidth}{!}{
    \includegraphics{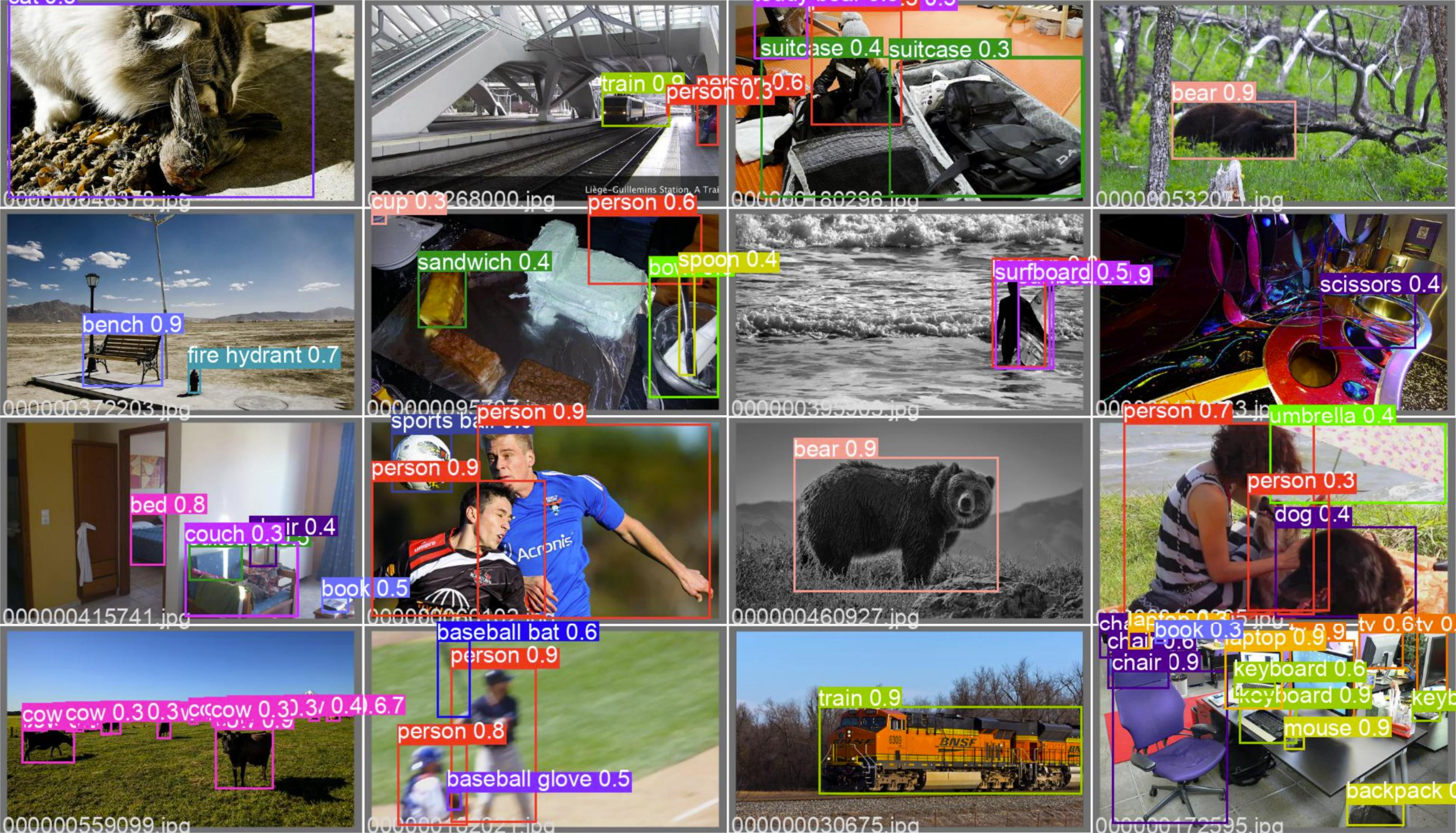}}
    \caption{The detection results of the pruned YOLOv5m on COCO2017, where different colors of bounding boxes refer to different classes and the number on the top left of each bounding box refers to the prediction confidence.}
    \label{vis_coco}
\end{figure}
\begin{figure}[ht]
    \centering
    \resizebox{\columnwidth}{!}{
    \includegraphics{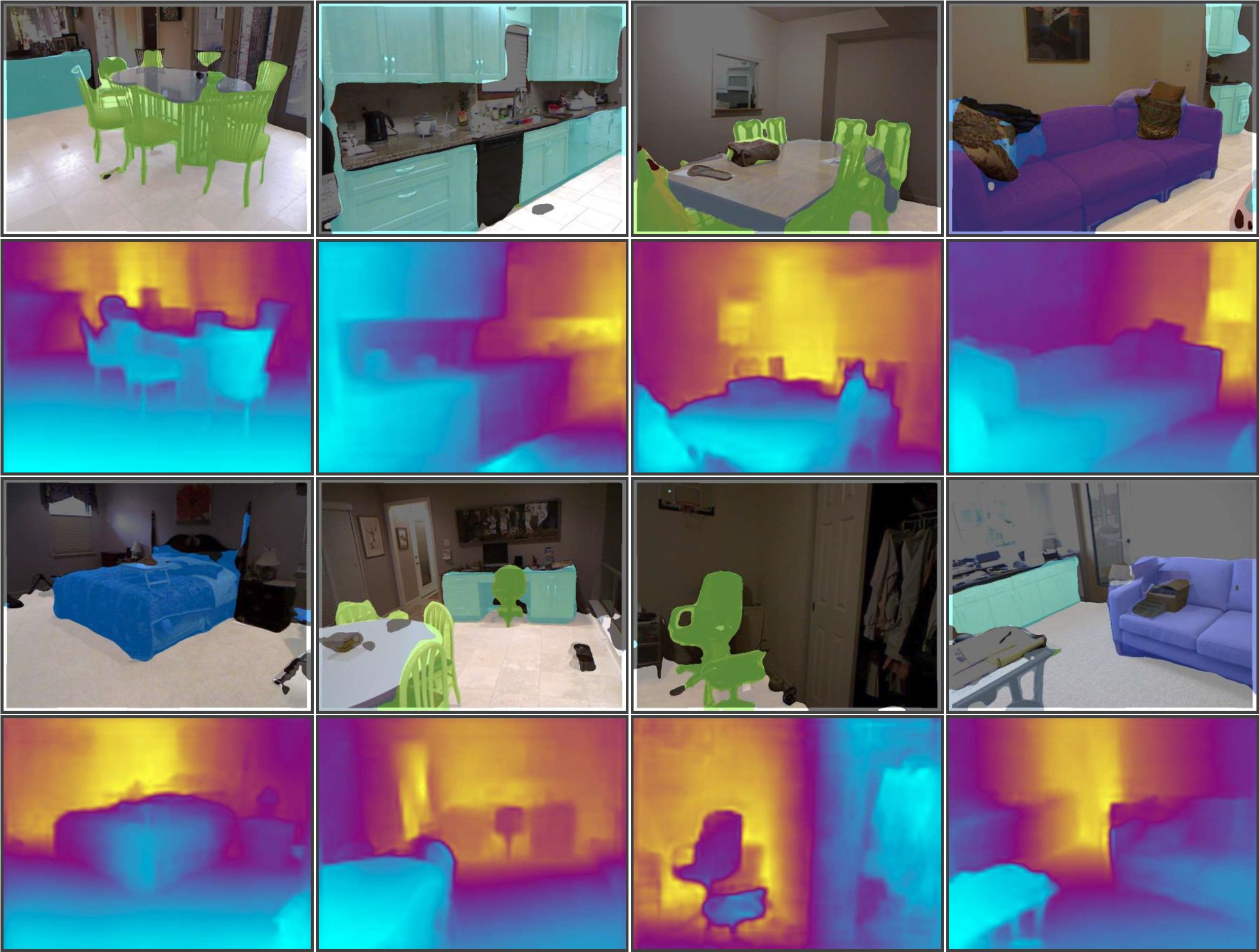}}
    \caption{The multitask prediction results of the pruned MTI-Net on NYUD-v2, where Row 1 and 3 are the semantic segmentation maps and Row 2 and 4 are the corresponding depth maps.}
    \label{vis_nyud}
\end{figure}

\noindent\textbf{Visualization of the Sensitivity Distribution.} We provide the distribution of the most sensitive task during pruning for each layer of SSD300, as shown in Fig. \ref{sensitivity_distribution}. It can be observed that any task in a model is likely to become the most sensitive task in a certain layer, where the classification-oriented tasks seem to have more sensitivity than regression-oriented tasks (See Appendix \ref{distribution} for more examples). The reason may be that the classification-related features are usually more abstract and have high-level semantics, which require a deeper and more complex structure by stacking multiple non-linear mapping layers for extraction than regression-related features.

\noindent\textbf{Visualization of Predictions. }We present some object detection results of the pruned YOLOv5m model on COCO and conventional multitask learning results of the pruned MTI-Net on NYUD-v2, which are shown in Fig. \ref{vis_coco} and Fig. \ref{vis_nyud}. The results indicate that the pruned models using PAGCP still perform well on multitask learning (See Appendix \ref{limitation_discussion} for some failure analysis).

\section{Conclusion}
\label{conclusion}
In this paper, we propose an effective Performance-Aware Global Channel Pruning (PAGCP) framework, to compress various models including object detection, conventional multitask dense predictions, and even classification. We theoretically derive the optimization objective of PAGCP, and develop a sequentially greedy pruning strategy to approximately solve the objective problem. In particular, the developed pruning strategy consists of two key modules, including a sequential channel pruning scheme considering different filters' co-influence on compression performance within and between layers, and a performance-aware oracle criterion considering different tasks have different performance-sensitive filters for keeping. Extensive experiments on multiple detection, MTL, and classification datasets demonstrate that the proposed PAGCP can achieve state-of-the-art compression performance in terms of prediction accuracy, parameter and FLOPs, and also can be well generalized to various one-stage and two-stage detectors, MTL models, and classification networks. We also give the real inference time of the compressed model using the proposed PAGCP, on both cloud and mobile platforms, to show the real-time application potential of our compressed model on mobile devices.

\ifCLASSOPTIONcaptionsoff
  \newpage
\fi

\bibliographystyle{IEEEtran}
\bibliography{main}

\begin{thebibliography}{10}
\providecommand{\url}[1]{#1}
\csname url@samestyle\endcsname
\providecommand{\newblock}{\relax}
\providecommand{\bibinfo}[2]{#2}
\providecommand{\BIBentrySTDinterwordspacing}{\spaceskip=0pt\relax}
\providecommand{\BIBentryALTinterwordstretchfactor}{4}
\providecommand{\BIBentryALTinterwordspacing}{\spaceskip=\fontdimen2\font plus
\BIBentryALTinterwordstretchfactor\fontdimen3\font minus
  \fontdimen4\font\relax}
\providecommand{\BIBforeignlanguage}[2]{{%
\expandafter\ifx\csname l@#1\endcsname\relax
\typeout{** WARNING: IEEEtran.bst: No hyphenation pattern has been}%
\typeout{** loaded for the language `#1'. Using the pattern for}%
\typeout{** the default language instead.}%
\else
\language=\csname l@#1\endcsname
\fi
#2}}
\providecommand{\BIBdecl}{\relax}
\BIBdecl

\bibitem{romero2014fitnets}
A.~Romero, N.~Ballas, S.~E. Kahou, A.~Chassang, C.~Gatta, and Y.~Bengio,
  ``Fitnets: Hints for thin deep nets,'' \emph{arXiv preprint arXiv:1412.6550},
  2014.

\bibitem{polino2018model}
A.~Polino, R.~Pascanu, and D.-A. Alistarh, ``Model compression via distillation
  and quantization,'' in \emph{6th International Conference on Learning
  Representations}, 2018.

\bibitem{chung2020feature}
I.~Chung, S.~Park, J.~Kim, and N.~Kwak, ``Feature-map-level online adversarial
  knowledge distillation,'' in \emph{International Conference on Machine
  Learning}.\hskip 1em plus 0.5em minus 0.4em\relax PMLR, 2020, pp. 2006--2015.

\bibitem{wu2016quantized}
J.~Wu, C.~Leng, Y.~Wang, Q.~Hu, and J.~Cheng, ``Quantized convolutional neural
  networks for mobile devices,'' in \emph{Proceedings of the IEEE Conference on
  Computer Vision and Pattern Recognition}, 2016, pp. 4820--4828.

\bibitem{gu2019projection}
J.~Gu, C.~Li, B.~Zhang, J.~Han, X.~Cao, J.~Liu, and D.~Doermann, ``Projection
  convolutional neural networks for 1-bit cnns via discrete back propagation,''
  in \emph{Proceedings of the AAAI Conference on Artificial Intelligence},
  vol.~33, no.~01, 2019, pp. 8344--8351.

\bibitem{han2015deep}
S.~Han, H.~Mao, and W.~J. Dally, ``Deep compression: Compressing deep neural
  networks with pruning, trained quantization and huffman coding,''
  \emph{International Conference on Learning Representations (ICLR)}, 2016.

\bibitem{8416559}
J.-H. Luo, H.~Zhang, H.-Y. Zhou, C.-W. Xie, J.~Wu, and W.~Lin, ``Thinet:
  Pruning cnn filters for a thinner net,'' \emph{IEEE Transactions on Pattern
  Analysis and Machine Intelligence}, vol.~41, no.~10, pp. 2525--2538, 2019.

\bibitem{yu2018nisp}
R.~Yu, A.~Li, C.-F. Chen, J.-H. Lai, V.~I. Morariu, X.~Han, M.~Gao, C.-Y. Lin,
  and L.~S. Davis, ``Nisp: Pruning networks using neuron importance score
  propagation,'' in \emph{Proceedings of the IEEE Conference on Computer Vision
  and Pattern Recognition}, 2018, pp. 9194--9203.

\bibitem{frankle2018lottery}
J.~Frankle and M.~Carbin, ``The lottery ticket hypothesis: Finding sparse,
  trainable neural networks,'' in \emph{International Conference on Learning
  Representations}, 2018.

\bibitem{liu2020autocompress}
N.~Liu, X.~Ma, Z.~Xu, Y.~Wang, J.~Tang, and J.~Ye, ``Autocompress: An automatic
  dnn structured pruning framework for ultra-high compression rates,'' in
  \emph{Proceedings of the AAAI Conference on Artificial Intelligence},
  vol.~34, no.~04, 2020, pp. 4876--4883.

\bibitem{8237417}
Y.~He, X.~Zhang, and J.~Sun, ``Channel pruning for accelerating very deep
  neural networks,'' in \emph{2017 IEEE International Conference on Computer
  Vision (ICCV)}, 2017, pp. 1398--1406.

\bibitem{coreset}
B.~Mussay, D.~Feldman, S.~Zhou, V.~Braverman, and M.~Osadchy,
  ``Data-independent structured pruning of neural networks via coresets,''
  \emph{IEEE Transactions on Neural Networks and Learning Systems}, pp. 1--13,
  2021.

\bibitem{Joo_Yi_Baek_Kim_2021}
\BIBentryALTinterwordspacing
D.~Joo, E.~Yi, S.~Baek, and J.~Kim, ``Linearly replaceable filters for deep
  network channel pruning,'' \emph{Proceedings of the AAAI Conference on
  Artificial Intelligence}, vol.~35, no.~9, pp. 8021--8029, May 2021. [Online].
  Available: \url{https://ojs.aaai.org/index.php/AAAI/article/view/16978}
\BIBentrySTDinterwordspacing

\bibitem{liu2016ssd}
W.~Liu, D.~Anguelov, D.~Erhan, C.~Szegedy, S.~Reed, C.-Y. Fu, and A.~C. Berg,
  ``Ssd: Single shot multibox detector,'' in \emph{European conference on
  computer vision}.\hskip 1em plus 0.5em minus 0.4em\relax Springer, 2016, pp.
  21--37.

\bibitem{gradcam}
R.~R. Selvaraju, M.~Cogswell, A.~Das, R.~Vedantam, D.~Parikh, and D.~Batra,
  ``Grad-cam: Visual explanations from deep networks via gradient-based
  localization,'' in \emph{2017 IEEE International Conference on Computer
  Vision (ICCV)}, 2017, pp. 618--626.

\bibitem{liu2018rethinking}
Z.~Liu, M.~Sun, T.~Zhou, G.~Huang, and T.~Darrell, ``Rethinking the value of
  network pruning,'' in \emph{International Conference on Learning
  Representations}, 2018.

\bibitem{Singh_2019}
\BIBentryALTinterwordspacing
P.~Singh, R.~Manikandan, N.~Matiyali, and V.~P. Namboodiri, ``Multi-layer
  pruning framework for compressing single shot multibox detector,'' \emph{2019
  IEEE Winter Conference on Applications of Computer Vision (WACV)}, Jan 2019.
  [Online]. Available: \url{http://dx.doi.org/10.1109/WACV.2019.00145}
\BIBentrySTDinterwordspacing

\bibitem{XIE2020400}
\BIBentryALTinterwordspacing
Z.~Xie, L.~Zhu, L.~Zhao, B.~Tao, L.~Liu, and W.~Tao, ``Localization-aware
  channel pruning for object detection,'' \emph{Neurocomputing}, vol. 403, pp.
  400--408, 2020. [Online]. Available:
  \url{https://www.sciencedirect.com/science/article/pii/S092523122030429X}
\BIBentrySTDinterwordspacing

\bibitem{Ren2015Faster}
S.~Ren, K.~He, R.~Girshick, and J.~Sun, ``Faster r-cnn: Towards real-time
  object detection with region proposal networks.'' \emph{IEEE Transactions on
  Pattern Analysis and Machine Intelligence}, vol.~39, no.~6, pp. 1137--1149,
  2015.

\bibitem{zhou2019objects}
X.~Zhou, D.~Wang, and P.~Kr{\"a}henb{\"u}hl, ``Objects as points,'' in
  \emph{arXiv preprint arXiv:1904.07850}, 2019.

\bibitem{glenn2021yolov5}
\BIBentryALTinterwordspacing
G.~Jocher, A.~Stoken, J.~Borovec, NanoCode012, A.~Chaurasia, TaoXie,
  L.~Changyu, A.~V, Laughing, tkianai, yxNONG, A.~Hogan, lorenzomammana,
  AlexWang1900, J.~Hajek, L.~Diaconu, Marc, Y.~Kwon, oleg, wanghaoyang0106,
  Y.~Defretin, A.~Lohia, ml5ah, B.~Milanko, B.~Fineran, D.~Khromov, D.~Yiwei,
  Doug, Durgesh, and F.~Ingham, ``{ultralytics/yolov5: v5.0 - YOLOv5-P6 1280
  models, AWS, Supervise.ly and YouTube integrations},'' Apr. 2021. [Online].
  Available: \url{https://doi.org/10.5281/zenodo.4679653}
\BIBentrySTDinterwordspacing

\bibitem{vandenhende2020mti}
S.~Vandenhende, S.~Georgoulis, and L.~Van~Gool, ``Mti-net: Multi-scale task
  interaction networks for multi-task learning,'' in \emph{Proceedings of the
  European Conference on Computer Vision (ECCV)}.\hskip 1em plus 0.5em minus
  0.4em\relax Springer, 2020, pp. 527--543.

\bibitem{everingham2015pascal}
M.~Everingham, S.~A. Eslami, L.~Van~Gool, C.~K. Williams, J.~Winn, and
  A.~Zisserman, ``The pascal visual object classes challenge: A
  retrospective,'' \emph{International journal of computer vision}, vol. 111,
  no.~1, pp. 98--136, 2015.

\bibitem{lin2014microsoft}
T.-Y. Lin, M.~Maire, S.~Belongie, J.~Hays, P.~Perona, D.~Ramanan,
  P.~Doll{\'a}r, and C.~L. Zitnick, ``Microsoft coco: Common objects in
  context,'' in \emph{European conference on computer vision}.\hskip 1em plus
  0.5em minus 0.4em\relax Springer, 2014, pp. 740--755.

\bibitem{silberman2012indoor}
\BIBentryALTinterwordspacing
N.~Silberman, D.~Hoiem, P.~Kohli, and R.~Fergus, ``Indoor segmentation and
  support inference from rgbd images,'' in \emph{ECCV'12 Proceedings of the
  12th European conference on Computer Vision - Volume Part V}.\hskip 1em plus
  0.5em minus 0.4em\relax Springer-Verlag Berlin, October 2012, pp. 746--760.
  [Online]. Available:
  \url{https://www.microsoft.com/en-us/research/publication/indoor-segmentation-support-inference-rgbd-images/}
\BIBentrySTDinterwordspacing

\bibitem{context}
X.~Chen, R.~Mottaghi, X.~Liu, S.~Fidler, R.~Urtasun, and A.~Yuille, ``Detect
  what you can: Detecting and representing objects using holistic models and
  body parts,'' in \emph{2014 IEEE Conference on Computer Vision and Pattern
  Recognition}, 2014, pp. 1979--1986.

\bibitem{resnet}
\BIBentryALTinterwordspacing
K.~He, X.~Zhang, S.~Ren, and J.~Sun, ``Deep residual learning for image
  recognition,'' in \emph{2016 IEEE Conference on Computer Vision and Pattern
  Recognition (CVPR)}.\hskip 1em plus 0.5em minus 0.4em\relax Los Alamitos, CA,
  USA: IEEE Computer Society, jun 2016, pp. 770--778. [Online]. Available:
  \url{https://doi.ieeecomputersociety.org/10.1109/CVPR.2016.90}
\BIBentrySTDinterwordspacing

\bibitem{he2015Spatial}
------, ``Spatial pyramid pooling in deep convolutional networks for visual
  recognition,'' \emph{IEEE Transactions on Pattern Analysis and Machine
  Intelligence}, vol.~37, no.~9, pp. 1904--16, 2015.

\bibitem{zhang2018single}
S.~Zhang, L.~Wen, X.~Bian, Z.~Lei, and S.~Z. Li, ``Single-shot refinement
  neural network for object detection,'' in \emph{CVPR}, 2018.

\bibitem{redmon2016you}
J.~Redmon, S.~Divvala, R.~Girshick, and A.~Farhadi, ``You only look once:
  Unified, real-time object detection,'' in \emph{Proceedings of the IEEE
  conference on computer vision and pattern recognition}, 2016, pp. 779--788.

\bibitem{misra2016cross}
I.~Misra, A.~Shrivastava, A.~Gupta, and M.~Hebert, ``Cross-stitch networks for
  multi-task learning,'' in \emph{Proceedings of the IEEE conference on
  computer vision and pattern recognition}, 2016, pp. 3994--4003.

\bibitem{xu2018pad}
D.~Xu, W.~Ouyang, X.~Wang, and N.~Sebe, ``Pad-net: Multi-tasks guided
  prediction-and-distillation network for simultaneous depth estimation and
  scene parsing,'' in \emph{Proceedings of the IEEE Conference on Computer
  Vision and Pattern Recognition}, 2018, pp. 675--684.

\bibitem{gao2019nddr}
Y.~Gao, J.~Ma, M.~Zhao, W.~Liu, and A.~L. Yuille, ``Nddr-cnn: Layerwise feature
  fusing in multi-task cnns by neural discriminative dimensionality
  reduction,'' in \emph{CVPR}, 2019.

\bibitem{liu2019end}
S.~Liu, E.~Johns, and A.~J. Davison, ``End-to-end multi-task learning with
  attention,'' in \emph{Proceedings of the IEEE/CVF Conference on Computer
  Vision and Pattern Recognition}, 2019, pp. 1871--1880.

\bibitem{KDSG17}
H.~Li, K.~Asim, D.~Igor, S.~Hanan, and P.~G. Hans, ``Pruning filters for
  efficient convnets,'' in \emph{5th International Conference on Learning
  Representations, {ICLR} 2017, Toulon, France, April 24-26, 2017, Conference
  Track Proceedings}, 2017.

\bibitem{hu2016network}
H.~Hu, R.~Peng, Y.-W. Tai, and C.-K. Tang, ``Network trimming: A data-driven
  neuron pruning approach towards efficient deep architectures,'' in
  \emph{arXiv preprint arXiv:1607.03250}, 2016.

\bibitem{liu2017learning}
Z.~Liu, J.~Li, Z.~Shen, G.~Huang, S.~Yan, and C.~Zhang, ``Learning efficient
  convolutional networks through network slimming,'' in \emph{Proceedings of
  the IEEE international conference on computer vision}, 2017, pp. 2736--2744.

\bibitem{8953464}
P.~Molchanov, A.~Mallya, S.~Tyree, I.~Frosio, and J.~Kautz, ``Importance
  estimation for neural network pruning,'' in \emph{2019 IEEE/CVF Conference on
  Computer Vision and Pattern Recognition (CVPR)}, 2019, pp. 11\,256--11\,264.

\bibitem{8953212}
Y.~He, P.~Liu, Z.~Wang, Z.~Hu, and Y.~Yang, ``Filter pruning via geometric
  median for deep convolutional neural networks acceleration,'' in \emph{2019
  IEEE/CVF Conference on Computer Vision and Pattern Recognition (CVPR)}, 2019,
  pp. 4335--4344.

\bibitem{9156677}
M.~Lin, R.~Ji, Y.~Wang, Y.~Zhang, B.~Zhang, Y.~Tian, and L.~Shao, ``Hrank:
  Filter pruning using high-rank feature map,'' in \emph{2020 IEEE/CVF
  Conference on Computer Vision and Pattern Recognition (CVPR)}, 2020, pp.
  1526--1535.

\bibitem{howard2017mobilenets}
A.~G. Howard, M.~Zhu, B.~Chen, D.~Kalenichenko, W.~Wang, T.~Weyand,
  M.~Andreetto, and H.~Adam, ``Mobilenets: Efficient convolutional neural
  networks for mobile vision applications,'' \emph{arXiv preprint
  arXiv:1704.04861}, 2017.

\bibitem{pmlr-v97-tan19a}
\BIBentryALTinterwordspacing
M.~Tan and Q.~Le, ``{E}fficient{N}et: Rethinking model scaling for
  convolutional neural networks,'' in \emph{Proceedings of the 36th
  International Conference on Machine Learning}, ser. Proceedings of Machine
  Learning Research, K.~Chaudhuri and R.~Salakhutdinov, Eds., vol.~97.\hskip
  1em plus 0.5em minus 0.4em\relax PMLR, 09--15 Jun 2019, pp. 6105--6114.
  [Online]. Available: \url{https://proceedings.mlr.press/v97/tan19a.html}
\BIBentrySTDinterwordspacing

\bibitem{liu2019metapruning}
Z.~Liu, H.~Mu, X.~Zhang, Z.~Guo, X.~Yang, K.-T. Cheng, and J.~Sun,
  ``Metapruning: Meta learning for automatic neural network channel pruning,''
  in \emph{Proceedings of the IEEE/CVF international conference on computer
  vision}, 2019, pp. 3296--3305.

\bibitem{AMP}
L.~Zhang, G.~Chen, Y.~Shi, Q.~Zhang, M.~Tan, Y.~Wang, Y.~Tian, and T.~Huang,
  ``Anonymous model pruning for compressing deep neural networks,'' in
  \emph{2020 IEEE Conference on Multimedia Information Processing and Retrieval
  (MIPR)}.\hskip 1em plus 0.5em minus 0.4em\relax IEEE, 2020, pp. 157--160.

\bibitem{ye2022efficient}
P.~Ye, B.~Li, T.~Chen, J.~Fan, Z.~Mei, C.~Lin, C.~Zuo, Q.~Chi, and W.~Ouyang,
  ``Efficient joint-dimensional search with solution space regularization for
  real-time semantic segmentation,'' \emph{International Journal of Computer
  Vision}, vol. 130, no.~11, pp. 2674--2694, 2022.

\bibitem{ye2022beta}
P.~Ye, B.~Li, Y.~Li, T.~Chen, J.~Fan, and W.~Ouyang, ``$\beta$-darts:
  Beta-decay regularization for differentiable architecture search,'' in
  \emph{2022 IEEE/CVF Conference on Computer Vision and Pattern Recognition
  (CVPR)}.\hskip 1em plus 0.5em minus 0.4em\relax IEEE, 2022, pp.
  10\,864--10\,873.

\bibitem{li2022pruning}
Y.~Li, P.~Zhao, G.~Yuan, X.~Lin, Y.~Wang, and X.~Chen, ``Pruning-as-search:
  Efficient neural architecture search via channel pruning and structural
  reparameterization,'' in \emph{Thirty-First International Joint Conference on
  Artificial Intelligence}, 2022.

\bibitem{guo2021towards}
Y.~Guo, Y.~Zheng, M.~Tan, Q.~Chen, Z.~Li, J.~Chen, P.~Zhao, and J.~Huang,
  ``Towards accurate and compact architectures via neural architecture
  transformer,'' \emph{IEEE Transactions on Pattern Analysis and Machine
  Intelligence}, vol.~44, no.~10, pp. 6501--6516, 2021.

\bibitem{resrep}
X.~Ding, T.~Hao, J.~Tan, J.~Liu, J.~Han, Y.~Guo, and G.~Ding, ``Resrep:
  Lossless cnn pruning via decoupling remembering and forgetting,'' in
  \emph{Proceedings of the IEEE/CVF International Conference on Computer
  Vision}, 2021, pp. 4510--4520.

\bibitem{zhuang2018discrimination}
Z.~Zhuang, M.~Tan, B.~Zhuang, J.~Liu, Y.~Guo, Q.~Wu, J.~Huang, and J.~Zhu,
  ``Discrimination-aware channel pruning for deep neural networks,'' in
  \emph{Advances in Neural Information Processing Systems, 2018}.\hskip 1em
  plus 0.5em minus 0.4em\relax Neural Information Processing Systems (NIPS),
  2018, pp. 875--886.

\bibitem{he2021pruning}
X.~He, D.~Gao, Z.~Zhou, Y.~Tong, and L.~Thiele, ``Pruning-aware merging for
  efficient multitask inference,'' in \emph{Proceedings of the 27th ACM SIGKDD
  Conference on Knowledge Discovery \& Data Mining}, 2021, pp. 585--595.

\bibitem{dempster1977maximum}
A.~P. Dempster, N.~M. Laird, and D.~B. Rubin, ``Maximum likelihood from
  incomplete data via the em algorithm,'' \emph{Journal of the Royal
  Statistical Society: Series B (Methodological)}, vol.~39, no.~1, pp. 1--22,
  1977.

\bibitem{shen2017dsod}
Z.~Shen, Z.~Liu, J.~Li, Y.-G. Jiang, Y.~Chen, and X.~Xue, ``Dsod: Learning
  deeply supervised object detectors from scratch,'' in \emph{Proceedings of
  the IEEE international conference on computer vision}, 2017, pp. 1919--1927.

\bibitem{8078500}
D.~Anisimov and T.~Khanova, ``Towards lightweight convolutional neural networks
  for object detection,'' in \emph{2017 14th IEEE International Conference on
  Advanced Video and Signal Based Surveillance (AVSS)}, 2017, pp. 1--8.

\bibitem{Wang_2019}
\BIBentryALTinterwordspacing
T.~Wang, L.~Yuan, X.~Zhang, and J.~Feng, ``Distilling object detectors with
  fine-grained feature imitation,'' \emph{2019 IEEE/CVF Conference on Computer
  Vision and Pattern Recognition (CVPR)}, Jun 2019. [Online]. Available:
  \url{http://dx.doi.org/10.1109/CVPR.2019.00507}
\BIBentrySTDinterwordspacing

\bibitem{li2020eagleeye}
B.~Li, B.~Wu, J.~Su, and G.~Wang, ``Eagleeye: Fast sub-net evaluation for
  efficient neural network pruning,'' in \emph{European conference on computer
  vision}.\hskip 1em plus 0.5em minus 0.4em\relax Springer, 2020, pp. 639--654.

\bibitem{sfp}
\BIBentryALTinterwordspacing
Y.~He, G.~Kang, X.~Dong, Y.~Fu, and Y.~Yang, ``Soft filter pruning for
  accelerating deep convolutional neural networks,'' in \emph{IJCAI}, 2018, pp.
  2234--2240. [Online]. Available:
  \url{https://doi.org/10.24963/ijcai.2018/309}
\BIBentrySTDinterwordspacing

\bibitem{chu2020fair}
X.~Chu, T.~Zhou, B.~Zhang, and J.~Li, ``Fair darts: Eliminating unfair
  advantages in differentiable architecture search,'' in \emph{European
  conference on computer vision}.\hskip 1em plus 0.5em minus 0.4em\relax
  Springer, 2020, pp. 465--480.

\bibitem{chu2020darts}
X.~Chu, X.~Wang, B.~Zhang, S.~Lu, X.~Wei, and J.~Yan, ``Darts-: Robustly
  stepping out of performance collapse without indicators,'' in
  \emph{International Conference on Learning Representations}, 2020.

\bibitem{wang2021scaled}
C.-Y. Wang, A.~Bochkovskiy, and H.-Y.~M. Liao, ``Scaled-yolov4: Scaling cross
  stage partial network,'' in \emph{Proceedings of the IEEE/CVF Conference on
  Computer Vision and Pattern Recognition}, 2021, pp. 13\,029--13\,038.

\bibitem{tan2020efficientdet}
M.~Tan, R.~Pang, and Q.~V. Le, ``Efficientdet: Scalable and efficient object
  detection,'' in \emph{Proceedings of the IEEE/CVF conference on computer
  vision and pattern recognition}, 2020, pp. 10\,781--10\,790.

\bibitem{yolox2021}
Z.~Ge, S.~Liu, F.~Wang, Z.~Li, and J.~Sun, ``Yolox: Exceeding yolo series in
  2021,'' \emph{arXiv preprint arXiv:2107.08430}, 2021.

\bibitem{hrnet}
J.~Wang, K.~Sun, T.~Cheng, B.~Jiang, C.~Deng, Y.~Zhao, D.~Liu, Y.~Mu, M.~Tan,
  X.~Wang \emph{et~al.}, ``Deep high-resolution representation learning for
  visual recognition,'' \emph{IEEE transactions on pattern analysis and machine
  intelligence}, 2020.

\bibitem{He_2019_CVPR}
Y.~He, P.~Liu, Z.~Wang, Z.~Hu, and Y.~Yang, ``Filter pruning via geometric
  median for deep convolutional neural networks acceleration,'' in
  \emph{Proceedings of the IEEE/CVF Conference on Computer Vision and Pattern
  Recognition (CVPR)}, June 2019.

\bibitem{Lin_2019_CVPR}
S.~Lin, R.~Ji, C.~Yan, B.~Zhang, L.~Cao, Q.~Ye, F.~Huang, and D.~Doermann,
  ``Towards optimal structured cnn pruning via generative adversarial
  learning,'' in \emph{Proceedings of the IEEE/CVF Conference on Computer
  Vision and Pattern Recognition (CVPR)}, June 2019.

\bibitem{zhou2019accelerate}
Y.~Zhou, Y.~Zhang, Y.~Wang, and Q.~Tian, ``Accelerate cnn via recursive
  bayesian pruning,'' in \emph{Proceedings of the IEEE/CVF International
  Conference on Computer Vision}, 2019, pp. 3306--3315.

\bibitem{dong2019tas}
X.~Dong and Y.~Yang, ``Network pruning via transformable architecture search,''
  in \emph{Neural Information Processing Systems (NeurIPS)}, 2019.

\bibitem{luo2020autopruner}
J.-H. Luo and J.~Wu, ``Autopruner: An end-to-end trainable filter pruning
  method for efficient deep model inference,'' \emph{Pattern Recognition}, vol.
  107, p. 107461, 2020.

\bibitem{liebenwein2019provable}
L.~Liebenwein, C.~Baykal, H.~Lang, D.~Feldman, and D.~Rus, ``Provable filter
  pruning for efficient neural networks,'' in \emph{International Conference on
  Learning Representations}, 2019.

\bibitem{tang2020scop}
Y.~Tang, Y.~Wang, Y.~Xu, D.~Tao, C.~Xu, C.~Xu, and C.~Xu, ``Scop: Scientific
  control for reliable neural network pruning,'' \emph{Advances in Neural
  Information Processing Systems}, vol.~33, pp. 10\,936--10\,947, 2020.

\bibitem{NPPM}
S.~Gao, F.~Huang, W.~Cai, and H.~Huang, ``Network pruning via performance
  maximization,'' in \emph{Proceedings of the IEEE/CVF Conference on Computer
  Vision and Pattern Recognition}, 2021, pp. 9270--9280.

\bibitem{DCP}
J.~Liu, B.~Zhuang, Z.~Zhuang, Y.~Guo, J.~Huang, J.~Zhu, and M.~Tan,
  ``Discrimination-aware network pruning for deep model compression,''
  \emph{IEEE Transactions on Pattern Analysis and Machine Intelligence},
  vol.~44, no.~8, pp. 4035--4051, 2022.

\bibitem{sui2021chip}
Y.~Sui, M.~Yin, Y.~Xie, H.~Phan, S.~Aliari~Zonouz, and B.~Yuan, ``Chip: Channel
  independence-based pruning for compact neural networks,'' \emph{Advances in
  Neural Information Processing Systems}, vol.~34, pp. 24\,604--24\,616, 2021.

\bibitem{deng2009imagenet}
J.~Deng, W.~Dong, R.~Socher, L.-J. Li, K.~Li, and L.~Fei-Fei, ``Imagenet: A
  large-scale hierarchical image database,'' in \emph{CVPR}, 2009.

\bibitem{He_2018_ECCV}
Y.~He, J.~Lin, Z.~Liu, H.~Wang, L.-J. Li, and S.~Han, ``Amc: Automl for model
  compression and acceleration on mobile devices,'' in \emph{Proceedings of the
  European Conference on Computer Vision (ECCV)}, September 2018.

\bibitem{He_2020_CVPR}
Y.~He, Y.~Ding, P.~Liu, L.~Zhu, H.~Zhang, and Y.~Yang, ``Learning filter
  pruning criteria for deep convolutional neural networks acceleration,'' in
  \emph{Proceedings of the IEEE/CVF Conference on Computer Vision and Pattern
  Recognition (CVPR)}, June 2020.

\bibitem{cifar10}
A.~Krizhevsky \emph{et~al.}, ``Learning multiple layers of features from tiny
  images,'' 2009.

\end{thebibliography}









\clearpage
\setcounter{page}{1}
\setcounter{equation}{0}
\setcounter{figure}{0}
\setcounter{table}{0}
\renewcommand\thefigure{A.\arabic{figure}}
\renewcommand\theequation{A.\arabic{equation}}
\renewcommand\thetable{A.\arabic{table}}
\appendices
\section{Supplementary Theory Justification}
\subsection{The Rationality of Using Norm over the State Change}
\label{norm}
The reason for using norm over the state change in Eq. (\ref{sensitivity}) is for measuring the distance between two high-dimension states, \textit{e.g.}, the feature maps or logits vectors, which is more concerned than the state improvement in the saliency computation. Usually, for high-dimension states in the intermediate layers, as they are not directly related with the model predication loss, it is thus difficult to judge whether their change direction is correct and the state is improved (notes: not the improvement of the saliency) by pruning some filters. On the other hand, for those task-related states such as the model prediction loss, they may be considered as improved if becoming smaller by zeroing (pruning) some filters. In such case, to preserve the distance measurement ability of the filters improving the state, meanwhile keeping the ability to assign the maximal pruning probability to the filters harming the state, we can slightly tune the state function into Eq. (\ref{finetune_state}):
\begin{equation}
    \boldsymbol{f}'(x;\theta_{lk}) =\max( \boldsymbol{f}(x,\theta _{lk}) ,\boldsymbol{f}(x,\theta _{lk}=0)). 
\label{finetune_state}
\end{equation}

By replacing $\boldsymbol{f}\left( x,\theta _{lk} \right)$ with $\boldsymbol{f}'\left( x;\theta _{lk} \right)$ in Eq. (\ref{sensitivity}) of the manuscript, those filters harming the state will be assigned with the lowest saliency, \textit{i.e.}, $\mathcal{S} =0$. Consequently, $\mathcal{P}$ of these filters in Eq. \ref{joint_probability}) will be mapped to 1, \textit{i.e.}, the maximal pruning probability.

In summary, for most state criteria, the norm of the state change is to measure the distance of the state change for high-dimension states. Considering some special cases like filters damaging the state, we can tune the state function like Eq. (\ref{finetune_state}) to maintain the lowest saliency for them, which will not affect the general form of the saliency criterion.

\subsection{The Proof of Theorem \ref{thm1}}
\label{proof}
To prove Theorem \ref{thm1}, according to the definition in Eq. (\ref{joint_sensitivity}), the joint saliency can be derived as follows:
\begin{equation}
\begin{aligned}
\mathcal{S}_{pq}^{uv}\left( x \right)& =\lVert \boldsymbol{f}\left( x;\theta _{pu},\theta _{qv}=1 \right) -\boldsymbol{f}\left( x;\theta _{pu},\theta _{qv}=0 \right) \rVert \\
&=\left.\lVert \boldsymbol{f}\left( x;\theta _{pu},\theta _{qv}=1 \right) -\boldsymbol{f}\left( x;\theta _{pu}=0 \right) + \right.\\
&\phantom{=\;\;}\left.\boldsymbol{f}\left( x;\theta _{pu}=0 \right) -\boldsymbol{f}\left( x;\theta _{qv}=1 \right) + \right.\\
&\phantom{=\;\;}\left.\boldsymbol{f}\left( x;\theta _{qv}=1 \right) -\boldsymbol{f}\left( x;\theta _{pu},\theta _{qv}=0 \right) \rVert \right.\\
&=\left.\lVert \boldsymbol{f}\left( x;\theta _{pu}=0 \right) -\boldsymbol{f}\left( x;\theta _{pu}=0 \right) + \right.\\
&\phantom{=\;\;}\left.\boldsymbol{f}\left( x;\theta _{qv}=1 \right) -\boldsymbol{f}\left( x;\theta _{pu},\theta _{qv}=0 \right) \rVert \right.\\
&\leqslant \left.\lVert \boldsymbol{f}\left( x;\theta _{qv}=1 \right) -\boldsymbol{f}\left( x;\theta _{pu}=0 \right) \rVert + \right.\\
&\phantom{=\;\;}\left.\lVert \boldsymbol{f}\left( x;\theta _{pu}=0 \right) -\boldsymbol{f}\left( x;\theta _{pu},\theta _{qv}=0 \right) \rVert \right.\\
&=\left.\lVert \boldsymbol{f}\left( x;\theta _{pu}=1 \right) -\boldsymbol{f}\left( x;\theta _{pu}=0 \right) \rVert + \right.\\
&\phantom{=\;\;}\left.\lVert \boldsymbol{f}\left( x;\theta _{pu}=\text{0},\theta _{qv}=1 \right) -\boldsymbol{f}\left( x;\theta _{pu},\theta _{qv}=0 \right) \rVert \right.\\
&=\mathcal{S}_{p}^{u}\left( x \right) +\mathcal{S}_{q}^{v}|_{p}^{u}\left( x \right). 
\end{aligned}
\end{equation}

The proof utilizes the property that $\boldsymbol{f}\left( x;\theta_{pu}=1\right)=\boldsymbol{f}\left(x;\theta _{qv}=1 \right)=\boldsymbol{f}\left(x;\theta_{pu},\theta_{qv}=1\right)$ as well as the in-equation of absolute value in the form of vectors. Eq. (\ref{conditional_sensitivity}) can also be written as: $\mathcal{S}_{pq}^{uv}\left( x \right) \leqslant \mathcal{S}_{q}^{v}\left( x \right) + \mathcal{S}_{p}^{u}|_{q}^{v}\left( x \right)$ for the symmetry.
\subsection{The Rationality of the Exponential Mapping for $\mathcal{P}$}
\label{exponential}
The idea of exponential mapping in Eq. (\ref{joint_probability}) is inspired by the classification task, where output logits are mapped through the \textit{softmax} operation to a probability distribution vector with each dimension representing the prediction probability score for each class. Similar to the classification task, the saliency-based channel pruning task can be considered as a filter-wise binary classification task (1 for pruning, 0 for not), where the filter saliency criterion represents the mapping rule and outputs the saliency of the corresponding input filter(s).

To correlate the saliency with the prediction confidence, we need to map the original saliency value to a new probability space where the probability denotes the filter pruning chance. Since a higher filter saliency indicates a more important filter and less chances of pruning, the filter saliency is thus considered as negatively related to the probability of pruning the filter. Inspired by the \textit{softmax} operator, which maps the logits positively to the probability space by the exponential reweight of logits, we intuitively map the saliency to the pruning probability space by the negative exponential reweight. Note that the saliency computed by the norm of the state change is non-negative, which constrains the exponentiated output $\mathcal{P}$ of each filter to be between 0 and 1. Before mapping, we normalize each filter saliency by the filter size to avoid the effect of filter size differences. Therefore, for each filter, we hold that $\mathcal{P}$ might be a naïve but appropriate substitute to the \textit{sigmoid} or \textit{softmax} operator in this binary classification task. The final decision of whether to prune the high-$\mathcal{P}$ filter should consider the $\mathcal{P}$ distribution of the whole filter set, \textit{e.g.}, in a joint way like our method.

\subsection{The Mechanism of Approximating the Maximal Value of $\mathcal{P}_{pq}^{uv}$}
\label{EM}
The theoretical foundation of this approximation in Section \ref{approximation} is based on the bound optimization, which aims at using the bound function to optimize the objective if the bound objective is easier to optimize. Expectation-Maximization (EM) algorithm for maximum likelihood learning is a popular bound optimizer, which iteratively pushes up the evidence lower bound (ELBO) of the marginal likelihood to maximize the marginal likelihood \cite{dempster1977maximum}. Similar to the EM algorithm, the approximation mechanism of $\mathcal{P}_{pq}^{uv}$ can be decomposed into two steps:

\textbf{Step 1}. Find a contact point for the values of $\theta_{pu}, \theta_{qv}$ such that $\mathcal{P} _{pq}^{uv}\left( x \right) =\mathcal{P} _{p}^{u}\left( x \right) \mathcal{P} _{q}^{v}|_{p}^{u}\left( x \right)$.

\textbf{Step 2}. Adjust $\theta_{pu}, \theta_{qv}$ to maximize $\mathcal{P} _{p}^{u}\left( x \right) \mathcal{P} _{q}^{v}|_{p}^{u}\left( x \right)$ .

\begin{figure}[htbp]
    \centering
    \resizebox{\columnwidth}{!}{
    \includegraphics{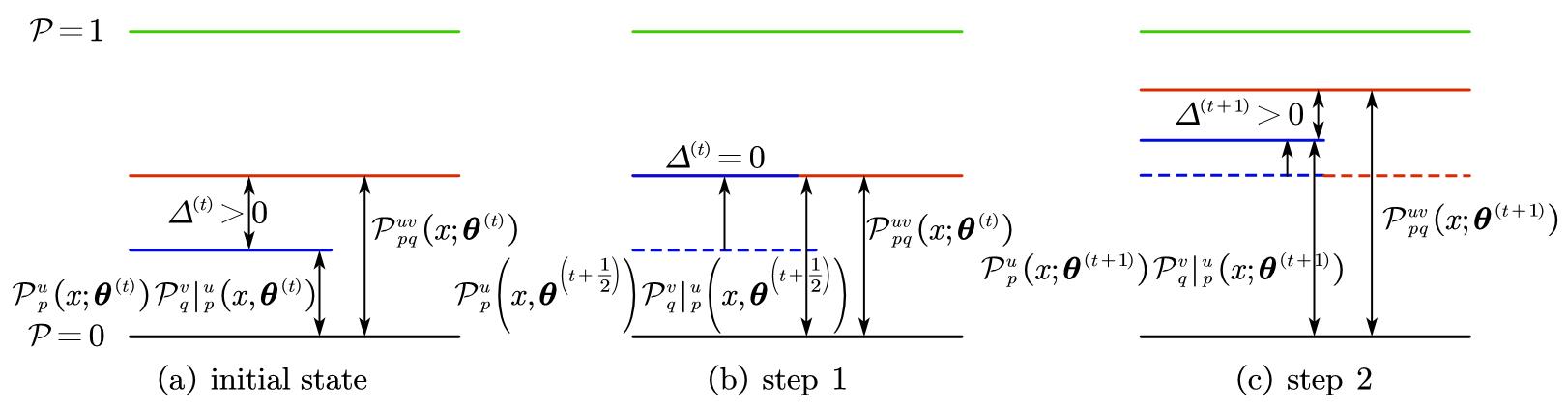}}
    \caption{The illustration of the \textit{t}-th optimization iteration in approximating the maximum of $\mathcal{P} _{pq}^{uv}(x)$.}
    \label{maximal_approximation}
\end{figure}

Through repeating the above two steps, $\mathcal{P} _{pq}^{uv}(x)$ will converge to a local maximum that is upper bounded by $\mathcal{P}=1$. Fig. \ref{maximal_approximation} illustrates the \textit{t}-th optimization iteration. Therefore, when maximizing $\mathcal{P} _{p}^{u}\left( x \right) \mathcal{P} _{q}^{v}|_{p}^{u}\left( x \right)$, $\mathcal{P} _{pq}^{uv}(x)$ is gradually pushed up more closely to the upper bound $\mathcal{P}=1$, approximating the local maximum.

\subsection{The Rationality of the Intra-layer Filter Saliency Independence}
\label{independence}
The independence of filter saliency mentioned in Section \ref{independence_analysis} varies in different designs of the saliency criterion. Existing saliency criteria can be roughly categorized as filter-guided and prediction-guided. Filter-guided saliency criteria judge the importance of filters from the model parameters, including the filter weights \cite{KDSG17}, filter scaling factors \cite{liu2017learning}, and the gradients \cite{8953464}, \textit{etc.}, while prediction-guided saliency criteria evaluate the filter importance from the output of either the model or filters, \textit{i.e.}, the prediction loss or feature maps \cite{9156677}. The former saliency criteria are considered to follow intra-layer filter saliency independence, since the elements for computing the filter saliency are each filter's own weights or parameters independent of the distribution of other intra-layer filters. In contrast, the latter saliency criteria do not follow such independence property since the element for computing each intra-layer filter saliency, \textit{i.e.}, the prediction loss or each output feature map is jointly determined by a group of intra-layer filters on the input feature maps. Since our filter saliency criterion originates from the $\ell_1$ norm of filter weights, one type of the filter-guided saliency criteria, the assumption of the intra-layer filter saliency thus holds in our work.

\section{Supplementary Experiments}
\subsection{The Influence of Different Pruning Ratios on the Pruning Sequence}
\label{pruning_sequence}
\begin{figure}[htbp]
    \centering
    \resizebox{\columnwidth}{!}{
    \includegraphics{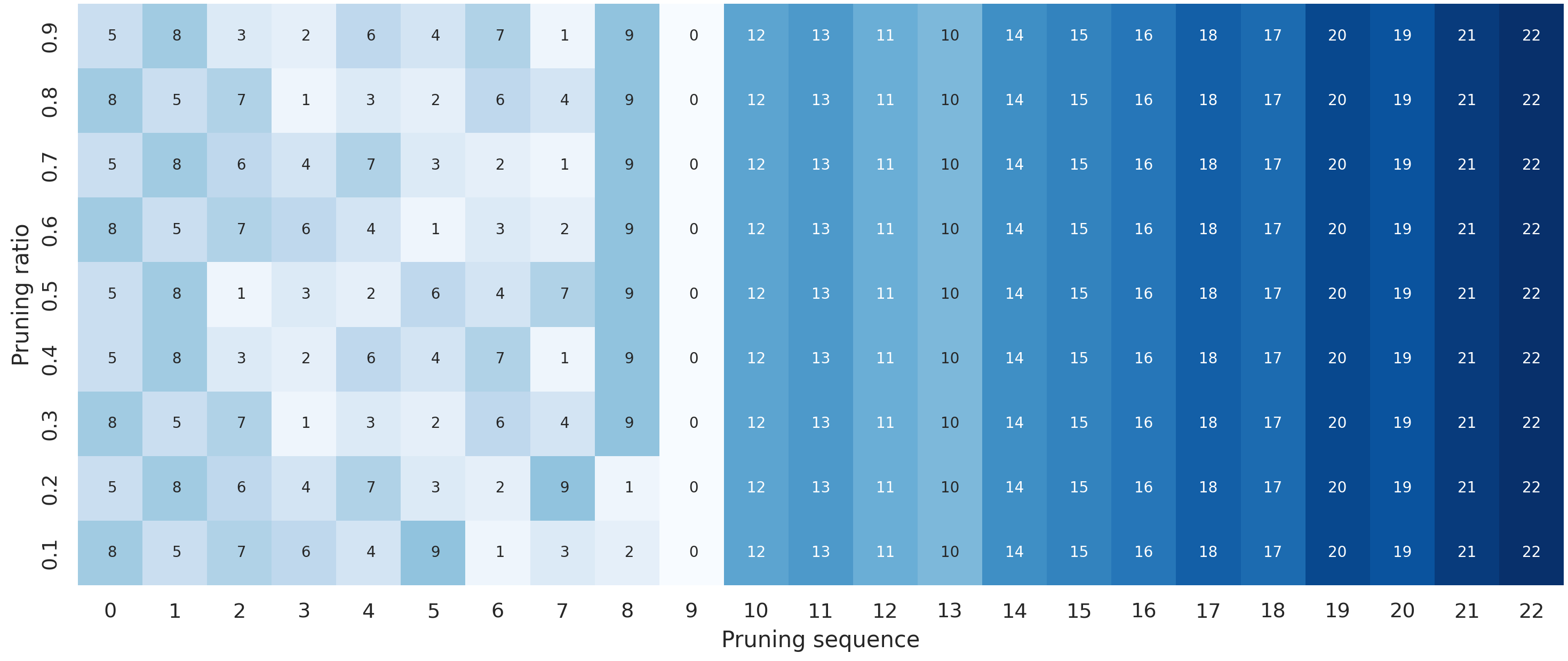}}
    \caption{The influence of different pruning ratios on the pruning sequence. The value in each grid represents the layer index in the original model, with the horizontal and vertical axis representing its index in the pruning sequence and the pruning ratio for this layer, respectively. The experiments are based on SSD300.}
    \label{sequence_matrix}
\end{figure}

We compare the sorting differences of compression contribution for each layer with different pruning ratios, as visualized in Fig. \ref{sequence_matrix}. It can be observed that most layers are insensitive to the pruning ratio setting, located at constant pruning positions (from the 9th to 22nd pruning index in the horizontal axis). The impact of the rest pruning layers on compression performance has been explored in Section \ref{ablation}.

\begin{table}[t]
\centering

\caption{Compression results of ResNet56 and ResNet110 on CIFAR-10.}
\label{cifar10}
\resizebox{\columnwidth}{!}{
\begin{tabular}{@{}llcccc@{}}
\toprule
Model & Method & \begin{tabular}[c]{@{}c@{}}Base\\ Top-1 (\%)\end{tabular} & \begin{tabular}[c]{@{}c@{}}Pruned\\ Top-1 (\%)\end{tabular} & \begin{tabular}[c]{@{}c@{}}Top-1\\ ↓(\%)\end{tabular} & \begin{tabular}[c]{@{}c@{}}FLOPs\\ ↓(\%)\end{tabular} \\ \midrule
\multirow{7}{*}{R56} & AMC\cite{He_2018_ECCV} & 92.8 & 91.9 & 0.9 & 50.0 \\
 & FPGM\cite{He_2019_CVPR} & 93.59 & 93.26 & 0.33 & 52.6 \\
  & TAS \cite{dong2019tas} & 94.46 & 93.69 & 0.77 & 52.7\\
 & SFP\cite{sfp} & 93.59 & 93.35 & 0.24 & 52.6 \\
 & LFPC\cite{He_2020_CVPR} & 93.59 & 93.24 & 0.35 & 52.9 \\
 & HRank\cite{9156677} & 93.26 & 93.17 & 0.09 & 42.4 \\
 & ResRep\cite{resrep} & 93.71 & 93.71 & 0.00 & 52.9 \\
 & Ours & \textbf{93.26} & \textbf{93.71} & \textbf{-0.35} & \textbf{56.6} \\\midrule
\multirow{5}{*}{R110} & $\ell_1$ norm \cite{KDSG17} & 93.53 & 93.30 & 0.23 & 38.6 \\
 & GAL-0.5\cite{Lin_2019_CVPR} & 93.50 & 92.74 & 0.76 & 48.5 \\
 & TAS \cite{dong2019tas} & 94.97 & 94.33 & 0.64 & 53.0 \\
 & HRank\cite{9156677} & 93.50 & 93.36 & 0.14 & 58.2 \\
 & ResRep\cite{resrep} & 94.64 & 94.62 & 0.02 & 58.2 \\
 & Ours & \textbf{93.50} & \textbf{93.81} & \textbf{-0.31} & \textbf{70.0} \\ \bottomrule
\end{tabular}}
\end{table}

\subsection{CIFAR-10 Benchmark}
\label{cifar_benchmark}
In addition to the ImageNet benchmark, we also conduct experiments with ResNet56 \cite{resnet} and ResNet110 \cite{resnet} on CIFAR-10 \cite{cifar10}. The base models are consistent with the models in HRank \cite{9156677}, and we fine-tune the pruned models for 200 epochs with the initial learning rate of 0.1 and batch size of 128. Table \ref{cifar10} shows the superiority of PAGCP. The pruned models achieve +0.3\% Top-1 accuracy increase with +56\% FLOPs reduction on both models, which outperform the state-of-the-art method ResRep although the base accuracy is biased in each method. An exciting observation is that for ResNet 110, we achieve 70\% FLOPs reduction in one pruning iteration and still have the accuracy gain by 0.31\%. These results also validate the superiority of our method in single-task model compression.

\begin{figure}[ht]
    \centering
    \resizebox{\columnwidth}{!}{
    \includegraphics{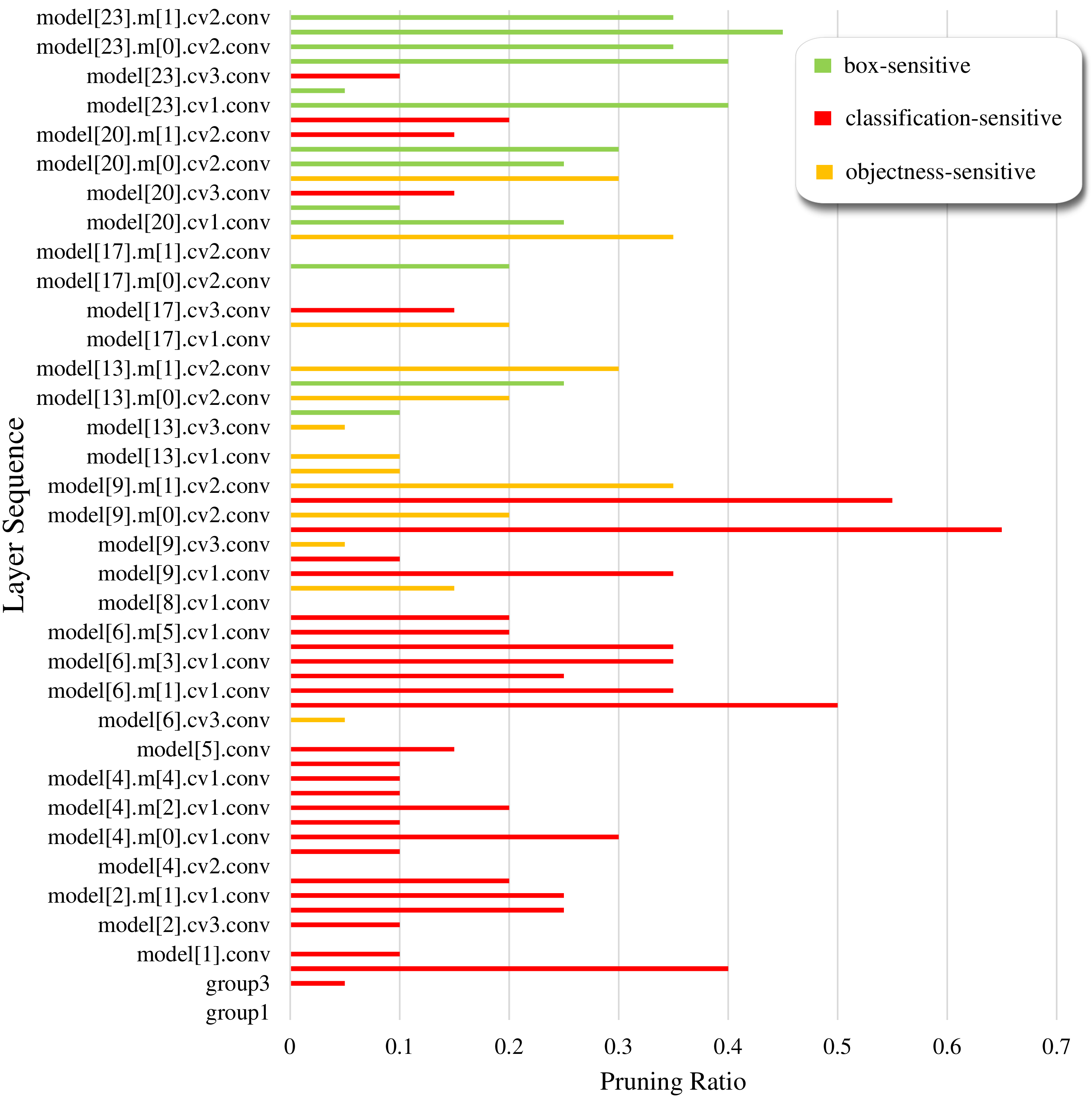}}
    \caption{The sensitivity distribution of each layer in YOLOv5 on COCO.}
    \label{yolov5_sens}
\end{figure}

\begin{figure}[htbp]
    \centering
    \resizebox{\columnwidth}{!}{
    \includegraphics{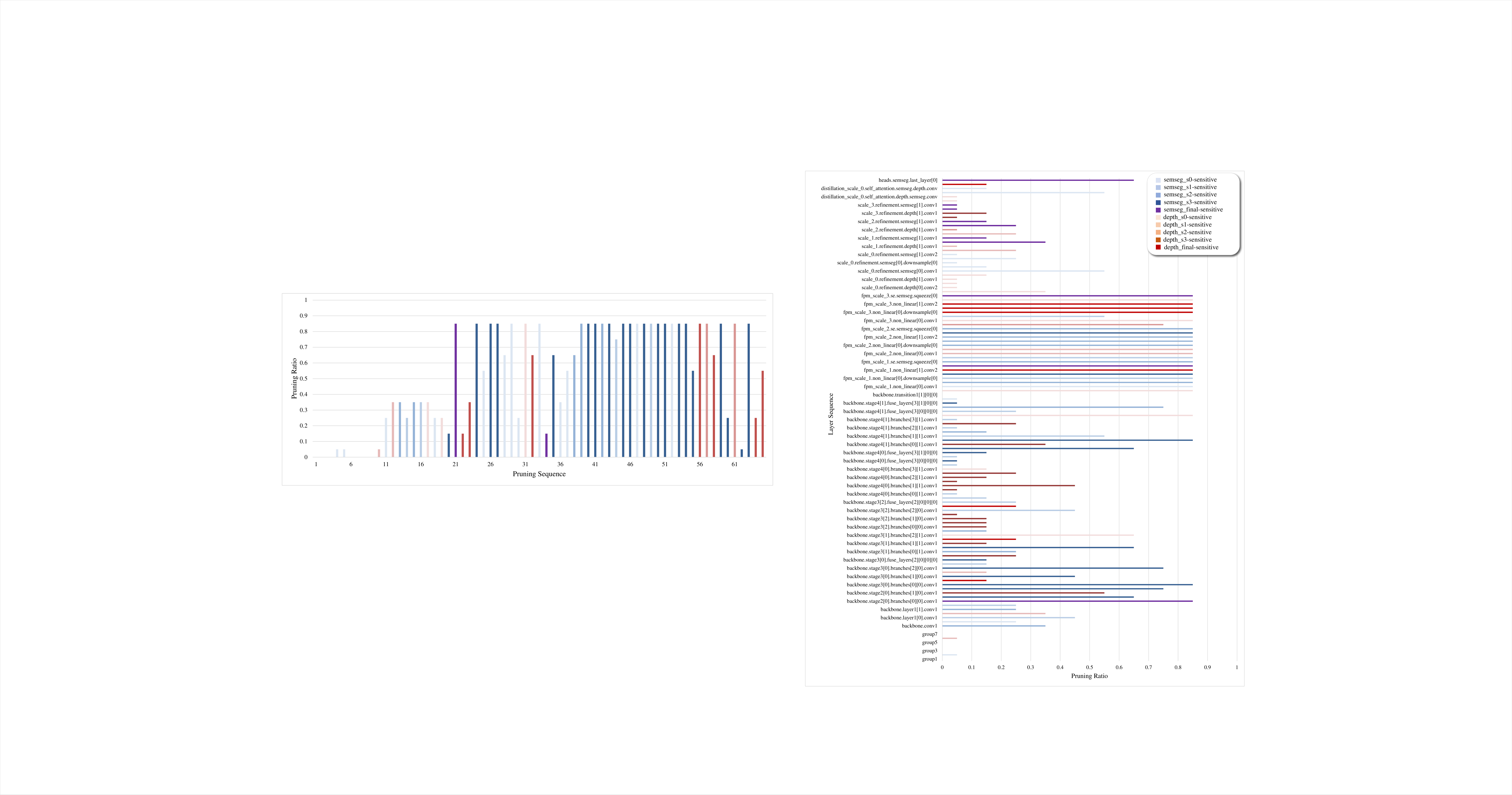}}
    \caption{The sensitivity distribution of MTI-Net on NYUD-v2 with ten tasks, where `\_s\textit{x}\_' refers to the initial prediction at scale \textit{x} and `\_final\_' refers to the prediction at the final stage.}
    \label{mti_sens}
\end{figure}

\subsection{The Distribution of the Most Sensitive Task along the Process}
\label{distribution}
Besides Fig. \ref{sensitivity_distribution}, we provide more sensitivity distributions on other two benchmarks, as shown in Fig. \ref{yolov5_sens} and Fig. \ref{mti_sens}. Fig. \ref{yolov5_sens} shows the sensitivity distribution of each layer in YOLOv5 on COCO, while Fig. \ref{mti_sens} presents the distribution of MTI-Net on NYUD-v2. From these results, we can further observe a common characteristic that shallow layers usually have higher chances to become classification-sensitive, while deep layers have higher chances to become regression-sensitive. For YOLOv5, shallow layers are more sensitive to the classification-oriented tasks (including objectness and object classifications), while deep layers are more sensitive to the regression-oriented task (box regression). 
For MTI-Net, although there are more tasks than SSD300 and YOLOv5, it shares the same characteristic that shallow layers are more sensitive to classification-oriented tasks (including segmentation tasks of multiple scales), while deep layers are more sensitive to regression-oriented tasks (including depth estimation tasks of multiple scales). In addition, the pruning of MTI-Net shows one more characteristic that, the pruning ratio of layers sensitive to large-scale prediction tasks is averagely larger than that of ones sensitive to small-scale tasks (represented by longer dark bars in the chart).

\subsection{Visualization of Pruned and Preserved Filters}
\begin{figure}[htbp]
    \centering
    \resizebox{\linewidth}{!}{
    \includegraphics{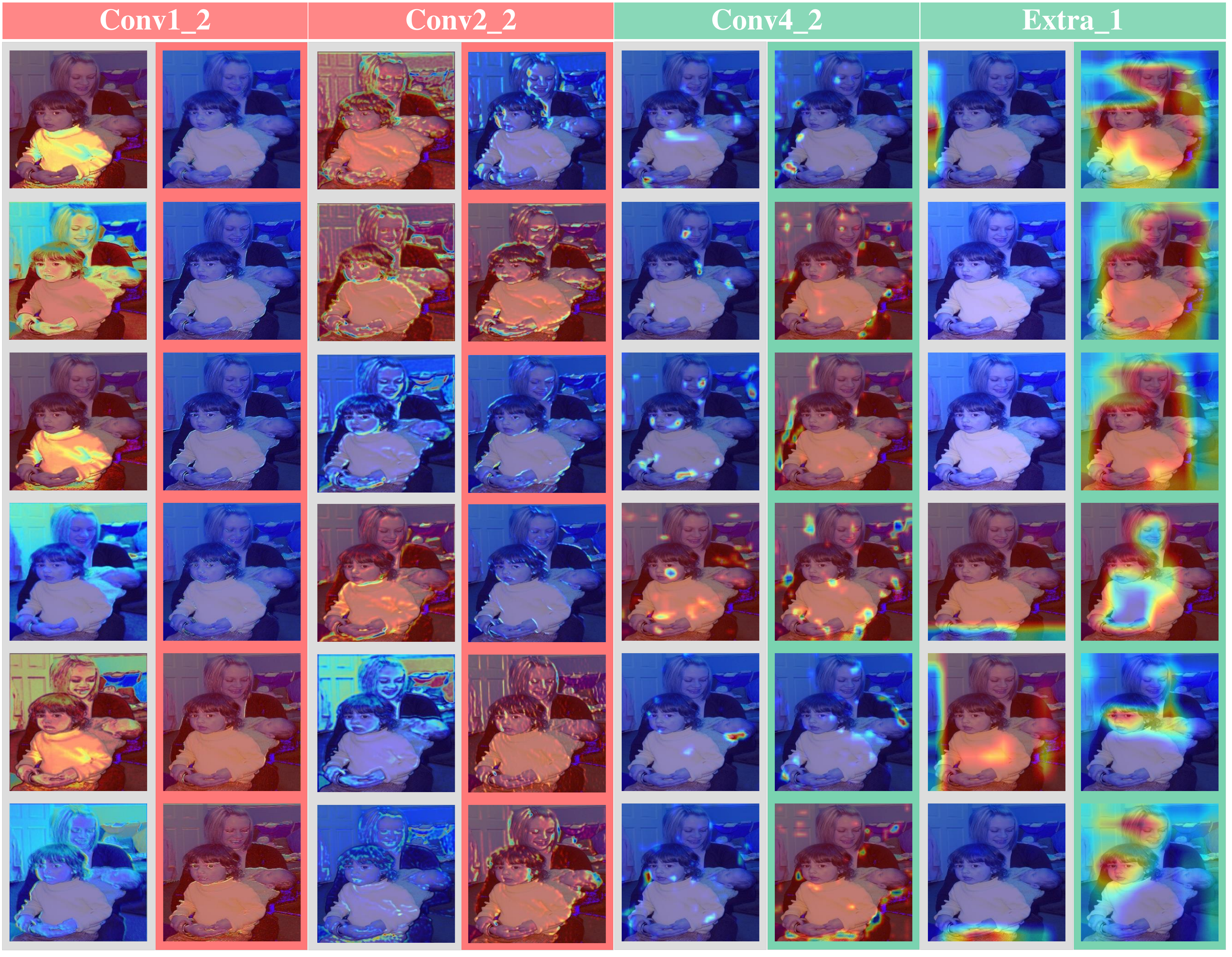}}
    \caption{Visualization of the attention maps generated by pruned (first column for each conv layer) and preserved (second column for each conv layer) filters in different task-sensitive layers of SSD300 on PASCAL VOC, where gray-grounded maps (Column 1, 3, 5, 7) refer to the dropped ones, red-grounded maps (Column 2, 4) refer to the preserved ones in the classification-sensitive layers, and green-grounded maps (Column 6, 8) refer to the preserved ones in the regression-sensitive layers.}
    \label{gradcam}
\end{figure}
We take Layer Conv1\_2, Conv2\_2, Conv4\_2, and Extra\_1 of SSD300 as examples to visualize the pruned and preserved filters, which represent two different classification-sensitive and two different regression-sensitive layers, respectively, according to previous experimental results as shown in Fig. \ref{sensitivity_distribution}. Fig. \ref{gradcam} shows the visualization results, where we utilize GradCAM to visualize attention maps generated by pruned and preserved filters in these four layers under our criterion. It can be observed that our proposed method focuses on preserving filters that can generate higher task-related responses for two types of layers. For example, in classification-sensitive layers (Conv1\_2 and Conv 2\_2), the selected filters have higher responses on spatial details of persons than those dropped filters, and these detail features are preserved in more refined attention areas, keeping shallow layers robust to the regression task and sensitive to the classification task. While in regression-sensitive layers (Conv4\_2 and Extra\_1), more semantic context information of persons is activated by the selected filters with high quality (represented by more rich and meaningful contexts), thus keeping deep layers robust to the classification task and sensitive to the regression task.

\subsection{Pruning Complexity of PAGCP}
\label{pruning_complexity}
We first provide the theoretical complexity measure and then analyze the most important elements for the complexity measure according to the experimental results. Given the layer number \textit{N} to be pruned, the filter number $K_i$ of the \textit{i}-th pruning layer, the mask ratio $\gamma$ per pruning in a layer, the probability $p_i\left( j \right)$ of pruning top-\textit{j} least important intra-layer filters in the \textit{i}-th layer, and $t_i\left( j \right)$ denoting the total inference time of all batches images by the pruned model according to $p_i\left( j \right)$, the theoretical expectation of pruning complexity can be computed as follows:
\begin{equation}
    E\left( C \right) =\sum_{i=1}^N{\sum_{j=1}^{\left[ K_i/\gamma \right] -1}{p_i\left( j \right) \cdot j\cdot t_i\left( j \right)}.}
\label{complexity}
\end{equation}

From Eq. \ref{complexity}, the complexity \textit{C} is related to multiple variants, including \textit{t}, \textit{p}, $K_i$, $\gamma$, and \textit{N}. To provide a clear illustration, we show the pruning time consumption of different models on different benchmarks, as listed in Table \ref{consumption}. We report the FLOPs drop rate to indicate the influence of $\{K, \gamma, p, t\}$ on \textit{C}, as these parameters cooperate to affect the pruning efficiency in a layer. It can be observed that with the increase of the to-prune layer number, the time also increases. Note that the pruning occurs in the training stage, its complexity is thus less concerned than the inference time of the pruned model.
\begin{table}[t]
\centering
\caption{Pruning time consumption of YOLOv5m and MTI-Net on different benchmarks.}
\label{consumption}
\resizebox{\columnwidth}{!}{%
\begin{tabular}{@{}llccc@{}}
\toprule
Model & Benchmark & N & FLOPs↓(\%) & Time (hour) \\ \midrule
YOLOv5m & COCO & 69 & 30 & $\sim$2 \\
MTI-Net & NYUD-v2 & 117 & 33 & $\sim$4 \\ \bottomrule
\end{tabular}%
}
\end{table}

\begin{figure}[]
    \centering
    \resizebox{\columnwidth}{!}{
    \includegraphics{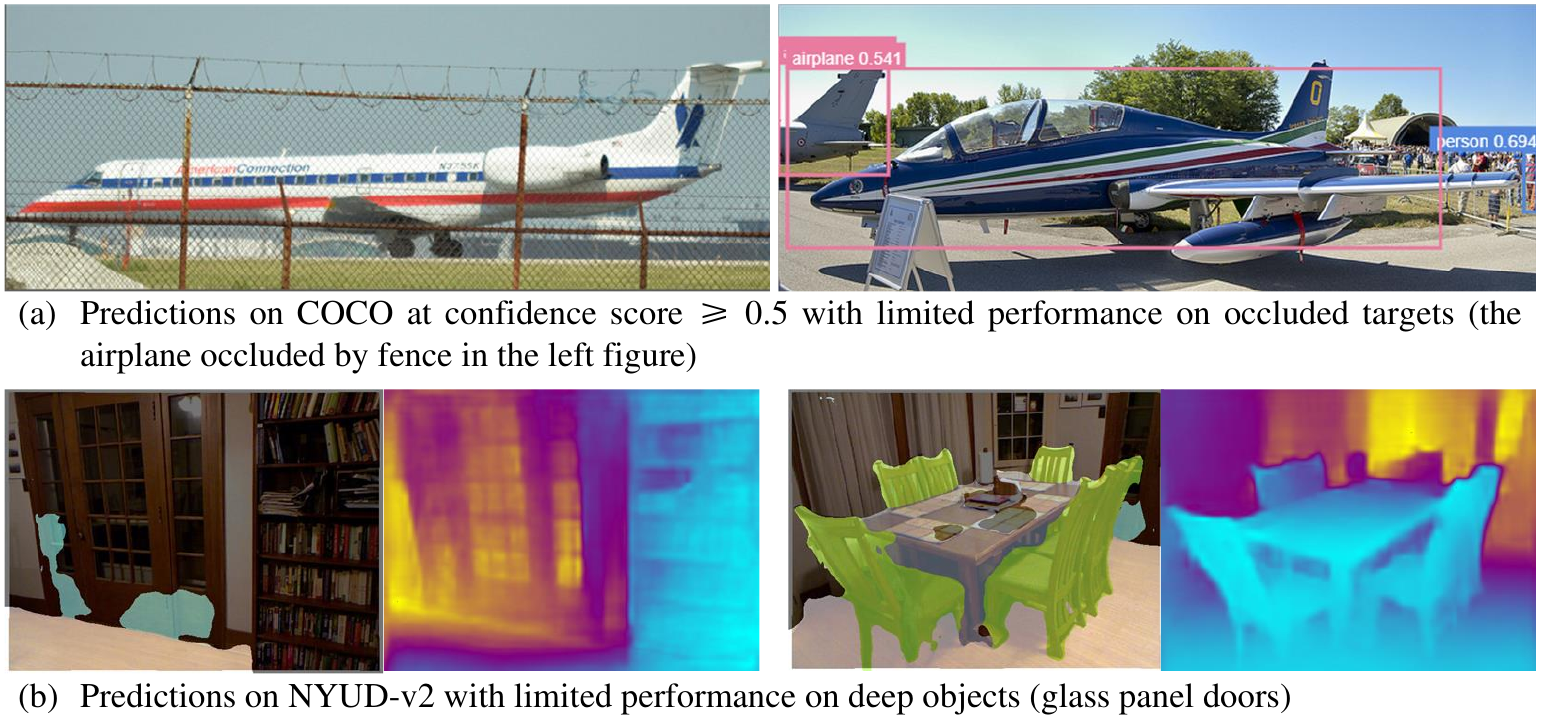}}
    \caption{Some failure cases on two multitask benchmarks. (a) The detection results of the pruned model for airplane objects, where the number on the top left of each bounding box refers to the prediction confidence score. (b) The multitask prediction results of the pruned model for glass panel doors, where Column 1 and 3 are the semantic segmentation maps and the other two columns (Column 2 and 4) are the corresponding depth maps.}
    \label{limitation}
\end{figure}
\section{Limitation Discussion}
\label{limitation_discussion}
The results in Fig. \ref{limitation}-(a) show the failure of the pruned model for predicting the noisy objects, that is, when the object is severely occluded by other objects, the pruned model may detect it with low confidence or miss it. The reason may be the missing of some deep-layer features robust to interference during the filter pruning. Another example in Fig. \ref{limitation}-(b) shows the failure of the pruned model for segmenting far-away objects. The reason may be that the far-away objects with large depth variation may interfere the semantic segmentation for the nearby glass panel doors.

\end{document}